\RequirePackage{fix-cm}
\documentclass[smallextended]{svjour3} 
\smartqed
\usepackage[english]{babel}
\usepackage[utf8]{inputenc}
\usepackage{url}
\usepackage{tikz}
\usepackage{graphicx} 
\usepackage{amssymb}
\usepackage{multirow}
\usepackage{adjustbox}
\usepackage{booktabs}
\usepackage{tabularx}
\usepackage{natbib}
\usetikzlibrary{quotes,arrows.meta,positioning,decorations.pathreplacing,calc,3d,arrows}
\usepackage{subfig}
\usepackage[misc]{ifsym}

\usepackage[keeplastbox]{flushend}
\usepackage[a4paper=false,
            citecolor=blue,
            colorlinks=true,
            urlcolor=blue,
            linkcolor=blue,
            pdfauthor={Hassan Ismail Fawaz},
            pdftitle={Deep learning for time series classification: a review},
            pdfsubject={Deep learning for time series classification: a review},
            pdfkeywords={deep learning, time series, classification, review}
            ]{hyperref}
\usepackage{textcomp}
\usepackage{tikz}
\newcommand\copyrighttext{%
\small This is the author's version of an article published in Data Mining and Knowledge Discovery.
The final authenticated version is available online at: \url{https://doi.org/10.1007/s10618-019-00619-1}.}
\newcommand\copyrightnotice{%
\begin{tikzpicture}[remember picture,overlay]
\node[anchor=north,yshift=-30pt] at (current page.north) {\fbox{\parbox{\textwidth}{\copyrighttext}}};
\end{tikzpicture}
}

\usepackage[11pt]{extsizes}

\usepackage{geometry}
\newcommand{\subparagraph}{}
\usepackage{titlesec}
\titleformat*{\section}{\sffamily\Large\bfseries}
\titleformat*{\subsection}{\sffamily\large\bfseries}
\titleformat*{\subsubsection}{\sffamily\large\bfseries}

\usepackage{enumitem}
\usepackage{setspace}
\setlist{itemsep=0pt,parsep=0pt}

\begin{document}

\hyphenation{avai-lability}

\def\makeheadbox{\relax}
\newcommand{\gfc}[1]{{\color{magenta}~{\bf [Germain:} #1{\bf ]}}}
\newcommand{\hfc}[1]{{\color{red}~{\bf [Hassan:} #1{\bf ]}}}
\newcommand{\jwc}[1]{{\color{lime}~{\bf [Jonathan:} #1{\bf ]}}}

\title{{\sffamily Deep learning for time series classification: a review}}

\titlerunning{Deep learning for time series classification: a review}  

\author{\sffamily \large Hassan Ismail Fawaz\textsuperscript{1}         \and
        Germain Forestier\textsuperscript{1,2}  \and 
        Jonathan Weber\textsuperscript{1}  \and 
        Lhassane Idoumghar\textsuperscript{1} \and 
        Pierre-Alain Muller\textsuperscript{1}
}

\authorrunning{Hassan Ismail Fawaz et al.} 

\institute{\Letter{} H. Ismail Fawaz  \\ \medskip
              \email{hassan.ismail-fawaz@uha.fr}  \\
              \textsuperscript{1}IRIMAS, Universit\'e Haute Alsace, Mulhouse, France\\\
              \textsuperscript{2}Faculty of IT, Monash University, Melbourne, Australia
}

\date{}

\maketitle           

\copyrightnotice{}   

\vspace{-2cm}
\begin{abstract}
Time Series Classification (TSC) is an important and challenging problem in data mining. 
With the increase of time series data availability, hundreds of TSC algorithms have been proposed.
Among these methods, only a few have considered Deep Neural Networks (DNNs) to perform this task.
This is surprising as deep learning has seen very successful applications in the last years. 
DNNs have indeed revolutionized the field of computer vision especially with the advent of novel deeper architectures such as Residual and Convolutional Neural Networks. 
Apart from images, sequential data such as text and audio can also be processed with DNNs to reach state-of-the-art performance for document classification and speech recognition. 
In this article, we study the current state-of-the-art performance of deep learning algorithms for TSC by presenting an empirical study of the most recent DNN architectures for TSC. 
We give an overview of the most successful deep learning applications in various time series domains under a unified taxonomy of DNNs for TSC. 
We also provide an open source deep learning framework to the TSC community where we implemented each of the compared approaches and evaluated them on a univariate TSC benchmark (the UCR/UEA archive) and 12 multivariate time series datasets. 
By training 8,730 deep learning models on 97 time series datasets, we propose the most exhaustive study of DNNs for TSC to date. 

\end{abstract}

\section{Introduction}
During the last two decades, Time Series Classification (TSC) has been considered as one of the most challenging problems in data mining~\citep{yang200610,esling2012time}. 
With the increase of temporal data availability~\citep{silva2018speeding}, hundreds of TSC algorithms have been proposed since 2015~\citep{bagnall2017the}.
Due to their natural temporal ordering, time series data are present in almost every task that requires some sort of human cognitive process~\citep{langkvist2014a}.  
In fact, any classification problem, using data that is registered taking into account some notion of ordering, can be  cast as a TSC problem~\citep{gamboa2017deep}. 
Time series are encountered in many real-world applications ranging from electronic health records~\citep{rajkomar2018scalable} and human activity recognition~\citep{nweke2018deep,wang2018deep} to acoustic scene classification~\citep{nwe2017convolutional} and cyber-security~\citep{susto2018time}. 
In addition, the diversity of the datasets' types in the UCR/UEA archive~\citep{ucrarchive,bagnall2017the} (the largest repository of time series datasets) shows the different applications of the TSC problem. 

Given the need to accurately classify time series data, researchers have proposed hundreds of methods to solve this task~\citep{bagnall2017the}.
One of the most popular and traditional TSC approaches is the use of a nearest neighbor (NN) classifier coupled with a distance function~\citep{lines2015time}. 
Particularly, the Dynamic Time Warping (DTW) distance when used with a NN classifier has been shown to be a very strong baseline~\citep{bagnall2017the}. 
\cite{lines2015time} compared several distance measures and showed that there is no single distance measure that significantly outperforms DTW.
They also showed that ensembling the individual NN classifiers (with different distance measures) outperforms all of the ensemble's individual components.   
Hence, recent contributions have focused on developing ensembling methods that significantly outperforms the NN coupled with DTW (NN-DTW)~\citep{bagnall2016time,hills2014classification,bostrom2015binary,lines2016hive,schafer2015the,kate2016using,deng2013a,baydogan2013a}. 
These approaches use either an ensemble of decision trees (random forest)~\citep{baydogan2013a,deng2013a} or an ensemble of different types of discriminant classifiers (Support Vector Machine (SVM), NN with several distances) on one or several feature spaces~\citep{bagnall2016time,bostrom2015binary,schafer2015the,kate2016using}. 
Most of these approaches significantly outperform the NN-DTW~\citep{bagnall2017the} and share one common property, which is the data transformation phase where time series are transformed into a new feature space (for example using shapelets transform~\citep{bostrom2015binary} or DTW features~\citep{kate2016using}). 
This notion motivated the development of an ensemble of 35 classifiers named COTE (Collective Of Transformation-based Ensembles)~\citep{bagnall2016time} that does not only ensemble different classifiers over the same transformation, but instead ensembles different classifiers over different time series representations. 
~\cite{lines2016hive,lines2018time} extended COTE with a Hierarchical Vote system to become HIVE-COTE which has been shown to achieve a significant improvement over COTE by leveraging a new hierarchical structure with probabilistic voting, including two new classifiers and two additional representation transformation domains.
HIVE-COTE is currently considered the state-of-the-art algorithm for time series classification~\citep{bagnall2017the} when evaluated over the 85 datasets from the UCR/UEA archive. 

To achieve its high accuracy, HIVE-COTE becomes hugely computationally intensive and impractical to run on a real big data mining problem~\citep{bagnall2017the}. 
The approach requires training 37 classifiers as well as cross-validating each hyperparameter of these algorithms, which makes the approach infeasible to train in some situations~\citep{lucas2018proximity}. 
To emphasize on this infeasibility, note that one of these 37 classifiers is the Shapelet Transform~\citep{hills2014classification} whose time complexity is $O(n^2\cdot l^4)$ with $n$ being the number of time series in the dataset and $l$ being the length of a time series.
Adding to the training time's complexity is the high \emph{classification} time of one of the 37 classifiers: the nearest neighbor which needs to scan the training set before taking a decision at test time.
Therefore since the nearest neighbor constitutes an essential component of HIVE-COTE, its deployment in a real-time setting is still limited if not impractical. 
Finally, adding to the huge runtime of HIVE-COTE, the decision taken by 37 classifiers cannot be interpreted easily by domain experts, since researchers already struggle with understanding the decisions taken by an individual classifier.

After having established the current state-of-the-art of non deep classifiers for TSC~\citep{bagnall2017the}, we discuss the success of Deep Learning~\citep{lecun2015deep} in various classification tasks which motivated the recent utilization of deep learning models for TSC~\citep{wang2017time}. 
Deep Convolutional Neural Networks (CNNs) have revolutionized the field of computer vision~\citep{krizhevsky2012imagenet}. 
For example, in 2015, CNNs were used to reach human level performance in image recognition tasks~\citep{szegedy2015going}. 
Following the success of deep neural networks (DNNs) in computer vision, a huge amount of research proposed several DNN architectures to solve natural language processing (NLP) tasks such as machine translation~\citep{sutskever2014sequence,bahdanau2015neural}, learning word embeddings~\citep{mikolov2013distributed,mikolov2013efficient} and document classification~\citep{le2014distributed,goldberg2016a}. 
DNNs also had a huge impact on the speech recognition community~\citep{hinton2012deep,sainath2013deep}.
Interestingly, we should note that the intrinsic similarity between the NLP and speech recognition tasks is due to the sequential aspect of the data which is also one of the main characteristics of time series data.  

In this context, this paper targets the following open questions: \textit{What is the current state-of-the-art DNN for TSC}?
\textit{Is there a current DNN approach that reaches state-of-the-art performance for TSC and is less complex than HIVE-COTE}? 
\textit{What type of DNN architectures works best for the TSC task}? 
\textit{How does the random initialization affect the performance of deep learning classifiers}?
And finally: \textit{Could the black-box effect of DNNs be avoided to provide interpretability}? 
Given that the latter questions have not been addressed by the TSC community, it is surprising how much recent papers have neglected the possibility that TSC problems could be solved using a pure feature learning algorithm~\citep{neamtu2018generalized,bagnall2017the,lines2016hive}. 
In fact, a recent empirical study~\citep{bagnall2017the} evaluated 18 TSC algorithms on 85 time series datasets, none of which was a deep learning model. 
This shows how much the community lacks of an overview of the current performance of deep learning models for solving the TSC problem~\citep{lines2018time}. 

In this paper, we performed an empirical comparative study of the most recent deep learning approaches for TSC. 
With the rise of graphical processing units (GPUs), we show how deep architectures can be trained efficiently to learn hidden discriminative features from raw time series in an end-to-end manner.
Similarly to~\cite{bagnall2017the}, in order to have a fair comparison between the tested approaches, we developed a common framework in Python, Keras~\citep{chollet2015keras} and Tensorflow~\citep{tensorflow2015whitepaper} to train the deep learning models on a cluster of more than 60 GPUs.

In addition to the univariate datasets' evaluation, we tested the approaches on 12 Multivariate Time Series (MTS) datasets~\citep{baydogan2015mts}.
The multivariate evaluation shows another benefit of deep learning models, which is the ability to handle the curse of dimensionality~\citep{bellman2010dynamic,keogh2017curse} by leveraging different degrees of smoothness in compositional function~\citep{poggio2017why} as well as the parallel computations of the GPUs~\citep{lu2015a}. 

As for comparing the classifiers over multiple datasets, we followed the recommendations in ~\cite{demsar2006statistical} and used the Friedman test~\citep{friedman1940a} to reject the null hypothesis. 
Once we have established that a statistical difference exists within the classifiers' performance, we followed the pairwise post-hoc analysis recommended by~\cite{benavoli2016should} where the average rank comparison is replaced by a Wilcoxon signed-rank test~\citep{wilcoxon1945individual} with Holm's alpha correction~\citep{holm1979a,garcia2008an}. 
See Section~\ref{sec-results} for examples of critical difference diagrams~\citep{demsar2006statistical}, where a thick horizontal line shows a group of classifiers (a clique) that are not significantly different in terms of accuracy.

In this study, we have trained about 1 billion parameters across 97 univariate and multivariate time series datasets. 
Despite the fact that a huge number of parameters risks overfitting~\citep{zhang2017understanding} the relatively small train set in the UCR/UEA archive, our experiments showed that not only DNNs are able to significantly outperform the NN-DTW, but are also able to achieve results that are \emph{not significantly} different than COTE and HIVE-COTE using a deep residual network architecture~\citep{he2016deep,wang2017time}.
Finally, we analyze how poor random initializations can have a significant effect on a DNN's performance.

The rest of the paper is structured as follows. 
In Section~\ref{sec-background}, we provide some background materials concerning the main types of architectures that have been proposed for TSC.
In Section~\ref{sec-approaches}, the tested architectures are individually presented in details.
We describe our experimental open source framework in Section~\ref{sec-experiment}.    
The corresponding results and the discussions are presented in Section~\ref{sec-results}. 
In Section~\ref{sec-visualization}, we describe in detail a couple of methods that mitigate the black-box effect of the deep learning models.  
Finally, we present a conclusion in Section~\ref{sec-conclusion} to summarize our findings and discuss future directions. 

\smallskip

\noindent The main contributions of this paper can be summarized as follows: 
\begin{itemize}
	\item We explain with practical examples, how deep learning can be adapted to one dimensional time series data. 
    \item We propose a unified taxonomy that regroups the recent applications of DNNs for TSC in various domains under two main categories: generative and discriminative models. 
    \item We detail the architecture of nine end-to-end deep learning models designed specifically for TSC. 
    \item We evaluate these models on the univariate UCR/UEA archive benchmark and 12 MTS classification datasets.
    \item We provide the community with an open source deep learning framework for TSC in which we have implemented all nine approaches.
    \item We investigate the use of Class Activation Map (CAM) in order to reduce DNNs' black-box effect and explain the different decisions taken by various models.  
\end{itemize} 

\section{Background}\label{sec-background}
In this section, we start by introducing the necessary definitions for ease of understanding. 
We then follow by an extensive theoretical background on training DNNs for the TSC task. 
Finally we present our proposed taxonomy of the different DNNs with examples of their application in various real world data mining problems.   

\subsection{Time series classification}
Before introducing the different types of neural networks architectures, we go through some formal definitions for TSC. 
\begin{definition}
A univariate time series $X=[x_1,x_2, \dots, x_T]$ is an ordered set of real values. 
The length of $X$ is equal to the number of real values $T$. 
\end{definition}

\begin{definition}
An $M$-dimensional MTS, $X=[X^1,X^2, \dots, X^M]$ consists of $M$ different univariate time series with $X^i$ $\in \mathbb{R}^T$. 
\end{definition}

\begin{definition}
A dataset $D=\{(X_1,Y_1), (X_2,Y_2), \dots, (X_N,Y_N)\}$ is a collection of pairs $(X_i,Y_i)$ where $X_i$ could either be a univariate or multivariate time series with $Y_i$ as its corresponding one-hot label vector.
For a dataset containing $K$ classes, the one-hot label vector $Y_i$ is a vector of length $K$ where each element $j\in[1,K]$ is equal to $1$ if the class of $X_i$ is $j$ and $0$ otherwise.
\end{definition}

The task of TSC consists of training a classifier on a dataset $D$ in order to map from the space of possible inputs to a probability distribution over the class variable values (labels).

\begin{figure}
	\centering
    \includegraphics[width=0.9\linewidth]{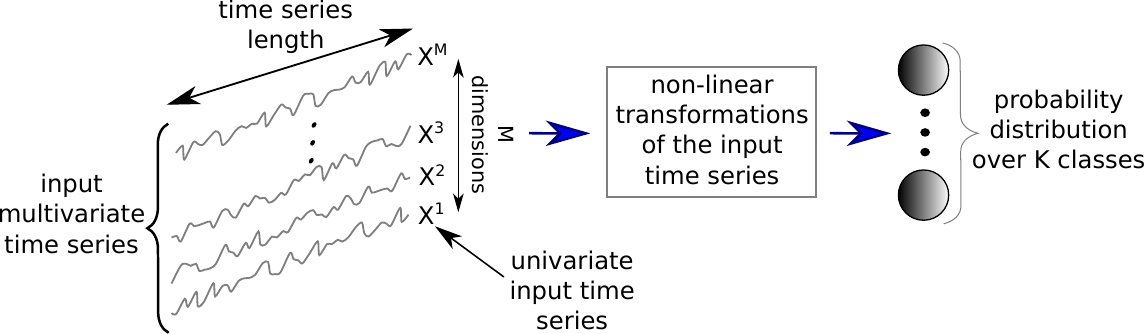}
    \caption{A unified deep learning framework for time series classification.}
    \label{fig-unified-framework}
\end{figure}

\subsection{Deep learning for time series classification}
In this review, we focus on the TSC task~\citep{bagnall2017the} using DNNs which are considered complex machine learning models~\citep{lecun2015deep}.
A general deep learning framework for TSC is depicted in Figure~\ref{fig-unified-framework}. 
These networks are designed to learn hierarchical representations of the data. 
A deep neural network is a composition of $L$ parametric functions referred to as layers 
where each layer is considered a representation of the input domain~\citep{papernot2018deep}.
One layer $l_i$, such as $i \in 1 \dots L$, contains neurons, which are small units that compute one element of the layer's output. 
The layer $l_i$ takes as input the output of its previous layer $l_{i-1}$ and applies a non-linearity (such as the sigmoid function) to compute its own output. 
The behavior of these non-linear transformations is controlled by a set of parameters $\theta_i$ for each layer. 
In the context of DNNs, these parameters are called weights which link the input of the previous layer to the output of the current layer.
Hence, given an input $x$, a neural network performs the following computations to predict the class:
\begin{equation}\label{eq-1}
f_L(\theta_L,x) = f_{L-1}(\theta_{L-1},f_{L-2}(\theta_{L-2}, \dots ,f_1(\theta_1,x)))
\end{equation}
where $f_i$ corresponds to the non-linearity applied at layer $l_i$. 
For simplicity, we will omit the vector of parameters $\theta$ and use $f(x)$ instead of $f(\theta,x)$. 
This process is also referred to as \emph{feed-forward} propagation in the deep learning literature. 

During training, the network is presented with a certain number of known input-output (for example a dataset $D$).
First, the weights are initialized randomly~\citep{lecun1998efficient}, although a robust alternative would be to take a pre-trained model on a source dataset and fine-tune it on the target dataset~\citep{pan2010a}. 
This process is known as transfer learning which we do not study empirically, rather we discuss the transferability of each model with respect to the architecture in Section~\ref{sec-approaches}. 
After the weight's initialization, a forward pass through the model is applied: using the function $f$ the output of an input $x$ is computed. 
The output is a vector whose components are the estimated probabilities of $x$ belonging to each class. 
The model's prediction loss is computed using a cost function, for example the negative log likelihood. 
Then, using gradient descent~\citep{lecun1998efficient}, the weights are updated in a backward pass to propagate the error. 
Thus, by iteratively taking a forward pass followed by backpropagation, the model's parameters are updated in a way that minimizes the loss on the training data. 

During testing, the probabilistic classifier (the model) is tested on unseen data which is also referred to as the inference phase: a forward pass on this unseen input followed by a class prediction.
The prediction corresponds to the class whose probability is maximum. 
To measure the performance of the model on the test data (generalization), we adopted the accuracy measure (similarly to~\cite{bagnall2017the}).
One advantage of DNNs over non-probabilistic classifiers (such as NN-DTW) is that a probabilistic decision is taken by the network~\citep{large2017the}, thus allowing to measure the confidence of a certain prediction given by an algorithm.

Although there exist many types of DNNs, in this review we focus on three main DNN architectures used for the TSC task: Multi Layer Perceptron (MLP), Convolutional Neural Network (CNN) and Echo State Network (ESN).  
These three types of architectures were chosen since they are widely adopted for end-to-end deep learning~\citep{lecun2015deep} models for TSC. 

\subsubsection{Multi Layer Perceptrons}
An MLP constitutes the simplest and most traditional architecture for deep learning models. 
This form of architecture is also known as a fully-connected (FC) network since the neurons in layer $l_i$ are  connected to every neuron in layer $l_{i-1}$ with $i\in [1,L]$.
These connections are modeled by the weights in a neural network.
A general form of applying a non-linearity to an input time series $X$ can be seen in the following equation: 
\begin{equation}\label{eq-activation}
A_{l_i}=f(\omega_{l_i}*X+b)
\end{equation}
with $\omega_{l_i}$ being the set of weights with length and number of dimensions identical to $X$'s, $b$ the bias term and $A_{l_i}$ the activation of the neurons in layer $l_i$. 
Note that the number of neurons in a layer is considered a hyperparameter. 

One impediment from adopting MLPs for time series data is that they do not exhibit any spatial invariance. 
In other words, each time stamp has its own weight and the temporal information is lost: meaning time series elements are treated independently from each other.
For example the set of weights $w_d$ of neuron $d$ contains $T\times M$ values denoting the weight of each time stamp $t$ for each dimension of the $M$-dimensional input MTS of length $T$. 
Then by cascading the layers we obtain a computation graph similar to equation~\ref{eq-1}. 

For TSC, the final layer is usually a discriminative layer that takes as input the activation of the previous layer and gives a probability distribution over the class variables in the dataset.
Most deep learning approaches for TSC employ a softmax layer which corresponds to an FC layer with softmax as activation function $f$ and a number of neurons equal to the number of classes in the dataset. 
Three main useful properties motivate the use of the softmax activation function: the sum of probabilities is guaranteed to be equal to $1$, the function is differentiable and it is an adaptation of logistic regression to the multinomial case. 
The result of a softmax function can be defined as follows: 
\begin{equation}\label{eq-softmax}
\hat{Y}_j(X) = \frac{e^{A_{L-1}*\omega_j+b_j}}{\sum_{k=1}^{K}{e^{A_{L-1}*\omega_k+b_k}}}
\end{equation}
with $\hat{Y}_j$ denoting the probability of $X$ having the class $Y$ equal to class $j$ out of $K$ classes in the dataset.
The set of weights $w_j$ (and the corresponding bias $b_j$) for each class $j$ are linked to each previous activation in layer $l_{L-1}$. 

The weights in equations~(\ref{eq-activation}) and~(\ref{eq-softmax}) should be learned automatically using an optimization algorithm that minimizes an objective cost function. 
In order to approximate the error of a certain given value of the weights, a differentiable cost (or loss) function that quantifies this error should be defined. 
The most used loss function in DNNs for the classification task is the categorical cross entropy as defined in the following equation: 
\begin{equation}
L(X)=-\sum_{j=1}^{K}{Y_j\log{\hat{Y}_j}}
\end{equation}
with $L$ denoting the loss or cost when classifying the input time series $X$. 
Similarly, the average loss when classifying the whole training set of $D$ can be defined using the following equation: 
\begin{equation}
J(\Omega)=\frac{1}{N}\sum_{n=1}^{N}{L(X_n)}
\end{equation}
with $\Omega$ denoting the set of weights to be learned by the network (in this case the weights $w$ from equations~\ref{eq-activation} and~\ref{eq-softmax}).
The loss function is minimized to learn the weights in $\Omega$ using a gradient descent method which is defined using the following equation: 
\begin{equation}
\omega = \omega - \alpha \frac{\partial J}{\partial \omega} ~ | ~ \forall ~ \omega \in \Omega
\end{equation}
with $\alpha$ denoting the learning rate of the optimization algorithm. 
By subtracting the partial derivative, the model is actually auto-tuning the parameters $\omega$ in order to reach a local minimum of $J$ in case of a non-linear classifier (which is almost always the case for a DNN). 
We should note that when the partial derivative cannot be directly computed with respect to a certain parameter $\omega$, the chain rule of derivative is employed which is in fact the main idea behind the backpropagation algorithm~\citep{lecun1998efficient}. 

\subsubsection{Convolutional Neural Networks}
Since AlexNet~\citep{krizhevsky2012imagenet} won the ImageNet competition in 2012, deep CNNs have seen a lot of successful applications in many different domains~\citep{lecun2015deep} such as reaching human level performance in image recognition problems~\citep{szegedy2015going} as well as different natural language processing tasks~\citep{sutskever2014sequence,bahdanau2015neural}. 
Motivated by the success of these CNN architectures in these various domains, researchers have started adopting them for time series analysis~\citep{gamboa2017deep}.

\begin{figure}
\centering
\includegraphics[width=.5\linewidth]{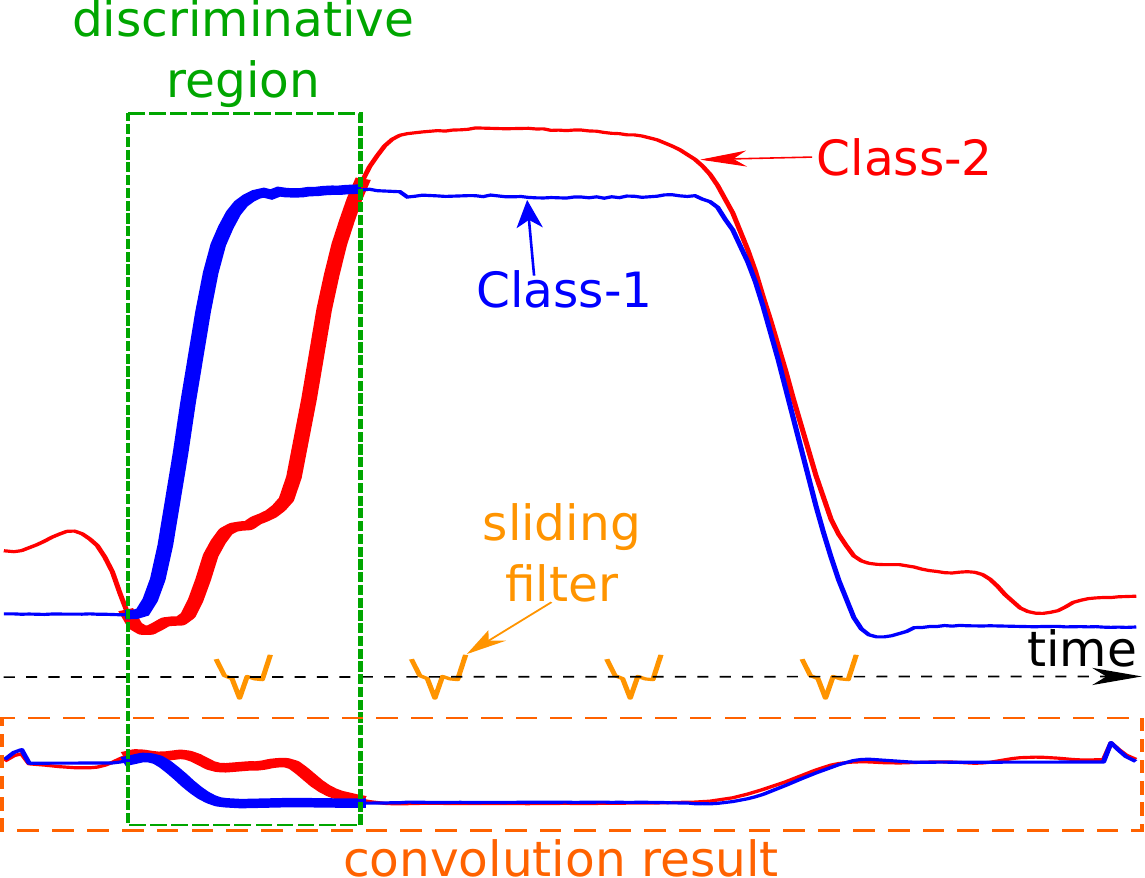}
\caption{The result of a applying a learned discriminative convolution on the GunPoint dataset}
\label{fig-conv-example}
\end{figure}

A convolution can be seen as applying and sliding a filter over the time series.
Unlike images, the filters exhibit only one dimension (time) instead of two dimensions (width and height). 
The filter can also be seen as a generic non-linear transformation of a time series.
Concretely, if we are convoluting (multiplying) a filter of length 3 with a univariate time series, by setting the filter values to be equal to $[\frac{1}{3},\frac{1}{3},\frac{1}{3}]$, the convolution will result in applying a moving average with a sliding window of length $3$.
A general form of applying the convolution for a centered time stamp $t$ is given in the following equation: 
\begin{equation}
C_t = f(\omega*X_{t-l/2:t+l/2}+b)   ~  | ~ \forall ~ t \in [1,T]
\end{equation}
where $C$ denotes the result of a convolution (dot product $*$) applied on a univariate time series $X$ of length $T$ with a filter $\omega$ of length $l$, a bias parameter $b$ and a final non-linear function $f$ such as the Rectified Linear Unit (ReLU). 
The result of a convolution (one filter) on an input time series $X$ can be considered as another univariate time series $C$ that underwent a filtering process.
Thus, applying several filters on a time series will result in a multivariate time series whose dimensions are equal to the number of filters used.  
An intuition behind applying several filters on an input time series would be to learn multiple discriminative features useful for the classification task. 

Unlike MLPs, the same convolution (the same filter values $w$ and $b$) will be used to find the result for all time stamps $t\in [1,T]$. 
This is a very powerful property (called weight sharing) of the CNNs which enables them to learn filters that are invariant across the time dimension.   

When considering an MTS as input to a convolutional layer, the filter no longer has one dimension (time) but also has dimensions that are equal to the number of dimensions of the input MTS.  

Finally, instead of setting manually the values of the filter $\omega$, these values should be learned automatically since they depend highly on the targeted dataset. 
For example, one dataset would have the optimal filter to be equal to $[1,2,2]$ whereas another dataset would have an optimal filter equal to $[2,0,-1]$. 
By \emph{optimal} we mean a filter whose application will enable the classifier to easily discriminate between the dataset classes (see Figure~\ref{fig-conv-example}).
In order to learn automatically a discriminative filter, the convolution should be followed by a discriminative classifier, which is usually preceded by a \emph{pooling} operation that can either be \emph{local} or \emph{global}. 

Local pooling such as \emph{average} or \emph{max} pooling takes an input time series and reduces its length $T$ by aggregating over a sliding window of the time series.
For example if the sliding window's length is equal to $3$ the resulting pooled time series will have a length equal to $\frac{T}{3}$ - this is only true if the stride is equal to the sliding window's length.
With a global pooling operation, the time series will be aggregated over the whole time dimension resulting in a single real value. 
In other words, this is similar to applying a local pooling with a sliding window's length equal to the length of the input time series. 
Usually a global aggregation is adopted to reduce drastically the number of parameters in a model thus decreasing the risk of overfitting while enabling the use of CAM to explain the model's decision~\citep{zhou2016learning}. 

In addition to pooling layers, some deep learning architectures include normalization layers to help the network converge quickly. 
For time series data, the batch normalization operation is performed over each channel therefore preventing the internal covariate shift across one mini-batch training of time series~\citep{ioffe2015batch}.
Another type of normalization was proposed by~\cite{ulyanov2016instance} to normalize each instance instead of a per batch basis, thus learning the mean and standard deviation of each training instance for each layer via gradient descent. 
The latter approach is called instance normalization and mimics learning the z-normalization parameters for the time series training data.

\begin{figure}[t]
\centering
\includegraphics[width=1.0\linewidth]{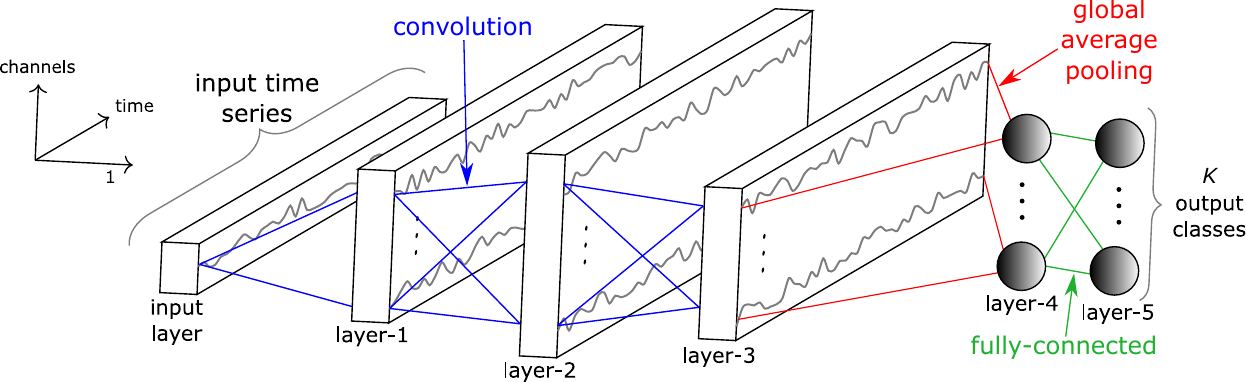}
\caption{Fully Convolutional Neural Network architecture}
\label{fig-fcn-archi}
\end{figure}

The final discriminative layer takes the representation of the input time series (the result of the convolutions) and give a probability distribution over the class variables in the dataset.  
Usually, this layer is comprised of a softmax operation similarly to the MLPs. 
Note that for some approaches, we would have an additional non-linear FC layer before the final softmax layer which increases the number of parameters in a network. 
Finally in order to train and learn the parameters of a deep CNN, the process is identical to training an MLP: a feed-forward pass followed by backpropagation~\citep{lecun1998efficient}. 
An example of a CNN architecture for TSC with three convolutional layers is illustrated in Figure~\ref{fig-fcn-archi}.

\subsubsection{Echo State Networks}
Another popular type of architectures for deep learning models is the Recurrent Neural Network (RNN). 
Apart from time series forecasting, we found that these neural networks were rarely applied for time series classification which is mainly due to three factors: (1) the type of this architecture is designed mainly to predict an output for each element (time stamp) in the time series~\citep{langkvist2014a}; (2) RNNs typically suffer from the vanishing gradient problem due to training on long time series~\citep{pascanu2012understanding}; (3) RNNs are considered hard to train and parallelize which led the researchers to avoid using them for computational reasons~\citep{pascanu2013on}.

Given the aforementioned limitations, a relatively recent type of recurrent architecture was proposed for time series: Echo State Networks (ESNs)~\citep{gallicchio2017deep}. 
ESNs were first invented by~\cite{jaeger2004Harnessing} for time series prediction in wireless communication channels.
They were designed to mitigate the challenges of RNNs by eliminating the need to compute the gradient for the hidden layers which reduces the training time of these neural networks thus avoiding the vanishing gradient problem.   
These hidden layers are initialized randomly and constitutes the \emph{reservoir}: the core of an ESN which is a sparsely connected random RNN. 
Each neuron in the reservoir will create its own nonlinear activation of the incoming signal. 
The inter-connected weights inside the reservoir and the input weights are not learned via gradient descent, only the output weights are tuned using a learning algorithm such as logistic regression or Ridge classifier~\citep{hoerl1970ridge}. 

\begin{figure}
\centering
\includegraphics[width=0.8\linewidth]{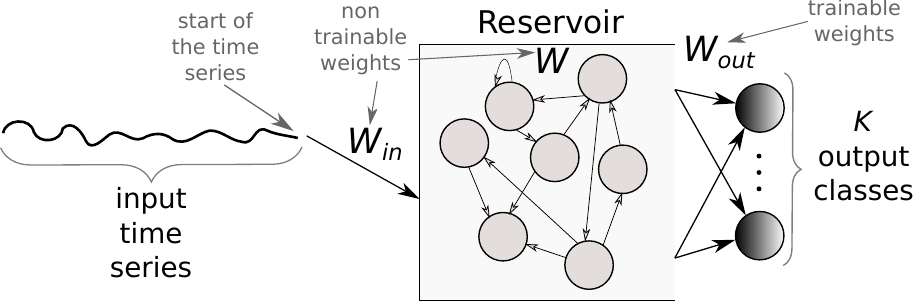}
\caption{An Echo State Network architecture for time series classification}
\label{fig-esn-archi}
\end{figure}

To better understand the mechanism of these networks, consider an ESN with input dimensionality $M$, neurons in the reservoir $N_r$ and an output dimensionality $K$ equal to the number of classes in the dataset. 
Let $X(t) \in \mathbb{R}^M$, $I(t) \in \mathbb{R}^{N_r}$ and $\hat{Y}(t) \in \mathbb{R}^K$ denote the vectors of the input $M$-dimensional MTS, the internal (or hidden) state and the output unit activity for time $t$ respectively. 
Further let $W_{in} \in \mathbb{R}^{N_r\times M}$ and $W \in \mathbb{R}^{N_r\times N_r}$ and $W_{out} \in \mathbb{R}^{C\times N_r}$ denote respectively the weight matrices for the input time series, the internal connections and the output connections as seen in Figure~\ref{fig-esn-archi}.    
The internal unit activity $I(t)$ at time $t$ is updated using the internal state at time step $t-1$ and the input time series element at time $t$. 
Formally the hidden state can be computed using the following recurrence: 
\begin{equation}
I(t)=f(W_{in}X(t)+W I(t-1))   ~  | ~ \forall ~ t \in [1,T]
\end{equation}
with f denoting an activation function of the neurons, a common choice is $tanh(\cdot)$ applied element-wise~\citep{tanisaro2016time}.
The output can be computed according to the following equation: 
\begin{equation}
\hat{Y}(t)=W_{out}I(t)
\end{equation}
thus classifying each time series element $X(t)$.  
Note that ESNs depend highly on the initial values of the reservoir that should satisfy a pre-determined hyperparameter: the spectral radius. 
Figure~\ref{fig-esn-archi} shows an example of an ESN with a univariate input time series to be classified into $K$ classes.

Finally, we should note that for all types of DNNs, a set of techniques was proposed by the deep learning community to enhance neural networks' generalization capabilities. 
Regularization methods such as $l2$-norm weight decay~\citep{bishop2006pattern} or Dropout~\citep{srivastava2014a} aim at reducing overfitting by limiting the activation of the neurons. 
Another popular technique is data augmentation, which tackles the problem of overfitting a small dataset by increasing the number of training instances~\citep{baird1992document}. 
This method consists in cropping, rotating and blurring images which have been shown to improve the DNNs' performance for computer vision tasks~\citep{zhang2017understanding}. 
Although two approaches in this survey include a data augmentation technique, the study of its impact on TSC is currently limited~\citep{ismailFawaz2018data}. 

\subsection{Generative or discriminative approaches}
\begin{figure}
\centering
\includegraphics[width=1.0\linewidth]{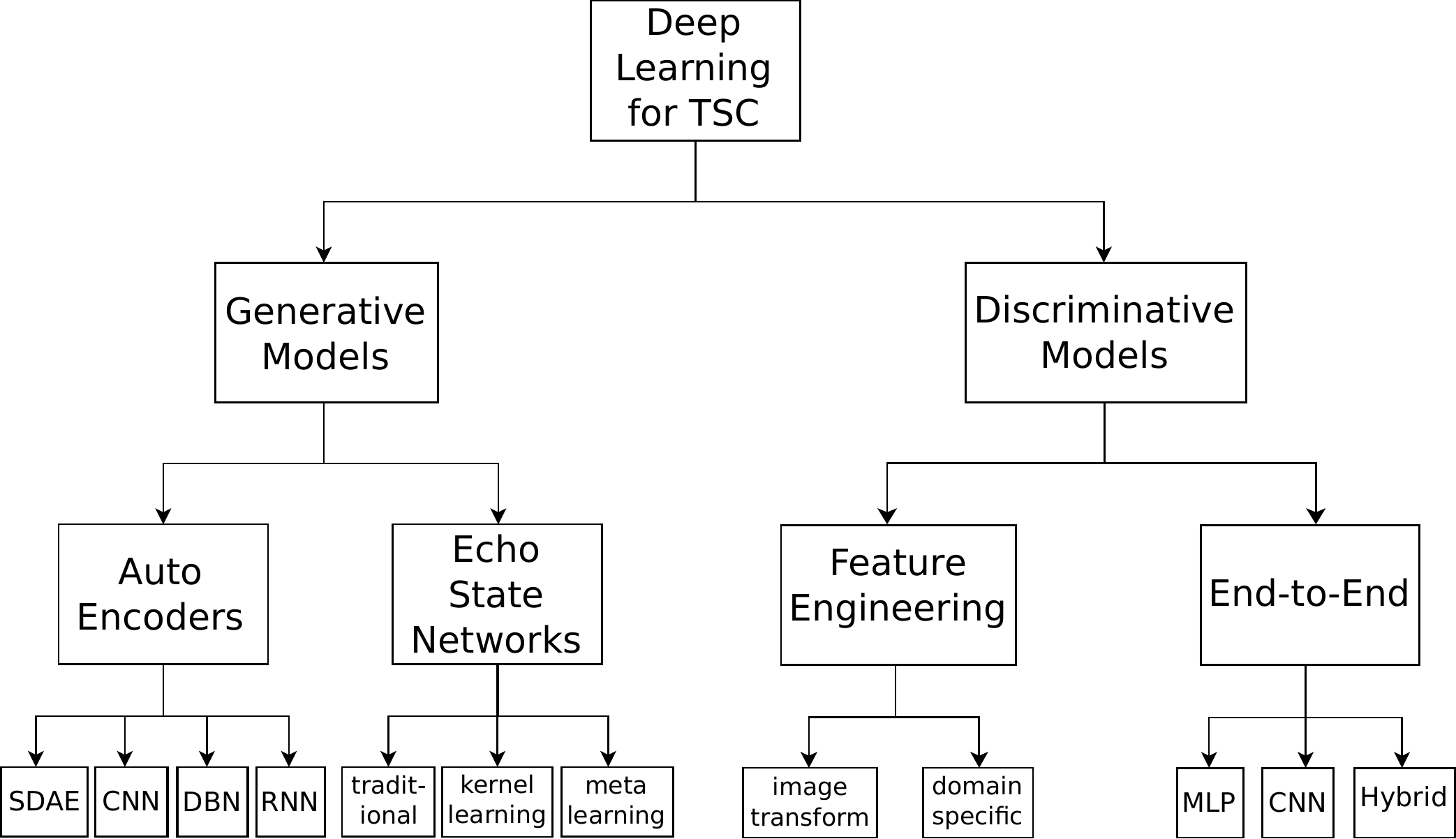}
\caption{An overview of the different deep learning approaches for time series classification}
\label{fig-diagram}
\end{figure}

Deep learning approaches for TSC can be separated into two main categories: the \emph{generative} and the \emph{discriminative} models (as proposed in \cite{langkvist2014a}).
We further separate these two groups into sub-groups which are detailed in the following subsections and illustrated in Figure~\ref{fig-diagram}. 

\subsubsection{Generative models}
Generative models usually exhibit an unsupervised training step that precedes the learning phase of the classifier~\citep{langkvist2014a}.
This type of network has been referred to as \emph{Model-based} classifiers in the TSC community~\citep{bagnall2017the}. 
Some of these generative non deep learning approaches include auto-regressive models~\citep{bagnall2014a}, hidden Markov models~\citep{kotsifakos2014model} and kernel models~\citep{chen2013model}.  

For all generative approaches, the goal is to find a good representation of time series prior to training a classifier~\citep{langkvist2014a}. 
Usually, to model the time series, classifiers are preceded by an unsupervised pre-training phase such as stacked denoising auto-encoders (SDAEs)~\citep{bengio2013generalized,hu2016transfer}. 
A generative CNN-based model was proposed in~\cite{wang2016representation,mittelman2015time} where the authors introduced a deconvolutional operation followed by an upsampling technique that helps in reconstructing a multivariate time series. 
Deep Belief Networks (DBNs) were also used to model the latent features in an unsupervised manner which are then leveraged to classify univariate and multivariate time series~\citep{wang2017a,banerjee2017a}.
In~\cite{mehdiyev2017time,malhotra2017timenet,rajan2018a}, an RNN auto-encoder was designed to first generate the time series then using the learned latent representation, they trained a classifier (such as SVM or Random Forest) on top of these representations to predict the class of a given input time series.

Other studies such as in~\cite{aswolinskiy2017time,bianchi2018reservoir,chouikhi2018genesis,ma2016functional} used self-predict modeling for time series classification where ESNs were first used to re-construct the time series and then the learned representation in the reservoir space was utilized for classification.  
We refer to this type of architecture by traditional ESNs in Figure~\ref{fig-diagram}. 
Other ESN-based approaches~\citep{chen2015model,chen2013model,che2017decade} define a kernel over the learned representation followed by an SVM or an MLP classifier.  
In~\cite{gong2018Multiobjective,wang2016an}, a meta-learning evolutionary-based algorithm was proposed to construct an optimal ESN architecture for univariate and multivariate time series. 
For more details concerning generative ESN models for TSC, we refer the interested reader to a recent empirical study~\citep{aswolinskiy2016time} that compared classification in reservoir and model-space for both multivariate and univariate time series.   

\subsubsection{Discriminative models}
A discriminative deep learning model is a classifier (or regressor) that directly learns the mapping between the raw input of a time series (or its hand engineered features) and outputs a probability distribution over the class variables in a dataset. 
Several discriminative deep learning architectures have been proposed to solve the TSC task, but we found that this type of model could be further sub-divided into two groups: (1) deep learning models with hand engineered features and (2) \emph{end-to-end} deep learning models. 

The most frequently encountered and computer vision inspired feature extraction method for hand engineering approaches is the transformation of time series into images using specific imaging methods such as Gramian fields~\citep{wang2015imaging,wang2015Encoding}, recurrence plots~\citep{hatami2017classification,tripathy2018use} and Markov transition fields~\citep{wang2015spatially}. 
Unlike image transformation, other feature extraction methods are not domain agnostic. 
These features are first hand-engineered using some domain knowledge, then fed to a deep learning discriminative classifier. 
For example in~\cite{uemura2018Feasibility}, several features (such as the velocity) were extracted from sensor data placed on a surgeon's hand in order to determine the skill level during surgical training. 
In fact, most of the deep learning approaches for TSC with some hand engineered features are present in human activity recognition tasks~\citep{ignatov2018activity}.
For more details on the different applications of deep learning for human motion detection using mobile and wearable sensor networks, we refer the interested reader to a recent survey~\citep{nweke2018deep} where deep learning approaches (with or without hand engineered features) were thoroughly described specifically for the human activity recognition task.

In contrast to feature engineering, \emph{end-to-end} deep learning aims to incorporate the feature learning process while fine-tuning the discriminative classifier~\citep{nweke2018deep}. 
Since this type of deep learning approach is domain agnostic and does not include any domain specific pre-processing steps, we decided to further separate these end-to-end approaches using their neural network architectures. 

In~\cite{wang2017time,geng2018cost}, an MLP was designed to learn from scratch a discriminative time series classifier.
The problem with an MLP approach is that temporal information is lost and the features learned are no longer time-invariant.
This is where CNNs are most useful, by learning spatially invariant filters (or features) from raw input time series~\citep{wang2017time}. 
During our study, we found that CNN is the most widely applied architecture for the TSC problem, which is probably due to their robustness and the relatively small amount of training time compared to complex architectures such as RNNs or MLPs. 
Several variants of CNNs have been proposed and validated on a subset of the UCR/UEA archive~\citep{ucrarchive,bagnall2017the} such as Residual Networks (ResNets)~\citep{wang2017time,geng2018cost} which add linear shortcut connections for the convolutional layers potentially enhancing the model's accuracy~\citep{he2016deep}. 
In~\cite{leguennec2016data,cui2016multi,wang2017time,zhao2017convolutional}, traditional CNNs were also validated on the UCR/UEA archive.
More recently in~\cite{wang2018multilevel}, the architectures proposed in~\cite{wang2017time} were modified to leverage a filter initialization technique based on the Daubechies 4 Wavelet values~\citep{alistair1995daubechies}. 
Outside of the UCR/UEA archive, deep learning has reached state-of-the-art performance on several datasets in different domains~\citep{langkvist2014a}. 
For spatio-temporal series forecasting problems, such as meteorology and oceanography, DNNs were proposed in~\cite{ziat2017spatio}.
~\cite{strodthoff2018detecting} proposed to detect myocardial infractions from electrocardiography data using deep CNNs.  
For human activity recognition from wearable sensors, deep learning is replacing the feature engineering approaches~\citep{nweke2018deep} where features are no longer hand-designed but rather learned by deep learning models trained through backpropagation.
One other type of time series data is present in Electronic Health Records, where a recent generative adversarial network with a CNN~\citep{che2017boosting} was trained for risk prediction based on patients historical medical records.
In~\cite{ismailFawaz2018evaluating}, CNNs were designed to reach state-of-the-art performance for surgical skills identification. 
~\cite{liu2018time} leveraged a CNN model for multivariate and lag-feature characteristics in order to achieve state-of-the-art accuracy on the Prognostics and Health Management (PHM) 2015 challenge data.
Finally, a recent review of deep learning for physiological signals classification revealed that CNNs were the most popular architecture~\citep{faust2018deep} for the considered task.  
We mention one final type of hybrid architectures that showed promising results for the TSC task on the UCR/UEA archive datasets, where mainly CNNs were combined with other types of architectures such as Gated Recurrent Units~\citep{lin2018gcrnn} and the attention mechanism~\citep{serra2018towards}.  
The reader may have noticed that CNNs appear under Auto Encoders as well as under End-to-End learning in Figure~\ref{fig-diagram}.
This can be explained by the fact that CNNs when trained as Auto Encoders have a complete different objective function than CNNs that are trained in an end-to-end fashion. 

Now that we have presented the taxonomy for grouping DNNs for TSC, we introduce in the following section the different approaches that we have included in our experimental evaluation. 
We also explain the motivations behind the selection of these algorithms.   

\section{Approaches}\label{sec-approaches}
In this section, we start by explaining the reasons behind choosing discriminative end-to-end approaches for this empirical evaluation.
We then describe in detail the nine different deep learning architectures with their corresponding advantages and drawbacks.

\subsection{Why discriminative end-to-end approaches ?}
As previously mentioned in Section~\ref{sec-background}, the main characteristic of a generative model is fitting a time series self-predictor whose latent representation is later fed into an off-the-shelf classifier such as Random Forest or SVM. 
Although these models do sometimes capture the trend of a time series, we decided to leave these generative approaches out of our experimental evaluation for the following reasons:   
\begin{itemize}
	\item This type of method is mainly proposed for tasks other than classification or as part of a larger classification scheme~\citep{bagnall2017the};
    \item The informal consensus in the literature is that generative models are usually less accurate than direct discriminative models~\citep{bagnall2017the,nguyen2017time};
    \item The implementation of these models is usually more complicated than for discriminative models since it introduces an additional step of fitting a time series generator - this has been considered a barrier with most approaches whose code was not publicly available such as~\cite{gong2018Multiobjective,che2017decade,chouikhi2018genesis,wang2017a};
    \item The accuracy of these models depends highly on the chosen off-the-shelf classifier which is sometimes not even a neural network classifier~\citep{rajan2018a}. 
\end{itemize}

Given the aforementioned limitations for generative models, we decided to limit our experimental evaluation to discriminative deep learning models for TSC. 
In addition of restricting the study to discriminative models, we decided to only consider end-to-end approaches, thus further leaving classifiers that incorporate feature engineering out of our empirical evaluation.
We made this choice because we believe that the main goal of deep learning approaches is to remove the bias due to manually designed features~\citep{ordonez2016deep}, thus enabling the network to learn the most discriminant useful features for the classification task.
This has also been the consensus in the human activity recognition literature, where the accuracy of deep learning methods depends highly on the quality of the extracted features~\citep{nweke2018deep}.  
Finally, since our goal is to provide an empirical study of domain agnostic deep learning approaches for any TSC task, we found that it is best to compare models that do not incorporate any domain knowledge into their approach.

As for why we chose the nine approaches (described in the next Section), it is first because among all the discriminative end-to-end deep learning models for TSC, we wanted to cover a wide range of architectures such as CNNs, Fully CNNs, MLPs, ResNets, ESNs, etc.
Second, since we cannot cover an empirical study of all approaches validated in all TSC domains, we decided to only include approaches that were validated on the whole (or a subset of) the univariate time series UCR/UEA archive~\citep{ucrarchive,bagnall2017the} and/or on the MTS archive~\citep{baydogan2015mts}.    
Finally, we chose to work with approaches that do not try to solve a sub task of the TSC problem such as in~\cite{geng2018cost} where CNNs were modified to classify imbalanced time series datasets. 
To justify this choice, we emphasize that imbalanced TSC problems can be solved using several techniques such as data augmentation~\citep{ismailFawaz2018data} and modifying the class weights~\citep{geng2018cost}.
However, any deep learning algorithm can benefit from this type of modification. 
Therefore if we did include modifications for solving imbalanced TSC tasks, it would be much harder to determine if it is the choice of the deep learning classifier or the modification itself that improved the accuracy of the model.
Another sub task that has been at the center of recent studies is early time series classification~\citep{wang2016earliness} where deep CNNs were modified to include an early classification of time series. 
More recently, a deep reinforcement learning approach was also proposed for the early TSC task~\citep{martinez2018a}. 
For further details, we refer the interested reader to a recent survey on deep learning for \emph{early} time series classification~\citep{santos2017a}.  

\subsection{Compared approaches}
After having presented an overview over the recent deep learning approaches for time series classification, we present the nine architectures that we have chosen to compare in this paper.

\subsubsection{Multi Layer Perceptron}
The MLP, which is the most traditional form of DNNs, was proposed in~\cite{wang2017time} as a baseline architecture for TSC.
The network contains 4 layers in total where each one is fully connected to the output of its previous layer. 
The final layer is a softmax classifier, which is fully connected to its previous layer's output and contains a number of neurons equal to the number of classes in a dataset. 
All three hidden FC layers are composed of 500 neurons with ReLU as the activation function. 
Each layer is preceded by a dropout operation~\citep{srivastava2014a} with a rate equal to 0.1, 0.2, 0.2 and 0.3 for respectively the first, second, third and fourth layer. 
Dropout is one form of regularization that helps in preventing overfitting~\citep{srivastava2014a}. 
The dropout rate indicates the percentage of neurons that are deactivated (set to zero) in a feed forward pass during training.

MLP does not have any layer whose number of parameters is invariant across time series of different lengths (denoted by $\#invar$ in \tablename~\ref{tab-archi}) which means that the transferability of the network is not trivial: the number of parameters (weights) of the network depends directly on the length of the input time series.

\subsubsection{Fully Convolutional Neural Network}
Fully Convolutional Neural Networks (FCNs) were first proposed in~\cite{wang2017time} for classifying univariate time series and validated on 44 datasets from the UCR/UEA archive. 
FCNs are mainly convolutional networks that do not contain any local pooling layers which means that the length of a time series is kept unchanged throughout the convolutions.
In addition, one of the main characteristics of this architecture is the replacement of the traditional final FC layer with a Global Average Pooling (GAP) layer which reduces drastically the number of parameters in a neural network while enabling the use of the CAM~\citep{zhou2016learning} that highlights which parts of the input time series contributed the most to a certain classification. 

The architecture proposed in~\cite{wang2017time} is first composed of three convolutional blocks where each block contains three operations: a convolution followed by a batch normalization~\citep{ioffe2015batch} whose result is fed to a ReLU activation function. 
The result of the third convolutional block is averaged over the whole time dimension which corresponds to the GAP layer. 
Finally, a traditional softmax classifier is fully connected to the GAP layer's output. 

All convolutions have a stride equal to 1 with a zero padding to preserve the exact length of the time series after the convolution. 
The first convolution contains 128 filters with a filter length equal to 8, followed by a second convolution of 256 filters with a filter length equal to 5 which in turn is fed to a third and final convolutional layer composed of 128 filters, each one with a length equal to 3. 

We can see that FCN does not hold any pooling nor a regularization operation.
In addition, one of the advantages of FCNs is the invariance (denoted by $\#invar$ in \tablename~\ref{tab-archi}) in the number of parameters for 4 layers (out of 5) across time series of different lengths. 
This invariance (due to using GAP) enables the use of a transfer learning approach where one can train a model on a certain source dataset and then fine-tune it on the target dataset~\citep{IsmailFawaz2018transfer}. 

\subsubsection{Residual Network}
The third and final proposed architecture in~\cite{wang2017time} is a relatively deep Residual Network (ResNet). 
For TSC, this is the deepest architecture with 11 layers of which the first 9 layers are convolutional followed by a GAP layer that averages the time series across the time dimension. 
The main characteristic of ResNets is the shortcut residual connection between consecutive convolutional layers.
Actually, the difference with the usual convolutions (such as in FCNs) is that a linear shortcut is added to link the output of a residual block to its input thus enabling the flow of the gradient directly through these connections, which makes training a DNN much easier by reducing the vanishing gradient effect~\citep{he2016deep}. 

The network is composed of three residual blocks followed by a GAP layer and a final softmax classifier whose number of neurons is equal to the number of classes in a dataset. 
Each residual block is first composed of three convolutions whose output is added to the residual block's input and then fed to the next layer. 
The number of filters for all convolutions is fixed to 64, with the ReLU activation function that is preceded by a batch normalization operation. 
In each residual block, the filter's length is set to 8, 5 and 3 respectively for the first, second and third convolution.  

\begin{figure}
\centering
\includegraphics[width=1.0\linewidth]{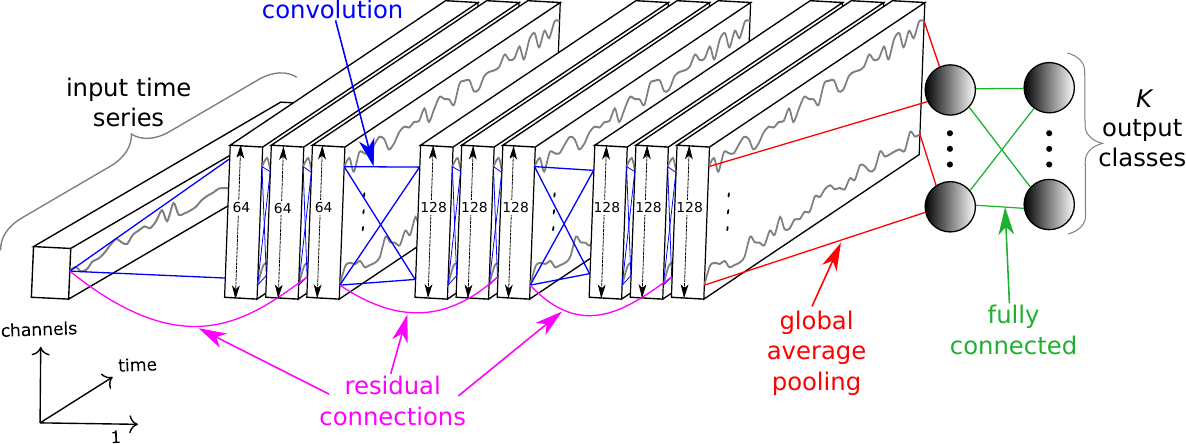}
\caption{The Residual Network's architecture for time series classification.}
\label{fig-resnet-archi}
\end{figure}

Similarly to the FCN model, the layers (except the final one) in the ResNet architecture have an invariant number of parameters across different datasets. 
That being said, we can easily pre-train a model on a source dataset, then transfer and fine-tune it on a target dataset without having to modify the hidden layers of the network. 
As we have previously mentioned and since this type of transfer learning approach can give an advantage for certain types of architecture, we leave the exploration of this area of research for future work. 
The ResNet architecture proposed by~\cite{wang2017time} is depicted in Figure~\ref{fig-resnet-archi}.

\subsubsection{Encoder}
Originally proposed by~\cite{serra2018towards}, Encoder is a hybrid deep CNN whose architecture is inspired by FCN~\citep{wang2017time} with a main difference where the GAP layer is replaced with an attention layer. 
In~\cite{serra2018towards}, two variants of Encoder were proposed: the first approach was to train the model from scratch in an end-to-end fashion on a target dataset while the second one was to pre-train this same architecture on a source dataset and then fine-tune it on a target dataset. 
The latter approach reached higher accuracy thus benefiting from the transfer learning technique. 
On the other hand, since almost all approaches can benefit to certain degree from a transfer learning method, we decided to implement only the end-to-end approach (training from scratch) which already showed high performance in the author's original paper. 

Similarly to FCN, the first three layers are convolutional with some relatively small modifications. 
The first convolution is composed of 128 filters of length 5; the second convolution is composed of 256 filters of length 11; the third convolution is composed of 512 filters of length 21. 
Each convolution is followed by an instance normalization operation~\citep{ulyanov2016instance} whose output is fed to the Parametric Rectified Linear Unit (PReLU)~\citep{he2015delving} activation function. 
The output of PReLU is followed by a dropout operation (with a rate equal to 0.2) and a final max pooling of length 2. 
The third convolutional layer is fed to an attention mechanism~\citep{bahdanau2015neural} that enables the network to learn which parts of the time series (in the time domain) are important for a certain classification. 
More precisely, to implement this technique, the input MTS is multiplied with a second MTS of the same length and number of channels, except that the latter has gone through the softmax function.
Each element in the second MTS will act as a weight for the first MTS, thus enabling the network to learn the importance of each element (time stamp).  
Finally, a traditional softmax classifier is fully connected to the latter layer with a number of neurons equal to the number of classes in the dataset. 

In addition to replacing the GAP layer with the attention layer, Encoder differs from FCN in three main core changes: (1) the PReLU activation function where an additional parameter is added for each filter to enable learning the slope of the function, (2) the dropout regularization technique and (3) the max pooling operation. 
One final note is that the careful design of Encoder's attention mechanism enabled the invariance across all layers which encouraged the authors to implement a transfer learning approach.

\subsubsection{Multi-scale Convolutional Neural Network}
Originally proposed by~\cite{cui2016multi}, Multi-scale Convolutional Neural Network (MCNN) is the earliest approach to validate an end-to-end deep learning architecture on the UCR Archive. 
MCNN's architecture is very similar to a traditional CNN model: with two convolutions (and max pooling) followed by an FC layer and a final softmax layer. 
On the other hand, this approach is very complex with its heavy data pre-processing step. 
\cite{cui2016multi} were the first to introduce the Window Slicing (WS) method as a data augmentation technique.
WS slides a window over the input time series and extract subsequences, thus training the network on the extracted subsequences instead of the raw input time series.
Following the extraction of a subsequence from an input time series using the WS method, a transformation stage is used. 
More precisely, prior to any training, the subsequence will undergo three transformations: (1) identity mapping; (2)  down-sampling and (3) smoothing; thus, transforming a univariate input time series into a multivariate input time series. 
This heavy pre-processing would question the end-to-end label of this approach, but since their method is generic enough we incorporated it into our developed framework. 

For the first transformation, the input subsequence is left unchanged and the raw subsequence will be used as an input for an independent first convolution. 
The down-sampling technique (second transformation) will result in shorter subsequences with different lengths which will then undergo another independent convolutions in parallel to the first convolution. 
As for the smoothing technique (third transformation), the result is a smoothed subsequence whose length is equal to the input raw subsequence which will also be fed to an independent convolution in parallel to the first and the second convolutions. 

The output of each convolution in the first convolutional stage is concatenated to form the input of the subsequent convolutional layer. 
Following this second layer, an FC layer is deployed with 256 neurons using the sigmoid activation function. 
Finally, the usual softmax classifier is used with a number of neurons equal to the number of classes in the dataset.  

Note that each convolution in this network uses 256 filters with the sigmoid as an activation function, followed by a max pooling operation. 
Two architecture hyperparameters are cross-validated, using a grid search on an unseen split from the training set: the filter length and the pooling factor which determines the pooling size for the max pooling operation.   
The total number of layers in this network is 4, out of which only the first two convolutional layers are invariant (transferable).  
Finally, since the WS method is also used at test time, the class of an input time series is determined by a majority vote over the extracted subsequences' predicted labels. 

\subsubsection{Time Le-Net}
Time Le-Net (t-LeNet) was originally proposed by~\cite{leguennec2016data} and inspired by the great performance of LeNet's architecture for the document recognition task~\citep{lecun1998gradient}. 
This model can be considered as a traditional CNN with two convolutions followed by an FC layer and a final softmax classifier. 
There are two main differences with the FCNs: (1) an FC layer and (2) local max-pooling operations.   
Unlike GAP, local pooling introduces invariance to small perturbations in the activation map (the result of a convolution) by taking the maximum value in a local pooling window.
Therefore for a pool size equal to 2, the pooling operation will halve the length of a time series by taking the maximum value between each two time steps. 

For both convolutions, the ReLU activation function is used with a filter length equal to 5. 
For the first convolution, 5 filters are used and followed by a max pooling of length equal to 2. 
The second convolution uses 20 filters followed by a max pooling of length equal to 4. 
Thus, for an input time series of length $l$, the resulting output of these two convolutions will divide the length of the time series by $8=4\times2$. 
The convolutional blocks are followed by a non-linear fully connected layer which is composed of 500 neurons, each one using the ReLU activation function.
Finally, similarly to all previous architectures, the number of neurons in the final softmax classifier is equal to the number of classes in a dataset. 

Unlike ResNet and FCN, this approach does not have much invariant layers (2 out of 4) due to the use of an FC layer instead of a GAP layer, thus increasing drastically the number of parameters needed to be trained which also depends on the length of the input time series. 
Thus, the transferability of this network is limited to the first two convolutions whose number of parameters depends solely on the number and length of the chosen filters. 

We should note that t-LeNet is one of the approaches adopting a data augmentation technique to prevent overfitting especially for the relatively small time series datasets in the UCR/UEA archive. 
Their approach uses two data augmentation techniques: WS and Window Warping (WW). 
The former me\-thod is identical to MCNN's data augmentation technique originally proposed in~\cite{cui2016multi}.  
As for the second data augmentation technique, WW employs a warping technique that squeezes or dilates the time series. 
In order to deal with multi-length time series the WS method is adopted to ensure that subsequences of the same length are extracted for training the network. 
Therefore, a given input time series of length $l$ is first dilated ($\times2$) then squeezed ($\times\frac{1}{2}$) resulting in three time series of length $l$, $2l$ and $\frac{1}{2}l$ that are fed to WS to extract equal length subsequences for training.
Not that in their original paper~\citep{leguennec2016data}, WS' length is set to $0.9l$. 
Finally similarly to MCNN, since the WS method is also used at test time, a majority vote over the extracted subsequences' predicted labels is applied. 

\subsubsection{Multi Channel Deep Convolutional Neural Network}
Multi Channel Deep Convolutional Neural Network (MCDCNN) was originally proposed and validated on two multivariate time series datasets~\citep{zheng2014time,zheng2016exploiting}.
The proposed architecture is mainly a traditional deep CNN with one modification for MTS data: the convolutions are applied independently (in parallel) on each dimension (or channel) of the input MTS.  

Each dimension for an input MTS will go through two convolutional stages with 8 filters of length 5 with ReLU as the activation function.
Each convolution is followed by a max pooling operation of length 2. 
The output of the second convolutional stage for all dimensions is concatenated over the channels axis and then fed to an FC layer with 732 neurons with ReLU as the activation function. 
Finally, the softmax classifier is used with a number of neurons equal to the number of classes in the dataset. 
By using an FC layer before the softmax classifier, the transferability of this network is limited to the first and second convolutional layers. 

\subsubsection{Time Convolutional Neural Network}
Time-CNN approach was originally proposed by~\cite{zhao2017convolutional} for both univariate and multivariate TSC. 
There are three main differences compared to the previously described networks.
The first characteristic of Time-CNN is the use of the mean squared error (MSE) instead of the traditional categorical cross-entropy loss function, which has been used by all the deep learning approaches we have mentioned so far.
Hence, instead of a softmax classifier, the final layer is a traditional FC layer with sigmoid as the activation function, which does not guarantee a sum of probabilities equal to 1.  
Another difference to traditional CNNs is the use of a local \emph{average} pooling operation instead of local \emph{max} pooling. 
In addition, unlike MCDCNN, for MTS data they apply one convolution for all the dimensions of a multivariate classification task. 
Another unique characteristic of this architecture is that the final classifier is fully connected directly to the output of the second convolution, which removes completely the GAP layer without replacing it with an FC non-linear layer. 

The network is composed of two consecutive convolutional layers with respectively 6 and 12 filters followed by a local average pooling operation of length 3.
The convolutions adopt the sigmoid as the activation function.   
The network's output consists of an FC layer with a number of neurons equal to the number of classes in the dataset. 

\subsubsection{Time Warping Invariant Echo State Network}
Time Warping Invariant Echo State Network (TWIESN)~\citep{tanisaro2016time} is the only non-convolutional \emph{recurrent} architecture tested and re-implemented in our study.
Although ESNs were originally proposed for time series forecasting,~\cite{tanisaro2016time} proposed a variant of ESNs that uses directly the raw input time series and predicts a probability distribution over the class variables.

In fact, for each element (time stamp) in an input time series, the reservoir space is used to project this element into a higher dimensional space. 
Thus, for a univariate time series, the element is projected into a space whose dimensions are inferred from the size of the reservoir. 
Then for each element, a Ridge classifier~\citep{hoerl1970ridge} is trained to predict the class of each time series element. 
During test time, for each element of an input test time series, the already trained Ridge classifier will output a probability distribution over the classes in a dataset. 
Then the a posteriori probability for each class is averaged over all time series elements, thus assigning for each input test time series the label for which the averaged probability is maximum.
Following the original paper of~\cite{tanisaro2016time}, using a grid-search on an unseen split ($20\%$) from the training set, we optimized TWIESN's three hyperparameters: the reservoir's size, sparsity and spectral radius.    

\subsection{Hyperparameters}

Tables~\ref{tab-archi} and~\ref{tab-optim} show respectively the architecture and the optimization hyperparameters for all the described approaches except for TWIESN, since its hyperparameters are not compatible with the eight other algorithms' hyperparameters. 
We should add that for all the other deep learning classifiers (with TWIESN omitted), a model checkpoint procedure was performed either on the training set or a validation set (split from the training set).
Which means that if the model is trained for 1000 epochs, the best one on the validation set (or the train set) loss will be chosen for evaluation. 
This characteristic is included in \tablename~\ref{tab-optim} under the ``valid'' column. 
In addition to the model checkpoint procedure, we should note that all deep learning models in \tablename~\ref{tab-archi} were initialized randomly using Glorot's uniform initialization method~\citep{glorot2010understanding}. 
All models were optimized using a variant of Stochastic Gradient Descent (SGD) such as Adam~\citep{kingma2015adam} and AdaDelta~\citep{zeiler2012adadelta}.
We should add that for FCN, ResNet and MLP proposed in~\cite{wang2017time}, the learning rate was reduced by a factor of $0.5$ each time the model's training loss has not improved for $50$ consecutive epochs (with a minimum value equal to $0.0001$).  
One final note is that we have no way of controlling the fact that those described architectures might have been overfitted for the UCR/UEA archive and designed empirically to achieve a high performance, which is always a risk when comparing classifiers on a benchmark~\citep{bagnall2017the}. 
We therefore think that challenges where only the training data is publicly available and the testing data are held by the challenge organizer for evaluation might help in mitigating this problem.   

\begin{table}
	\centering
	\setlength\tabcolsep{7.2pt}
	{ \small
		\begin{tabularx}{\textwidth}{lllllllll}
			\midrule
			\multirow{3}{*}{Methods} &
			\multicolumn{8}{l}{Architecture}\\ \cline{2-9}\\
			
			& \#Layers & \#Conv & \#Invar & Normalize & Pooling & Feature & Activate & Regularize \\ 
			\midrule
			
			MLP
			& 4 & 0 & 0 & None & None & FC & ReLU & Dropout \\
			
			FCN
			& 5 & 3 & 4 & Batch & None & GAP & ReLU & None\\
			
			ResNet
			& 11 & 9 & 10 & Batch & None & GAP & ReLU & None\\
			
			Encoder
			& 5 & 3 & 4 & Instance & Max & Att & PReLU & Dropout\\
			
			MCNN
			& 4 & 2 & 2 & None & Max & FC & Sigmoid & None \\
			
			t-LeNet
			& 4 & 2 & 2 & None & Max & FC & ReLU & None\\
			
			MCDCNN
			& 4 & 2 & 2 & None & Max & FC & ReLU & None\\
			
			Time-CNN
			& 3 & 2 & 2 & None & Avg & Conv & Sigmoid & None\\
			\hline
		\end{tabularx}
	}
	\caption{Architecture's hyperparameters for the deep learning approaches}\label{tab-archi}
\end{table}
\begin{table}
	\centering
	\setlength\tabcolsep{10.1pt}
	{\small
		\begin{tabularx}{\textwidth}{llllllll}
			\midrule
			\multirow{3}{*}{Methods} &
			
			\multicolumn{7}{l}{Optimization} \\ \cline{2-8} \\
			
			& Algorithm & Valid & Loss & Epochs & Batch & Learning rate & Decay \\ 
			\midrule
			
			MLP
			& AdaDelta & Train & Entropy & 5000 & 16 & 1.0 & 0.0  \\
			
			FCN
			& Adam & Train & Entropy & 2000 & 16 & 0.001 & 0.0 \\
			
			ResNet
			& Adam & Train & Entropy & 1500 & 16 & 0.001 & 0.0\\
			
			Encoder
			& Adam & Train & Entropy & 100 & 12 & 0.00001 & 0.0 \\
			
			MCNN
			& Adam & Split$_{20\%}$ & Entropy & 200 & 256 & 0.1 & 0.0 \\
			
			t-LeNet
			& Adam & Train & Entropy & 1000 & 256 & 0.01 & 0.005  \\
			
			MCDCNN
			& SGD & Split$_{33\%}$ & Entropy & 120 & 16 & 0.01 & 0.0005 \\
			
			Time-CNN
			& Adam & Train & MSE & 2000 & 16 & 0.001 & 0.0 \\
			\hline
		\end{tabularx}
	}
	\caption{Optimization's hyperparameters for the deep learning approaches}\label{tab-optim}
\end{table}

\section{Experimental setup}\label{sec-experiment}
We first start by presenting the datasets' properties we have adopted in this empirical study. 
We then describe in details our developed open-source framework of deep learning for time series classification. 

\subsection{Datasets}
\subsubsection{Univariate archive}
In order to have a thorough and fair experimental evaluation of all approaches, we tested each algorithm on the whole UCR/UEA archive~\citep{ucrarchive,bagnall2017the} which contains 85 univariate time series datasets.
The datasets possess different varying characteristics such as the length of the series which has a minimum value of 24 for the ItalyPowerDemand dataset and a maximum equal to 2,709 for the HandOutLines dataset. 
One important characteristic that could impact the DNNs' accuracy is the size of the training set which varies between 16 and 8926 for respectively DiatomSizeReduction and ElectricDevices datasets.   
We should note that twenty datasets contains a relatively small training set (50 or fewer instances) which surprisingly was not an impediment for obtaining high accuracy when applying a very deep architecture such as ResNet. 
Furthermore, the number of classes varies between 2 (for 31 datasets) and 60 (for the ShapesAll dataset). 
Note that the time series in this archive are already z-normalized~\citep{bagnall2017the}.

Other than the fact of being publicly available, the choice of validating on the UCR/UEA archive is motivated by having datasets from different domains which have been broken down into seven different categories (Image Outline, Sensor Readings, Motion Capture, Spectrographs, ECG, Electric Devices and Simulated Data) in~\cite{bagnall2017the}. 
Further statistics, which we do not repeat for brevity, were conducted on the UCR/UEA archive in~\cite{bagnall2017the}. 

\subsubsection{Multivariate archive}
We also evaluated all deep learning models on Baydogan's archive~\citep{baydogan2015mts} that contains 13 MTS classification datasets. 
For memory usage limitations over a single GPU, we left the MTS dataset Performance Measurement System (PeMS) out of our experimentations.
This archive also exhibits datasets with different characteristics such as the length of the time series which, unlike the UCR/UEA archive, varies among the same dataset.
This is due to the fact that the datasets in the UCR/UEA archive are already re-scaled to have an equal length among one dataset~\citep{bagnall2017the}.

In order to solve the problem of unequal length time series in the MTS archive we decided to linearly interpolate the time series of each dimension for every given MTS, thus each time series will have a length equal to the longest time series' length.
This form of pre-processing has also been used by~\cite{ratanamahatana2005three} to show that the length of a time series is not an issue for TSC problems.
This step is very important for deep learning models whose architecture depends on the length of the input time series (such as a MLP) and for parallel computation over the GPUs.
We did not z-normalize any time series, but we emphasize that this traditional pre-processing step~\citep{bagnall2017the} should be further studied for univariate as well as multivariate data, especially since normalization is known to have a huge effect on DNNs' learning capabilities~\citep{zhang2017understanding}. 
Note that this process is only true for the MTS datasets whereas for the univariate benchmark, the time series are already z-normalized.
Since the data is pre-processed using the same technique for all nine classifiers, we can safely say, to some extent, that the accuracy improvement of certain models can be solely attributed to the model itself. 
\tablename~\ref{tab-mts} shows the different characteristics of each MTS dataset used in our experiments. 

\begin{table}
	\centering
	\setlength\tabcolsep{11pt}
	{\small
		\begin{tabularx}{\textwidth}{lllllll}
			\midrule
			\multirow{1}{*}{Dataset} 
			& Old length & New length & Classes & Dimensions & Train & Test \\ 
			\toprule
			ArabicDigits & 4-93 & 93 & 10 & 13 & 6600 & 2200 \\
			AUSLAN & 45-136 & 136 & 95 & 22 & 1140 & 1425 \\
			CharacterTrajectories & 109-205 & 205 & 20 & 3 & 300 & 2558 \\
			CMUsubject16 & 127-580 & 580 & 2 & 62 & 29 & 29 \\
			ECG & 39-152 & 152 & 2 & 2 & 100 & 100 \\ 
			JapaneseVowels & 7-29 & 29 & 9 & 12 & 270 & 370 \\
			KickVsPunch & 274-841 & 841 & 2 & 62 & 16 & 10 \\
			Libras & 45-45 & 45 & 15 & 2 & 180 & 180 \\
			Outflow & 50-997 & 997 & 2 & 4 & 803 & 534 \\
			UWave & 315-315 & 315 & 8 & 3 & 200 & 4278 \\ 
			Wafer & 104-198 & 198 & 2 & 6 & 298 & 896 \\
			WalkVsRun & 128-1918 & 1919 & 2 & 62 & 28 & 16 \\    
			\bottomrule
		\end{tabularx}
		\caption{The multivariate time series classification archive.}\label{tab-mts}
		
	}
\end{table}

\subsection{Experiments}
For each dataset in both archives (97 datasets in total), we have trained the nine deep learning models (presented in the previous Section) with 10 different runs each.
Each run uses the same original train/test split in the archive but with a different random weight initialization, which enables us to take the mean accuracy over the 10 runs in order to reduce the bias due to the weights' initial values.
In total, we have performed 8730 experiments for the 85 univariate and 12 multivariate TSC datasets.
Thus, given the huge number of models that needed to be trained, we ran our experiments on a cluster of 60 GPUs. 
These GPUs were a mix of four types of Nvidia graphic cards: GTX 1080 Ti, Tesla K20, K40 and K80. 
The total sequential running time was approximately 100 days, that is if the computation has been done on a single GPU. 
However, by leveraging the cluster of 60 GPUs, we managed to obtain the results in less than one month.
We implemented our framework using the open source deep learning library Keras~\citep{chollet2015keras} with the Tensorflow~\citep{tensorflow2015whitepaper} back-end\footnote{The implementations are available on \url{https://github.com/hfawaz/dl-4-tsc}}.

Following~\cite{lucas2018proximity,forestier2017generating,petitjean2016faster,grabocka2014learning} we used the mean accuracy measure averaged over the 10 runs on the test set. 
When comparing with the state-of-the-art results published in~\cite{bagnall2017the} we averaged the accuracy using the median test error. 
Following the recommendation in~\cite{demsar2006statistical} we used the Friedman test~\citep{friedman1940a} to reject the null hypothesis. 
Then we performed the pairwise post-hoc analysis recommended by~\cite{benavoli2016should} where the average rank comparison is replaced by a Wilcoxon signed-rank test~\citep{wilcoxon1945individual} with Holm's alpha ($5\%$) correction~\citep{holm1979a,garcia2008an}. 
To visualize this type of comparison we used a critical difference diagram proposed by~\cite{demsar2006statistical}, where a thick horizontal line shows a group of classifiers (a clique) that are not-significantly different in terms of accuracy. 

\section{Results}\label{sec-results}
In this section, we present the accuracies for each one of the nine approaches. 
All accuracies are absolute and not relative to each other that is if we claim algorithm A is 5\% better than algorithm B, this means that the average accuracy is 0.05 higher for algorithm A than B.

\begin{figure}
\centering
\includegraphics[width=0.8\linewidth]{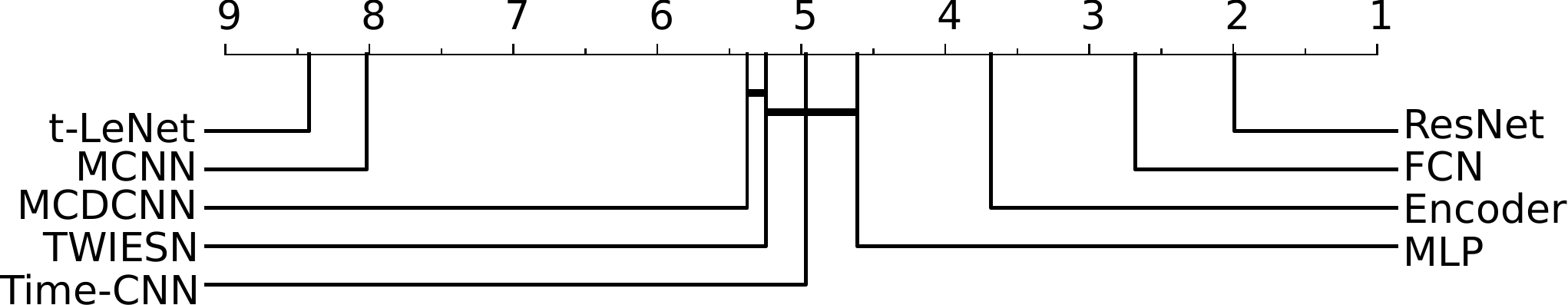}
\caption{Critical difference diagram showing pairwise statistical difference comparison of nine deep learning classifiers on the univariate UCR/UEA time series classification archive.}
\label{fig-cd-diagram-ucr}
\end{figure}

\subsection{Results for univariate time series}
We provide on the companion GitHub repository the raw accuracies over the 10 runs for the nine deep learning models we have tested on the 85 univariate time series datasets: the UCR/UEA archive~\citep{ucrarchive,bagnall2017the}.
The corresponding critical difference diagram is shown in Figure~\ref{fig-cd-diagram-ucr}.
The ResNet significantly outperforms the other approaches with an average rank of almost 2. 
ResNet wins on 50 problems out of 85 and significantly outperforms the FCN architecture. 
This is in contrast to the original paper's results where FCN was found to outperform ResNet on 18 out of 44 datasets, which shows the importance of validating on a larger archive in order to have a robust statistical significance. 

We believe that the success of ResNet is highly due to its deep flexible architecture.
First of all, our findings are in agreement with the deep learning for computer vision literature where deeper neural networks are much more successful than shallower architectures~\citep{he2016deep}.
In fact, in a space of 4 years, neural networks went from 7 layers in AlexNet 2012~\citep{krizhevsky2012imagenet} to \emph{1000} layers for ResNet 2016~\citep{he2016deep}.  
These types of deep architectures generally need a huge amount of data in order to generalize well on unseen examples~\citep{he2016deep}.
Although the datasets used in our experiments are relatively small compared to the billions of labeled images (such as {ImageNet}~\citep{russakovsky2015imagenet} and {OpenImages}~\citep{openimages} challenges), the deepest networks did reach competitive accuracies on the UCR/UEA archive benchmark. 

We give two potential reasons for this high generalization capabilities of deep CNNs on the TSC tasks. 
First, having seen the success of convolutions in classification tasks that require learning features that are spatially invariant in a \emph{two} dimensional space (such as width and height in images), it is only natural to think that discovering patterns in a \emph{one} dimensional space (time) should be an easier task for CNNs thus requiring less data to learn from.   
The other more direct reason behind the high accuracies of deep CNNs on time series data is its success in other sequential data such as speech recognition~\citep{hinton2012deep} and sentence classification~\citep{kim2014convolutional} where text and audio, similarly to time series data, exhibit a natural temporal ordering.   

The MCNN and t-LeNet architectures yielded very low accuracies with only one win for the Earthquakes dataset. 
The main common idea between both of these approaches is extracting subsequences to augment the training data. 
Therefore the model learns to classify a time series from a shorter subsequence instead of the whole one, then with a majority voting scheme the time series at test time are assigned a class label. 
The poor performances (worst average ranks) for these two approaches suggest that this ad-hoc method of slicing the time series does not guarantee that the discriminative information of a time series has not been lost.     
These two classifiers are similar to the phase dependent intervals TSC algorithms~\citep{bagnall2017the} where the classifiers derive features from intervals of each series. 
Similarly to the recent comparative study of TSC algorithms, this type of Window Slicing based approach yielded the lowest average ranks. 

Although MCDCNN and Time-CNN were originally proposed to classify MTS datasets, we have evaluated them on the univariate UCR/UEA archive.
The MCDCNN did not manage to beat any of the classifiers except for the ECG5000 dataset which is already a dataset where almost all approaches reached the highest accuracy.
This low performance is probably due to the non-linear FC layer that replaces the GAP pooling of the best performing algorithms (FCN and ResNet). 
This FC layer reduces the effect of learning time invariant features which explains why MLP, Time-CNN and MCDCNN exhibit very similar performance.  

One approach that shows relatively high accuracy is Encoder~\citep{serra2018towards}. 
The statistical test indicates a significant difference between Encoder, FCN and ResNet. 
FCN wins on 36 datasets whereases Encoder wins only on 17 which suggests the superiority of the GAP layer compared to Encoder's attention mechanism.

\subsection{Comparing with state-of-the-art approaches} 
In this section, we compared ResNet (the most accurate DNN of our study) with the current state-of-the-art classifiers evaluated on the UCR/UEA archive in the great time series classification bake off~\citep{bagnall2017the}. 
Note that our empirical study strongly suggests to use ResNet instead of any other deep learning algorithm - it is the most accurate one with similar runtime to FCN (the second most accurate DNN). 
Finally, since ResNet's results were averaged over ten different random initializations, we chose to take one iteration of ResNet (the median) and compare it to other state-of-the-art algorithms that were executed once over the original train/test split provided by the UCR/UEA archive. 

\begin{figure}
\centering
\includegraphics[width=0.8\linewidth]{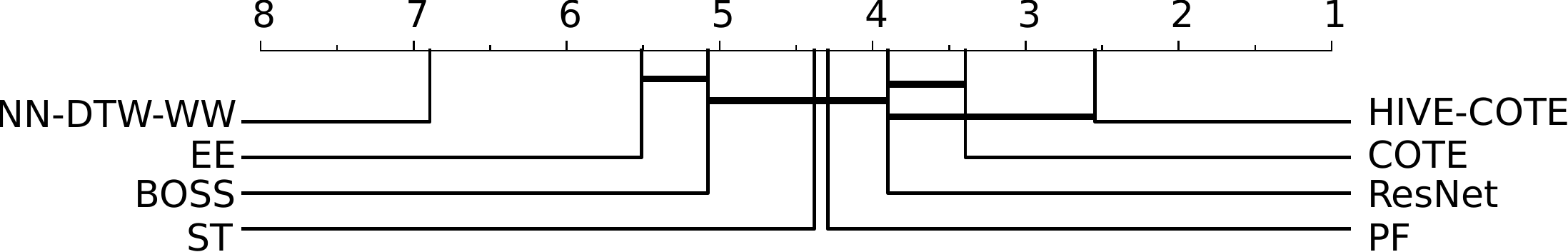}
\caption{Critical difference diagram showing pairwise statistical difference comparison of state-of-the-art classifiers on the univariate UCR/UEA time series classification archive.}
\label{fig-cd-diagram-uea}
\end{figure}

Out of the 18 classifiers evaluated by~\cite{bagnall2017the}, we have chosen the four best performing algorithms: (1) Elastic Ensemble (EE) proposed by~\cite{lines2015time} is an ensemble of nearest neighbor classifiers with 11 different time series similarity measures; (2) Bag-of-SFA-Symbols (BOSS) published in~\cite{schafer2015the} forms a discriminative bag of words by discretizing the time series using a Discrete Fourier Transform and then building a nearest neighbor classifier with a bespoke distance measure; (3) Shapelet Transform (ST) developed by~\cite{hills2014classification} extracts discriminative subsequences (shapelets) and builds a new representation of the time series that is fed to an ensemble of 8 classifiers; (4) Collective of Transformation-based Ensembles (COTE) proposed by~\cite{bagnall2017the} is basically a weighted ensemble of 35 TSC algorithms including EE and ST.
We also include the Hierarchical Vote Collective of Transformation-Based Ensembles (HIVE-COTE) proposed by~\cite{lines2018time} which improves significantly COTE's performance by leveraging a hierarchical voting system as well as adding two new classifiers and two additional transformation domains. 
In addition to these five state-of-the-art classifiers, we have included the classic nearest neighbor coupled with DTW and a warping window (WW) set through cross-validation on the training set (denoted by NN-DTW-WW), since it is still one of the most popular methods for classifying time series data~\citep{bagnall2017the}.
Finally, we  added a recent approach named Proximity Forest (PF) which is similar to Random Forest but replaces the attribute based splitting criteria by a random similarity measure chosen out of EE's elastic distances~\citep{lucas2018proximity}. 
Note that we did not implement any of the non-deep TSC algorithms. 
We used the results provided by~\cite{bagnall2017the} and the other corresponding papers to construct the critical difference diagram in Figure~\ref{fig-cd-diagram-uea}.

Figure~\ref{fig-cd-diagram-uea} shows the critical difference diagram over the UEA benchmark with ResNet added to the pool of six classifiers.
As we have previously mentioned, the state-of-the-art classifiers are compared to ResNet's median accuracy over the test set.
Nevertheless, we generated the ten different average ranks for each iteration of ResNet and observed that the ranking of the compared classifiers is stable for the ten different random initializations of ResNet. 
The statistical test failed to find any significant difference between COTE/HIVE-COTE and ResNet which is the only TSC algorithm that was able to reach similar performance to COTE. 
Note that for the ten different random initializations of ResNet, the pairwise statistical test always failed to find any significance between ResNet and COTE/HIVE-COTE.
PF, ST, BOSS and ResNet showed similar performances according to the Wilcoxon signed-rank test, but the fact that ResNet is not significantly different than COTE suggests that more datasets would give a better insight into these performances~\citep{demsar2006statistical}.
NN-DTW-WW and EE showed the lowest average rank suggesting that these methods are no longer competitive with current state-of-the-art algorithms for TSC. 
It is worthwhile noting that cliques formed by the Wilcoxon Signed Rank Test with Holm's alpha correction do not necessary reflect the rank order~\citep{lines2018time}. 
For example, if we have three classifiers ($C_1,C_2,C_3$) with average ranks ($C_1>C_2>C_3$), one can still encounter a case where $C_1$ is not significantly worse than $C_2$ and $C_3$ with $C_2$ and $C_3$ being significantly different. 
In our experiments, when comparing to state-of-the-art algorithms, we have encountered this problem with (ResNet$>$COTE$>$HIVE-COTE). 
Therefore we should emphasize that HIVE-COTE and COTE are \emph{significantly} different when performing the pairwise statistical test.

Although HIVE-COTE is still the most accurate classifier (when evaluated on the UCR/UEA archive) its use in a real data mining application is limited due to its huge training time complexity which is $O(N^2\cdot T^4)$ corresponding to the training time of one of its individual classifiers ST.
However, we should note that the recent work of~\cite{bostrom2015binary} showed that it is possible to use a random sampling approach to decrease significantly the running time of ST (HIVE-COTE's choke-point) without any loss of accuracy.
On the other hand, DNNs offer this type of scalability evidenced by its revolution in the field of computer vision when applied to images, which are thousand times larger than time series data~\citep{russakovsky2015imagenet}. 
In addition to the huge training time, HIVE-COTE's \emph{classification} time is bounded by a linear scan of the training set due to employing a nearest neighbor classifier, whereas the trivial GPU parallelization of DNNs provides instant classification.
Finally we should note that unlike HIVE-COTE, ResNet's hyperparameters were not tuned for each dataset but rather the same architecture was used for the whole benchmark suggesting further investigation of these hyperparameters should improve DNNs' accuracy for TSC.
These results should give an insight of deep learning for TSC therefore encouraging researchers to consider the DNNs as robust real time classifiers for time series data. 

\subsubsection{The need of a fair comparison}
In this section, we highlight the fairness of the comparison to other machine learning TSC algorithms. 
Since we did not train nor test any of the state-of-the-art non deep learning algorithms, it is possible that we allowed much more training time for the described DNNs. 
For example, for a lazy machine learning algorithm such as NN-DTW, training time is zero when allowing maximum warping whereas it has been shown that judicially setting the warping window~\cite{dau2017judicious} can lead to a significant increase in accuracy. 
Therefore, we believe that allowing a much more thorough search of DTW's warping window would lead to a fairer comparison between deep learning approaches and other state-of-the-art TSC algorithms.  
In addition to cross-validating NN-DTW's hyper-parameters, we can imagine spending more time on data pre-processing and cleansing (e.g. smoothing the input time series) in order to improve the accuracy of NN-DTW~\citep{hoppner2016improving,dau2018the}. 
Ultimately, in order to obtain a fair comparison between deep learning and current state-of-the-art algorithms for TSC, we think that the time spent on optimizing a network's weights should be also spent on optimizing non deep learning based classifiers especially lazy learning algorithms such as the $K$ nearest neighbor coupled with any similarity measure. 

\subsection{Results for multivariate time series}

\begin{figure}
\centering
\includegraphics[width=0.8\linewidth]{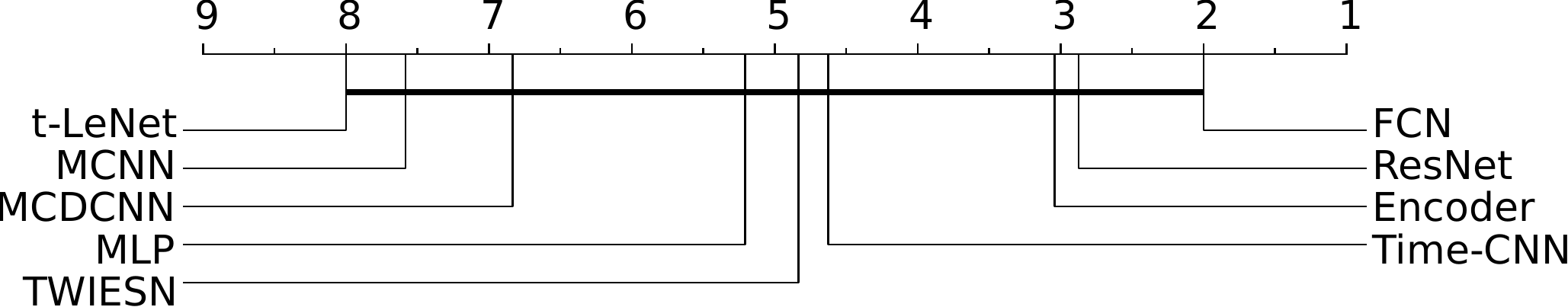}
\caption{Critical difference diagram showing pairwise statistical difference comparison of nine deep learning classifiers on the multivariate time series classification archive.}
\label{fig-cd-diagram-mts}
\end{figure}

\begin{figure}
\centering
\includegraphics[width=0.8\linewidth]{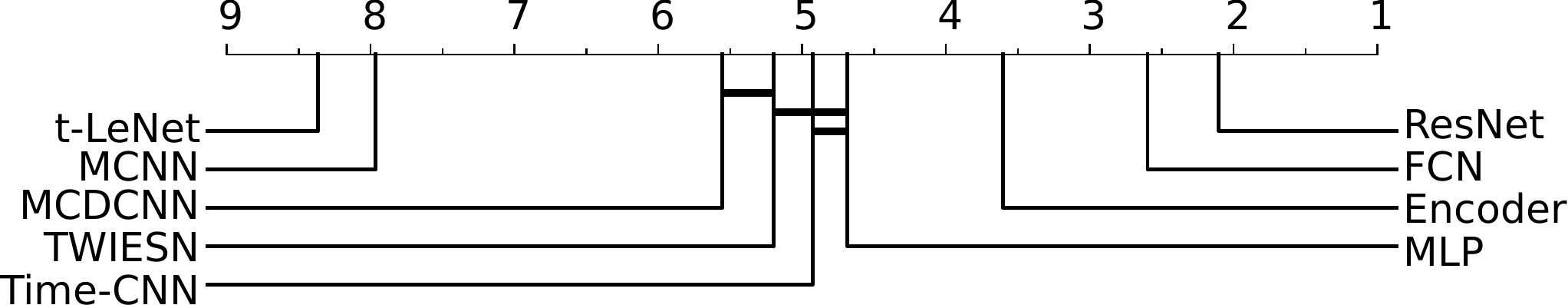}
\caption{Critical difference diagram showing pairwise statistical difference comparison of nine deep learning classifiers on both univariate and multivariate time series classification archives.}
\label{fig-cd-diagram-mts-ucr}
\end{figure}

We provide on our companion repository\footnote{\url{www.github.com/hfawaz/dl-4-tsc}} the detailed performance of the nine deep learning classifiers for 10 different random initializations over the 12 MTS classification datasets~\citep{baydogan2015mts}. 
Although Time-CNN and MCDCNN are the only architectures originally proposed for MTS data, they were outperformed by the three deep CNNs (ResNet, FCN and Encoder), which shows the superiority of these approaches on the MTS classification task. 
The corresponding critical difference diagram is depicted in Figure~\ref{fig-cd-diagram-mts}, where the statistical test failed to find any significant difference between the nine classifiers which is mainly due to the small number of datasets compared to their univariate counterpart. 
Therefore, we illustrated in Figure~\ref{fig-cd-diagram-mts-ucr} the critical difference diagram when both archives are combined (evaluation on 97 datasets in total).
At first glance, we can notice that when adding the MTS datasets to the evaluation, the critical difference diagram in Figure~\ref{fig-cd-diagram-mts-ucr} is not significantly different than the one in Figure~\ref{fig-cd-diagram-ucr} (where only the univariate UCR/UEA archive was taken into consideration). 
This is probably due to the fact that the algorithms' performance over the 12 MTS datasets is negligible to a certain degree when compared to the performance over the 85 univariate datasets. 
These observations reinforces the need to have an equally large MTS classification archive in order to evaluate hybrid univariate/multivariate time series classifiers.
The rest of the analysis is dedicated to studying the effect of the datasets' characteristics on the algorithms' performance.

\subsection{What can the dataset's characteristics tell us about the best architecture?}
The first dataset characteristic we have investigated is the problem's domain. 
\tablename~\ref{tab-perf-theme} shows the algorithms' performance with respect to the dataset's theme.
These themes were first defined in~\cite{bagnall2017the}. 
Again, we can clearly see the dominance of ResNet as the best performing approach across different domains.
One exception is the electrocardiography (ECG) datasets (7 in total) where ResNet was drastically beaten by the FCN model in 71.4\% of ECG datasets.
However, given the small sample size (only 7 datasets), we cannot conclude that FCN will almost always outperform the ResNet model for ECG datasets~\citep{bagnall2017the}. \\

\begin{table}
	\centering
	\setlength\tabcolsep{4.0pt}
	{ \small
		\begin{tabularx}{\textwidth}{llllllllll}
			\midrule
			\small Themes (\#)     & \small MLP  & \small FCN  & \small ResNet & \small Encoder & \small MCNN & \small t-LeNet & \small MCDCNN & \small Time-CNN & \small TWIESN  \\
			\toprule 
			\small DEVICE (6)    & 0.0  & 50.0 & \textbf{83.3}   & 0.0     & 0.0  & 0.0    & 0.0    & 0.0  & 0.0 \\
			\small ECG (7)       & 14.3 & \textbf{71.4} & 28.6   & 42.9    & 0.0  & 0.0    & 14.3   & 0.0  & 0.0     \\
			\small IMAGE (29)    & 6.9  & 34.5 & \textbf{48.3}   & 10.3    & 0.0  & 0.0    & 6.9    & 10.3 & 0.0     \\
			\small MOTION (14)   & 14.3 & 28.6 & \textbf{71.4}   & 21.4    & 0.0  & 0.0    & 0.0    & 0.0  & 0.0     \\
			\small SENSOR (16)   & 6.2  & 37.5 & \textbf{75.0}   & 31.2    & 6.2  & 6.2    & 6.2    & 0.0  & 12.5    \\
			\small SIMULATED (6) & 0.0  & 33.3 & \textbf{100.0}  & 33.3    & 0.0  & 0.0    & 0.0    & 0.0  & 0.0    \\
			\small SPECTRO (7) & 14.3 & 14.3 & \textbf{71.4}   & 0.0     & 0.0  & 0.0    & 0.0    & 28.6 & 28.6    \\
			\bottomrule
		\end{tabularx}
	}
	\caption{Deep learning algorithms' performance grouped by themes. 
		Each entry is the percentage of dataset themes an algorithm is most accurate for. 
		Bold indicates the best model.}\label{tab-perf-theme}
\end{table}

The second characteristic which we have studied is the time series length.
Similar to the findings for non deep learning models in~\cite{bagnall2017the}, the time series length does not give information on deep learning approaches' performance.
\tablename~\ref{tab-perf-lengths} shows the average rank of each DNN over the univariate datasets grouped by the datasets' lengths. 
One might expect that the relatively short filters (3) might affect the performance of ResNet and FCN since longer patterns cannot be captured by short filters.
However, since increasing the number of convolutional layers will increase the path length viewed by the CNN model~\citep{vaswani2017attention}, ResNet and FCN managed to outperform other approaches whose filter length is longer (21) such as Encoder.
For the recurrent TWIESN algorithm, we were expecting a poor accuracy for very long time series since a recurrent model may ``forget'' a useful information present in the early elements of a long time series.
However, TWIESN did reach competitive accuracies on several long time series datasets such as reaching a 96.8\% accuracy on Meat whose time series length is equal to 448. 
This would suggest that ESNs can solve the vanishing gradient problem especially when learning from long time series.  

\begin{table}
\centering
{ \small
\begin{tabularx}{\textwidth}{llllllllll}
\midrule
\small Length     & \small MLP  & \small FCN  & \small ResNet & \small Encoder & \small MCNN & \small t-LeNet & \small MCDCNN & \small Time-CNN & \small TWIESN \\
\toprule 
\small $<$81   & 5.43 & 3.36 & \textbf{2.43}   & 2.79    & 8.21 & 8.0    & 3.07   & 3.64 & 5.5  \\
\small 81-250   & 4.16 & \textbf{1.63} & 1.79   & 3.42    & 7.89 & 8.32   & 5.26   & 4.47 & 5.53   \\
\small 251-450  & 3.91 & 2.73 & \textbf{1.64}   & 3.32    & 8.05 & 8.36   & 6.0    & 4.68 & 4.91   \\
\small 451-700  & 4.85 & 2.69 & \textbf{1.92}   & 3.85    & 7.08 & 7.08   & 5.62   & 4.92 & 4.31   \\
\small 701-1000 & 4.6  & 1.9  & \textbf{1.6}    & 3.8     & 7.4  & 8.5    & 5.2    & 6.0  & 4.5  \\
\small $>$1000 & 3.29 & 2.71 & \textbf{1.43}   & 3.43    & 7.29 & 8.43   & 4.86   & 5.71 & 6.0 \\
\bottomrule
\end{tabularx}
}
\caption{Deep learning algorithms' average ranks grouped by the datasets' length.
Bold indicates the best model.}\label{tab-perf-lengths}
\end{table}

\begin{figure}
	\centering
    \includegraphics[width=0.7\linewidth]{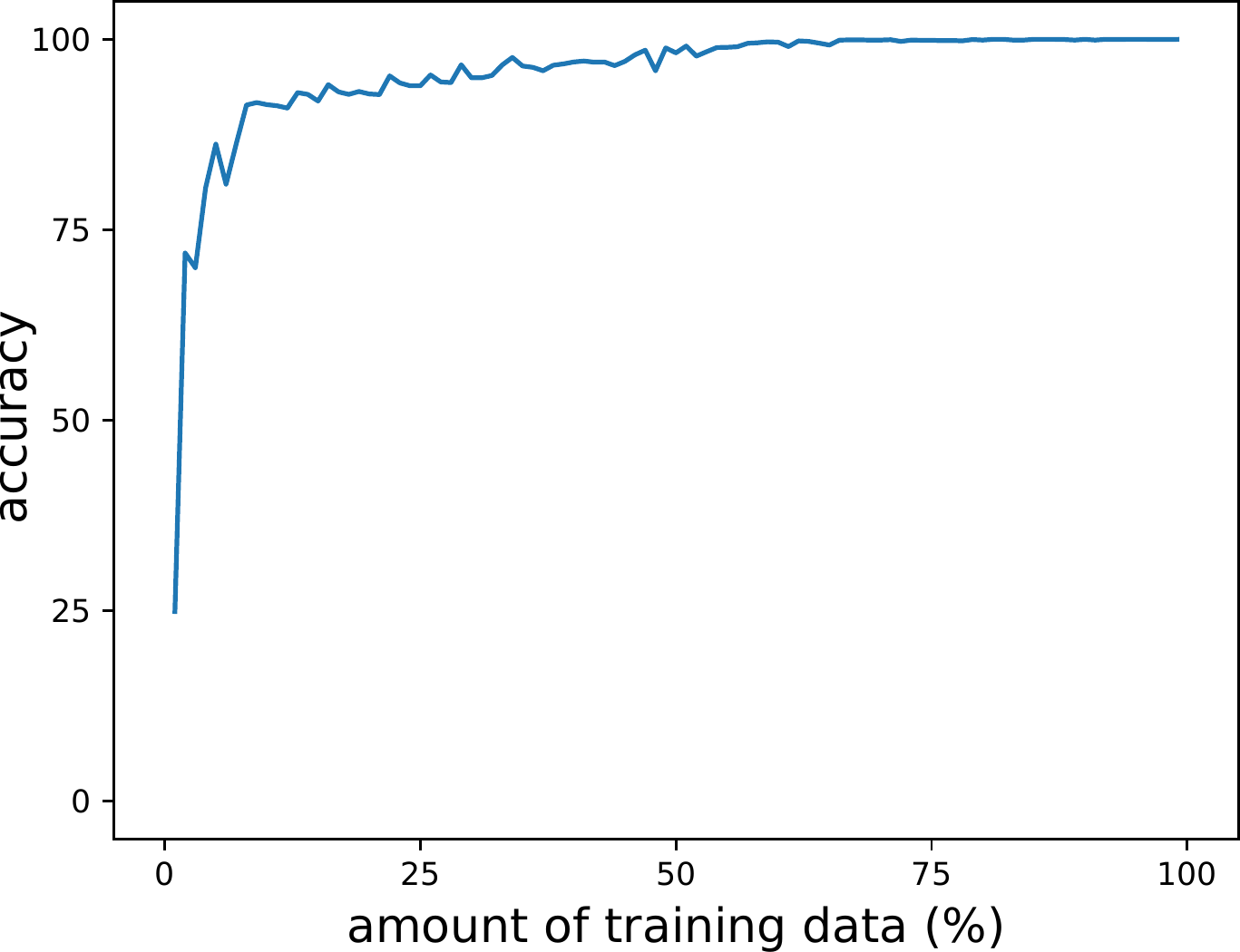}
    \caption{ResNet's accuracy variation with respect to the amount of training instances in the TwoPatterns dataset.}
    \label{fig-train-size-two-patterns}
\end{figure}

A third important characteristic is the training size of datasets and how it affects a DNN's performance. 
\tablename~\ref{tab-perf-train-size} shows the average rank for each classifier grouped by the train set's size. 
Again, ResNet and FCN still dominate with not much of a difference. 
However we found one very interesting dataset: DiatomSizeReduction. 
ResNet and FCN achieved the worst accuracy (30\%) on this dataset while Time-CNN reached the best accuracy (95\%).
Interestingly, DiatomSizeReduction is the smallest datasets in the UCR/UEA archive (with 16 training instances), which suggests that ResNet and FCN are easily overfitting this dataset. 
This suggestion is also supported by the fact that Time-CNN is the smallest model: it contains a very small number of parameters by design with only 18 filters compared to the 512 filters of FCN.
This simple architecture of Time-CNN renders overfitting the dataset much harder. 
Therefore, we conclude that the small number of filters in Time-CNN is the main reason behind its success on small datasets, however this shallow architecture is unable to capture the variability in larger time series datasets which is modeled efficiently by the FCN and ResNet architectures.
One final observation that is in agreement with the deep learning literature is that in order to achieve high accuracies while training a DNN, a large training set is needed. 
Figure~\ref{fig-train-size-two-patterns} shows the effect of the training size on ResNet's accuracy for the TwoPatterns dataset: the accuracy increases significantly when adding more training instances until it reaches 100\% for 75\% of the training data.    

\begin{table}
	\centering
	{\small
		\begin{tabularx}{\textwidth}{llllllllll}
			\midrule
			\small Train size   & \small MLP  & \small FCN  & \small ResNet & \small Encoder & \small MCNN & \small t-LeNet & \small MCDCNN & \small Time-CNN & \small TWIESN \\
			\toprule
			\small $<$100   & 4.3  & 2.03 & \textbf{1.67}   & 4.13    & 7.67 & 7.73   & 6.1    & 4.37 & 4.77  \\
			\small 100-399   & 4.85 & 2.76 & \textbf{2.06}   & 3.24    & 7.71 & 8.12   & 4.59   & 4.97 & 4.5    \\
			\small 400-799   & 3.62 & 2.38 & \textbf{1.75}   & 3.5     & 8.0  & 8.62   & 4.38   & 5.0  & 5.88   \\
			\small $>$799  & 3.85 & 2.85 & \textbf{1.62}   & 2.08    & 7.92 & 8.69   & 4.62   & 4.85 & 6.92 \\
			\bottomrule 
		\end{tabularx}
	}
	\caption{Deep learning algorithms' average ranks grouped by the training sizes.
		Bold indicates the best model.}\label{tab-perf-train-size}
\end{table}

Finally, we should note that the number of classes in a dataset - although it yielded some variability in the results for the recent TSC experimental study conducted by~\cite{bagnall2017the} - did not show any significance when comparing the classifiers based on this characteristic. 
In fact, most DNNs architectures, with the categorical cross-entropy as their cost function, employ mainly the same classifier: \emph{softmax} which is basically designed for multi-class classification. 

Overall, our results show that, on average, ResNet is the best architecture with FCN and Encoder following as second and third respectively.
ResNet performed very well in general except for the ECG datasets where it was outperformed by FCN. 
MCNN and t-LeNet, where time series were cropped into subsequences, were the worst on average. 
We found small variance between the approaches that replace the GAP layer with an FC dense layer (MCDCNN, CNN) which also showed similar performance to TWIESN and MLP. 

\subsection{Effect of random initializations}
The initialization of deep neural networks has received a significant amount of interest from many researchers in the field~\citep{lecun2015deep}. 
These advancement have contributed to a better understanding and initialization of deep learning models in order to maximize the quality of non-optimal solutions found by the gradient descent algorithm~\citep{glorot2010understanding}. 
Nevertheless, we observed in our experiments, that DNNs for TSC suffer from a significant decrease (increase) in accuracy when initialized with bad (good) random weights.  
Therefore, we study in this section, how random initializations can affect the performance of ResNet and FCN on the whole benchmark in a best and worst case scenario.  

\begin{figure}
\centering
\includegraphics[width=0.6\linewidth]{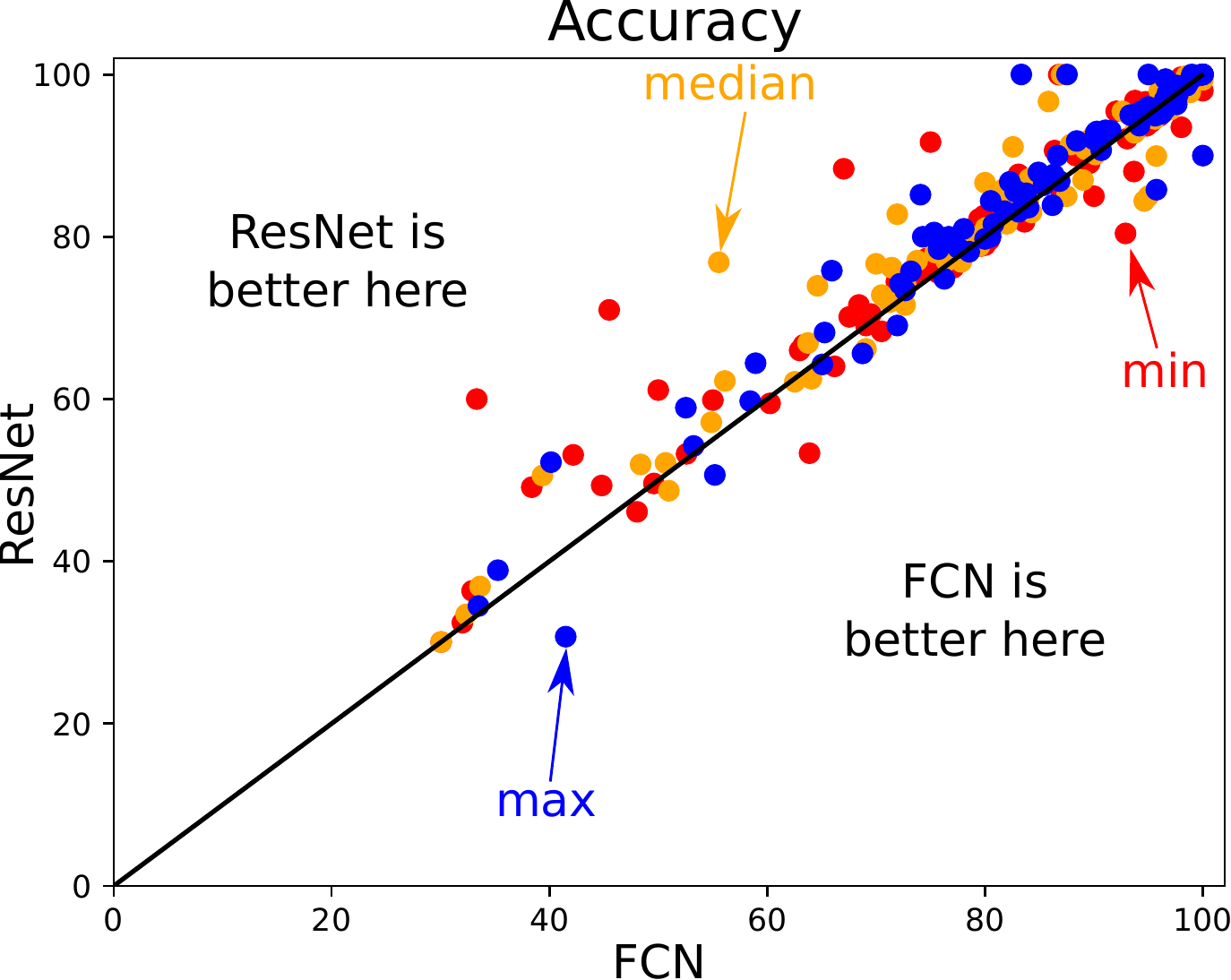}
\caption{Accuracy of ResNet versus FCN over the UCR/UEA archive when three different aggregations are taken: the minimum, median and maximum (Color figure online).}
\label{fig-resnet-vs-fcn}
\end{figure}

Figure~\ref{fig-resnet-vs-fcn} shows the accuracy plot of ResNet versus FCN on the 85 univariate time series datasets when aggregated over the 10 random initializations using three different functions: the minimum, median and maximum. 
When first observing Figure~\ref{fig-resnet-vs-fcn} one can easily conclude that ResNet has a better performance than FCN across most of the datasets regardless of the aggregation method.
This is in agreement with the critical difference diagram as well as the analysis conducted in the previous subsections, where ResNet was shown to achieve higher performance on most datasets with different characteristics.   
A deeper look into the \emph{minimum} aggregation (red points in Figure~\ref{fig-resnet-vs-fcn}) shows that FCN's performance is less stable compared to ResNet's. 
In other words, the weight's initial value can easily decrease the accuracy of FCN whereas ResNet maintained a relatively high accuracy when taking the worst initial weight values.
This is also in agreement with the average standard deviation of ResNet (1.48) which is less than FCN's (1.70).  
These observations would encourage a practitioner to avoid using a complex deep learning model since its accuracy may be unstable. 
Nevertheless we think that investigating different weight initialization techniques such as leveraging the weights of a pre-trained neural network would yield better and much more stable results~\citep{IsmailFawaz2018transfer}. 

\section{Visualization}\label{sec-visualization}
In this section, we start by investigating the use of Class Activation Map to provide an interpretable feedback that highlights the reason for a certain decision taken by the classifier.   
We then propose another visualization technique which is based on Multi-Dimensional Scaling~\citep{kruskal1978multidimensional} to understand the latent representation that is learned by the DNNs.    

\subsection{Class Activation Map}

We investigate the use of Class Activation Map (CAM) which was first introduced by~\cite{zhou2016learning} to highlight the parts of an image that contributed the most for a given class identification.
\cite{wang2017time} later introduced a one-dimensional CAM with an application to TSC. 
This method explains the classification of a certain deep learning model by highlighting the subsequences that contributed the most to a certain classification. 
Figure~\ref{fig-cam-gunpoint} and~\ref{fig-cam-meat} show the results of applying CAM respectively on GunPoint and Meat datasets.  
Note that employing the CAM is only possible for the approaches with a GAP layer preceding the softmax classifier~\citep{zhou2016learning}.
Therefore, we only considered in this section the ResNet and FCN models, who also achieved the best accuracies overall. 
Note that~\cite{wang2017time} was the only paper to propose an interpretable analysis of TSC with a DNN.
We should emphasize that this is a very important research area which is usually neglected for the sake of improving accuracy: only 2 out of the 9 approaches provided a method that explains the decision taken by a deep learning model.
In this section, we start by presenting the CAM method from a mathematical point of view and follow it with two interesting case studies on Meat and GunPoint datasets. 

By employing a Global Average Pooling (GAP) layer, ResNet and FCN benefit from the CAM method~\citep{zhou2016learning}, which makes it possible to identify which regions of an input time series constitute the reason for a certain classification.
Formally, let $A(t)$ be the result of the last convolutional layer which is an MTS with $M$ variables. 
$A_m(t)$ is the univariate time series for the variable $m\in [1,M]$, which is in fact the result of applying the $m^{th}$ filter. 
Now let $w_m^{c}$ be the weight between the $m^{th}$ filter and the output neuron of class $c$. 
Since a GAP layer is used then the input to the neuron of class $c$ ($z_c$) can be computed by the following equation: 
\begin{equation}
z_c=\sum_{m}{w_m^{c}}\sum_{t}{A_m(t)}
\end{equation}
The second sum constitutes the averaged time series over the whole time dimension but with the denominator omitted for simplicity.
The input $z_c$ can be also written by the following equation: 
\begin{equation}
z_c=\sum_{t}\sum_{m}{w_m^{c}A_m(t)}
\end{equation}
Finally the Class Activation Map ($CAM_c$) that explains the classification as label $c$ is given in the following equation: 
\begin{equation}
CAM_c(t)=\sum_{m}w_m^{c}A_m(t)
\end{equation}
CAM is actually a univariate time series where each element (at time stamp $t\in [1,T]$) is equal to the weighted sum of the $M$ data points at $t$, with the weights being learned by the neural network. 

\subsubsection{GunPoint dataset}
\begin{figure}
\centering
    \subfloat[FCN on GunPoint: Class-1]{
 \includegraphics[width=0.5\linewidth]{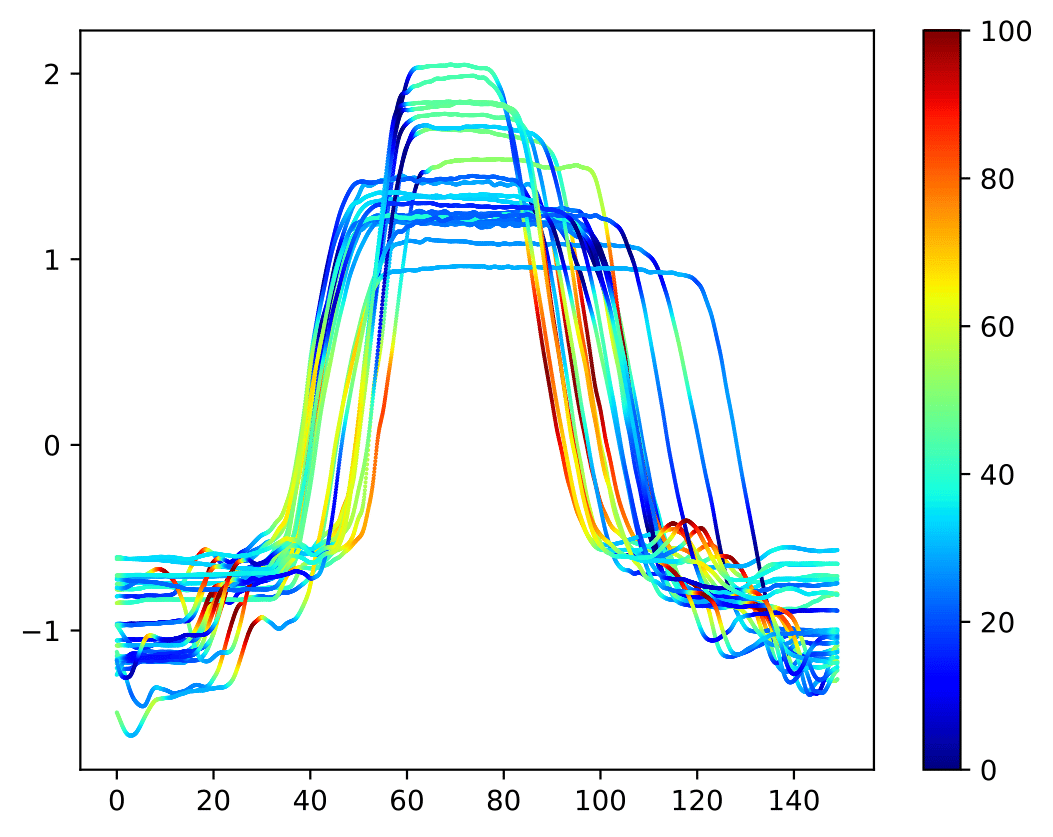}
      \label{sub-fcn-cam-GunPoint-class-1}}
    \subfloat[FCN on GunPoint: Class-2]{
 \includegraphics[width=0.5\linewidth]{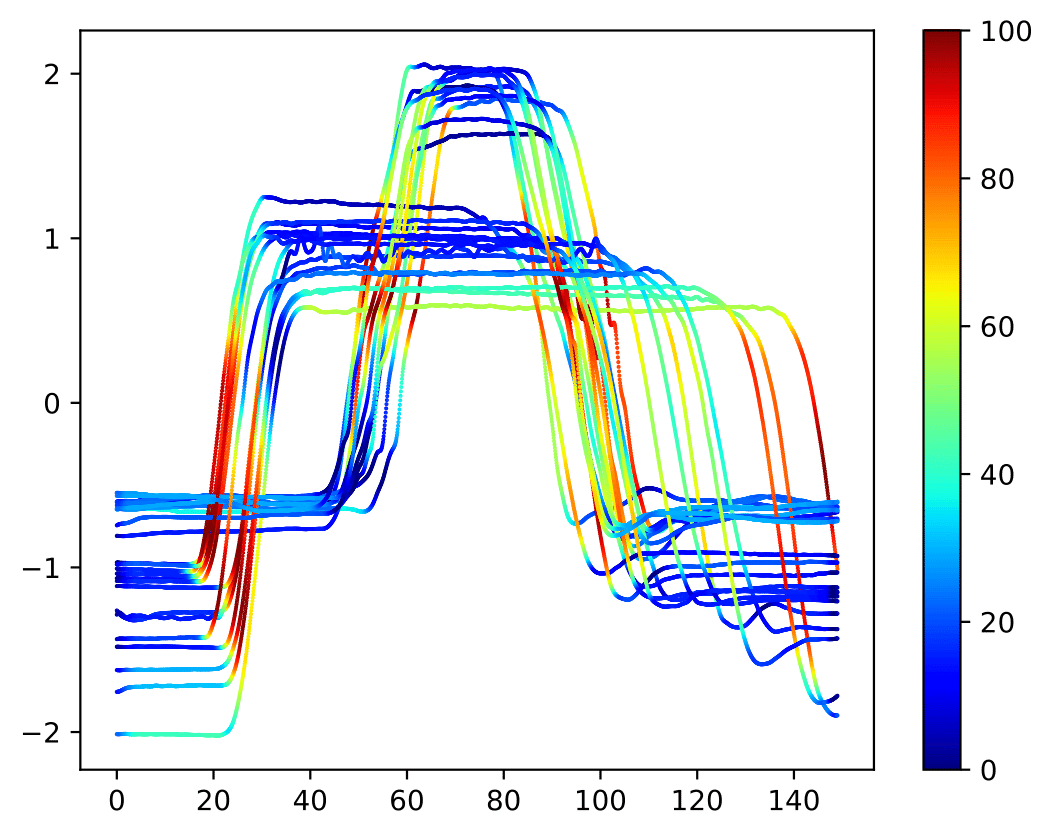}
      \label{sub-fcn-cam-GunPoint-class-2}}\\
      \subfloat[ResNet on GunPoint: Class-1]{
 \includegraphics[width=0.5\linewidth]{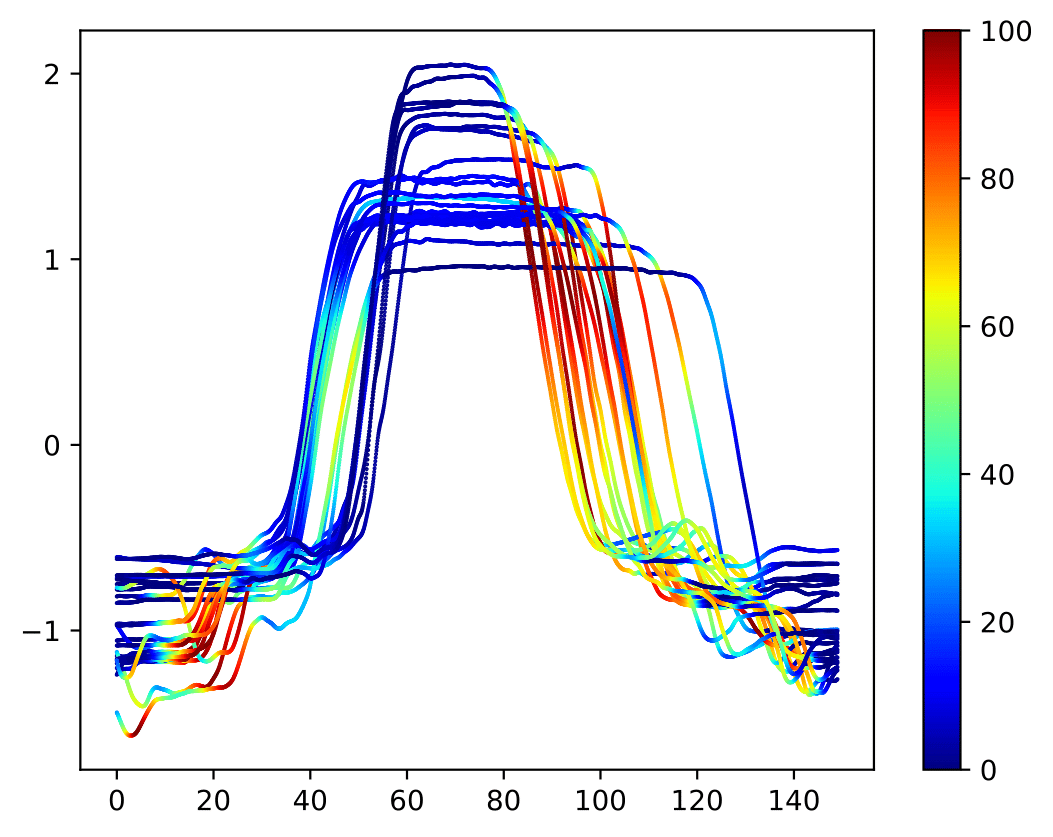}
      \label{sub-resnet-cam-GunPoint-class-1}}
    \subfloat[ResNet on GunPoint: Class-2]{
 \includegraphics[width=0.5\linewidth]{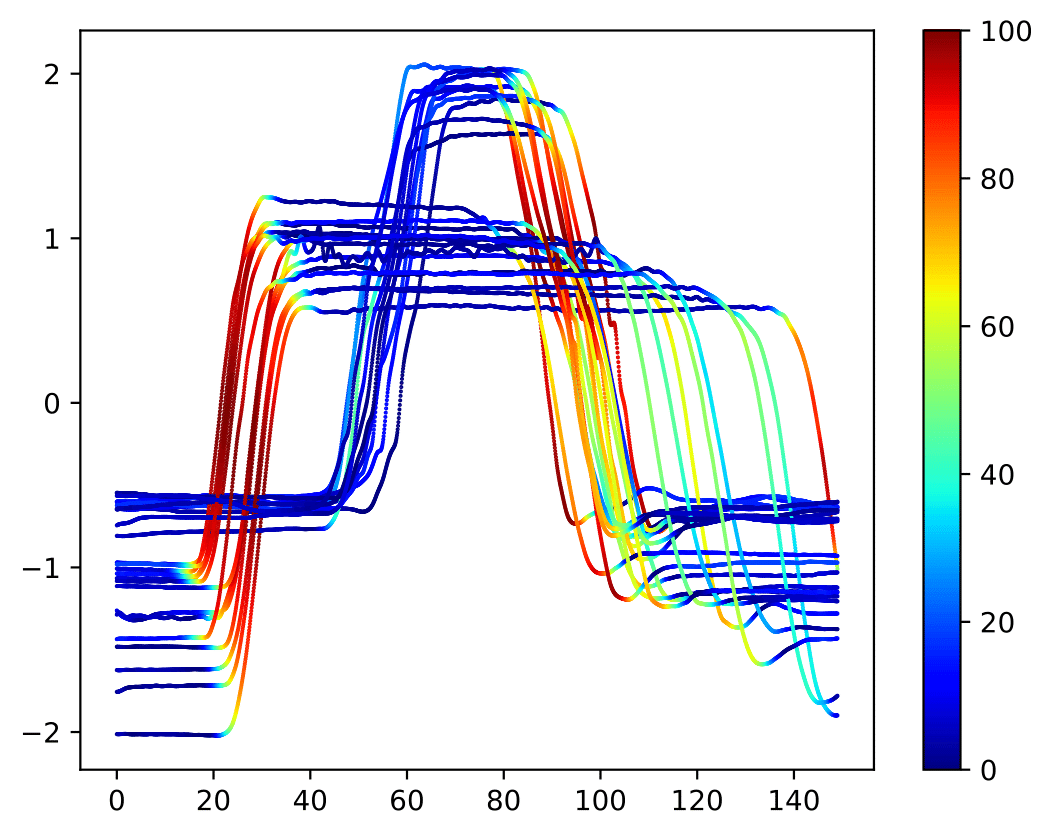}
      \label{sub-resnet-cam-GunPoint-class-2}}
    \caption{Highlighting with the Class Activation Map the contribution of each time series region for both classes in GunPoint when using the FCN and ResNet classifiers.
    Red corresponds to high contribution and blue to almost no contribution to the correct class identification (smoothed for visual clarity and best viewed in color) (Color figure online).}
    \label{fig-cam-gunpoint}
\usetikzlibrary{quotes,arrows.meta,positioning,decorations.pathreplacing,calc,3d,arrows}
\begin{tikzpicture}[overlay]
\node at (-6.25,16.5) (s1) {\small \textsf{discriminative}};
\node [below=-0.15 of s1] (s2) {\small \textsf{red bump}};
\node [below=0.2 of s1] (s3) {\small \textsf{detected}};
\draw [-latex,thick,below left=0 of s3]edge[color=black]+(-6.25,12.7)coordinate(x1);
\node at (4,12.2) (s1) {\small \textsf{non-discriminative}};
\node [below=-0.15 of s1] (s2) {\small \textsf{blue plateau}};
\node [below=0.2 of s1] (s3) {\small \textsf{detected}};
\draw [-latex,thick,above=0 of s1]edge[color=black]+(4,15)coordinate(x1);
\node at (-6.25,16.5-7.5) (s1) {\small \textsf{discriminative}};
\node [below=-0.15 of s1] (s2) {\small \textsf{red bump}};
\node [below=0.2 of s1] (s3) {\small \textsf{detected}};
\draw [-latex,thick,below left=0 of s3]edge[color=black]+(-6.25,12.7-7.5)coordinate(x1);
\node at (4,12.2-7.5) (s1) {\small \textsf{non-discriminative}};
\node [below=-0.15 of s1] (s2) {\small \textsf{blue plateau}};
\node [below=0.2 of s1] (s3) {\small \textsf{detected}};
\draw [-latex,thick,above=0 of s1]edge[color=black]+(4,15-7.5)coordinate(x1);
\end{tikzpicture}
\end{figure}
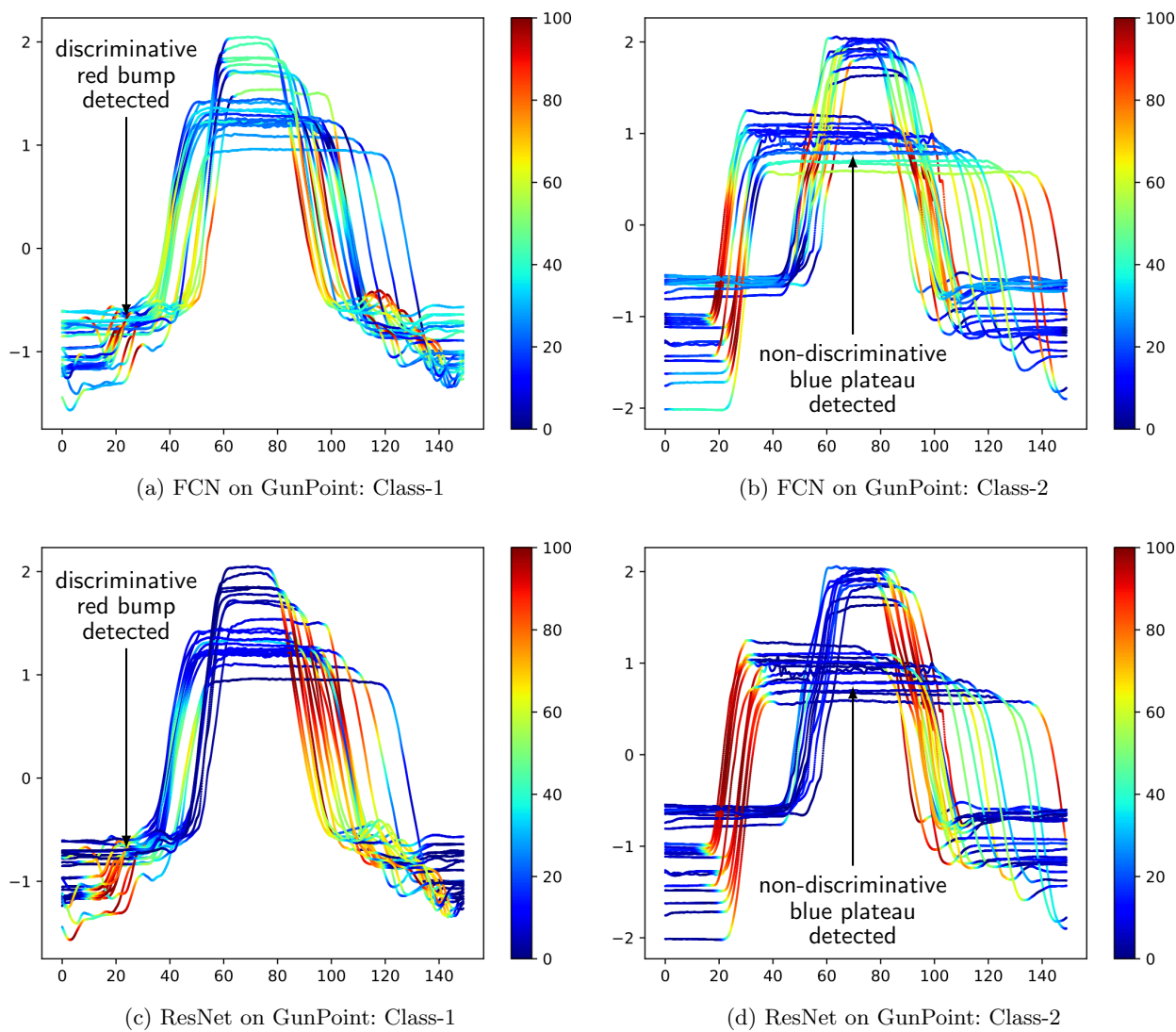 

The GunPoint dataset was first introduced by~\cite{ratanamahatana2005three} as a TSC problem. 
This dataset involves one male and one female actor performing two actions (Gun-Draw and Point) which makes it a binary classification problem.
For Gun-Draw (Class-1 in Figure~\ref{fig-cam-gunpoint}), the actors have first their hands by their sides, then draw a replicate gun from  hip-mounted holster, point it towards the target for one second, then finally place the gun in the holster and their hands to their initial position. 
Similarly to Gun-Draw, for Point (Class-2 in Figure~\ref{fig-cam-gunpoint}) the actors follow the same steps but instead of pointing a gun they point their index finger. 
For each task, the centroid of the actor's right hands on both $X$ and $Y$ axes were tracked and seemed to be very correlated, therefore the dataset contains only one univariate time series: the $X$-axis. 

We chose to start by visualizing the CAM for GunPoint for three main reasons. 
First, it is easy to visualize unlike other noisy datasets.
Second, both FCN and ResNet models achieved almost $100\%$ accuracy on this dataset which will help us to verify if both models are reaching the same decision for the same reasons.
Finally, it contains only two classes which allow us to analyze the data much more easily.  

Figure~\ref{fig-cam-gunpoint} shows the CAM's result when applied on each time series from both classes in the training set while classifying using the FCN model (Figure~\ref{sub-fcn-cam-GunPoint-class-1} and~\ref{sub-fcn-cam-GunPoint-class-2}) and the ResNet model (Figure~\ref{sub-resnet-cam-GunPoint-class-1} and~\ref{sub-resnet-cam-GunPoint-class-2}). 
At first glance, we can clearly see how both DNNs are neglecting the plateau non-discriminative regions of the time series when taking the classification decision. 
It is depicted by the blue flat parts of the time series which indicates no contribution to the classifier's decision. 
As for the highly discriminative regions (the red and yellow regions) both models were able to select the same parts of the time series which correspond to the points with high derivatives. 
Actually, the first most distinctive part of class-1 discovered by both classifiers is almost the same: the little red bump in the bottom left of Figure~\ref{sub-fcn-cam-GunPoint-class-1} and~\ref{sub-resnet-cam-GunPoint-class-1}.
Finally, another interesting observation is the ability of CNNs to localize a given discriminative shape regardless where it appears in the time series, which is evidence for CNNs' capability of learning time-invariant warped features. 

An interesting observation would be to compare the discriminative regions identified by a deep learning model with the most discriminative shapelets extracted by other shapelet-based approaches.
This observation would also be backed up by the mathematical proof provided by~\cite{cui2016multi}, that showed how the learned filters in a CNN can be considered a generic form of shapelets extracted by the learning shapelets algorithm~\citep{grabocka2014learning}.  
\cite{ye2011time} identified that the most important shapelet for the Gun/NoGun classification occurs when the actor's arm is lowered (about 120 on the horizontal axis in Figure~\ref{fig-cam-gunpoint}). 
\cite{hills2014classification} introduced a shapelet transformation based approach that discovered shapelets that are similar to the ones identified by~\cite{ye2011time}. 
For ResNet and FCN, the part where the actor lowers his/her arm (bottom right of Figure~\ref{fig-cam-gunpoint}) seems to be also identified as potential discriminative regions for some time series. 
On the other hand, the part where the actor raises his/her arm seems to be also a discriminative part of the data which suggests that the deep learning algorithms are identifying more ``shapelets''. 
We should note that this observation cannot confirm which classifier extracted the most discriminative subsequences especially because all algorithms achieved similar accuracy on GunPoint dataset. 
Perhaps a bigger dataset might provide a deeper insight into the interpretability of these machine learning models. 
Finally, we stress that the shapelet transformation classifier~\citep{hills2014classification} is an ensemble approach, which makes unclear how the shapelets affect the decision taken by the individual classifiers whereas for an end-to-end deep learning model we can directly explain the classification by using the Class Activation Map. 
\subsubsection{Meat dataset}
\begin{figure}
\centering
	\subfloat[FCN On Meat: Class-1]{
    \includegraphics[width=0.5\linewidth]{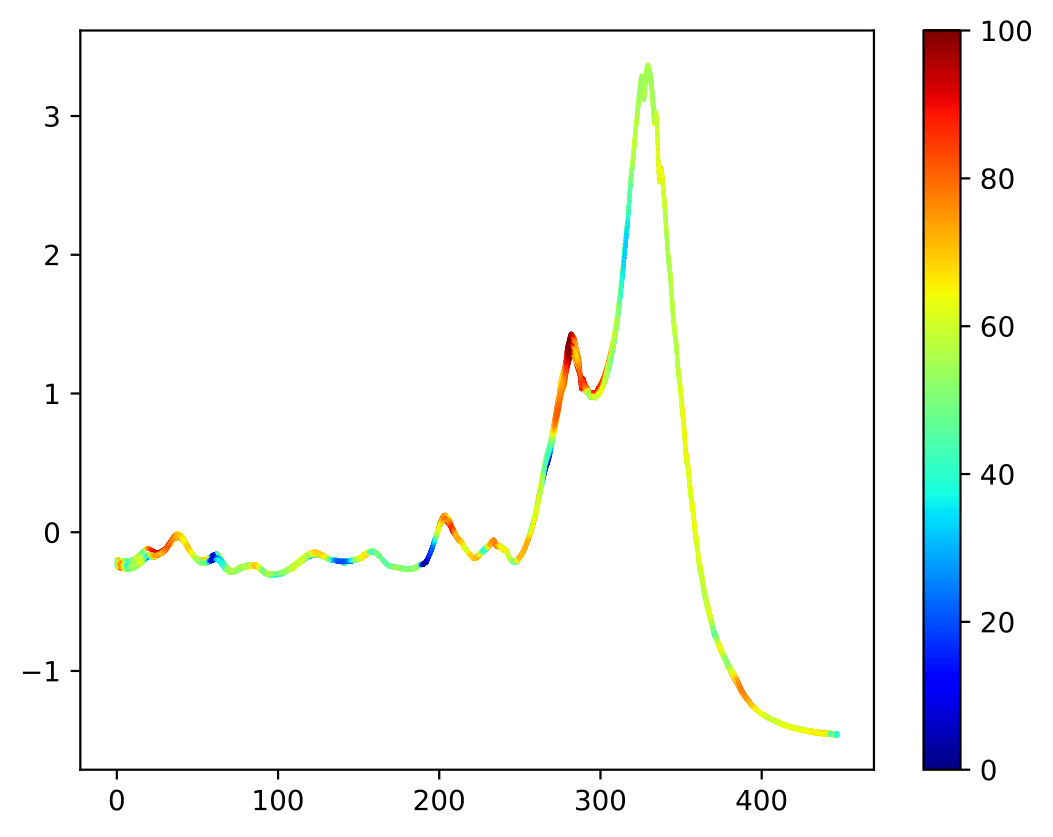}
    \label{sub-fcn-cam-Meat-class-1}}
    \subfloat[ResNet On Meat: Class-1]{
    \includegraphics[width=0.5\linewidth]{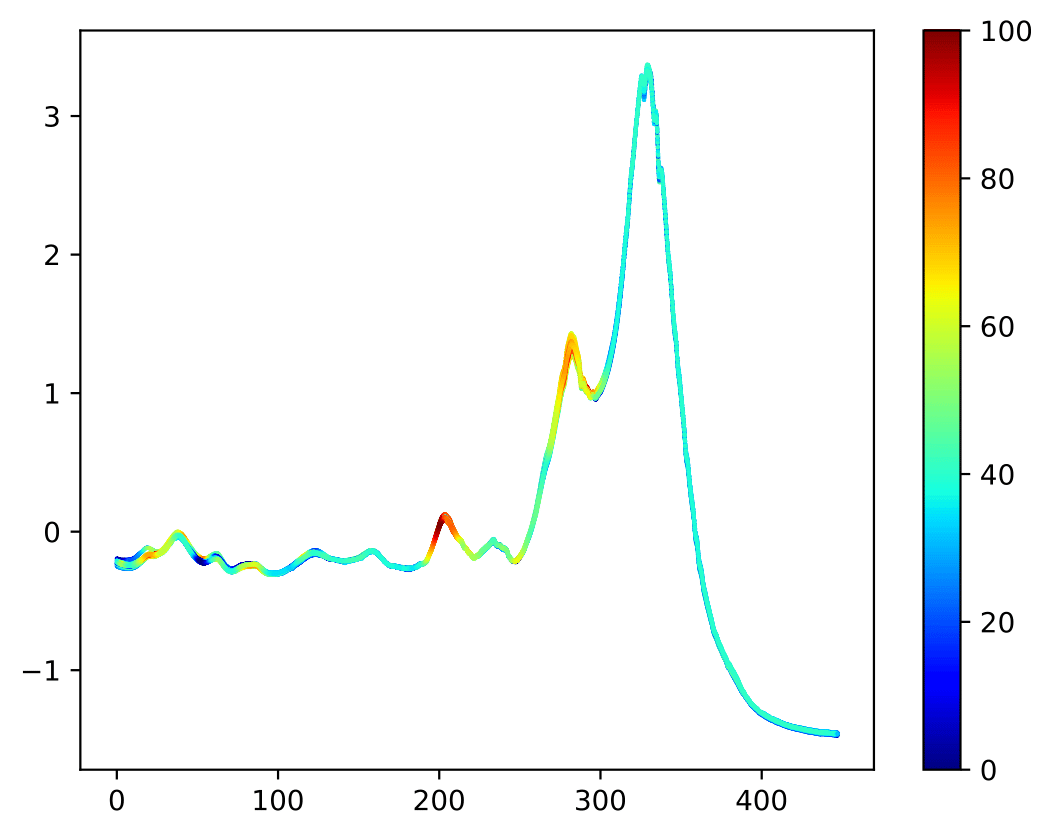}
    \label{sub-resnet-cam-Meat-class-1}}
    \\
    \subfloat[FCN On Meat: Class-2]{
    \includegraphics[width=0.5\linewidth]{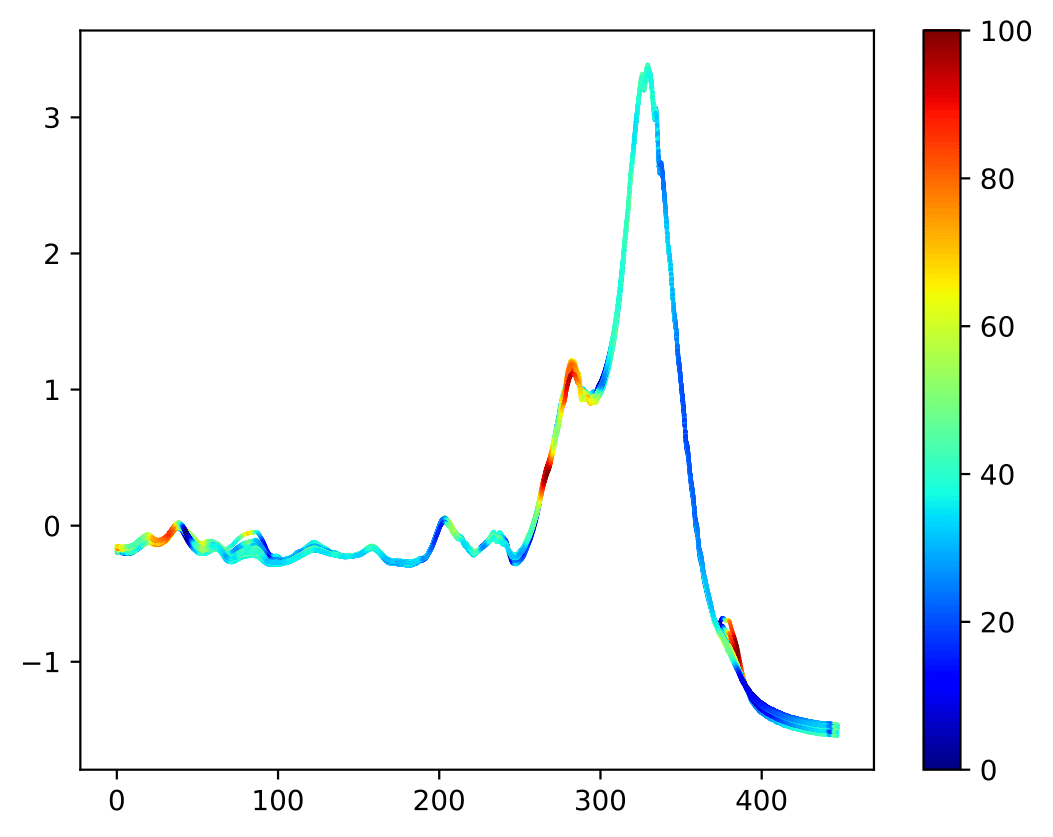}
    \label{sub-fcn-cam-Meat-class-2}}
    \subfloat[ResNet On Meat: Class-2]{
    \includegraphics[width=0.5\linewidth]{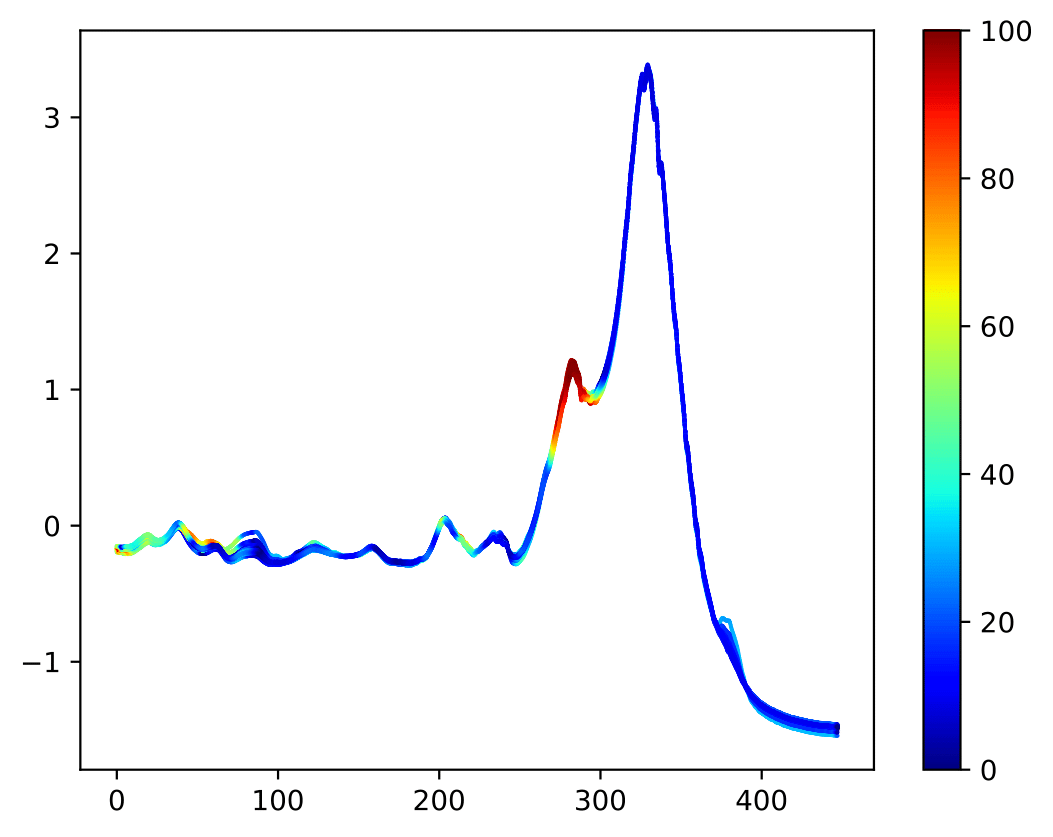}
    \label{sub-resnet-cam-Meat-class-2}}
    \\
    \subfloat[FCN On Meat: Class-3]{
    \includegraphics[width=0.5\linewidth]{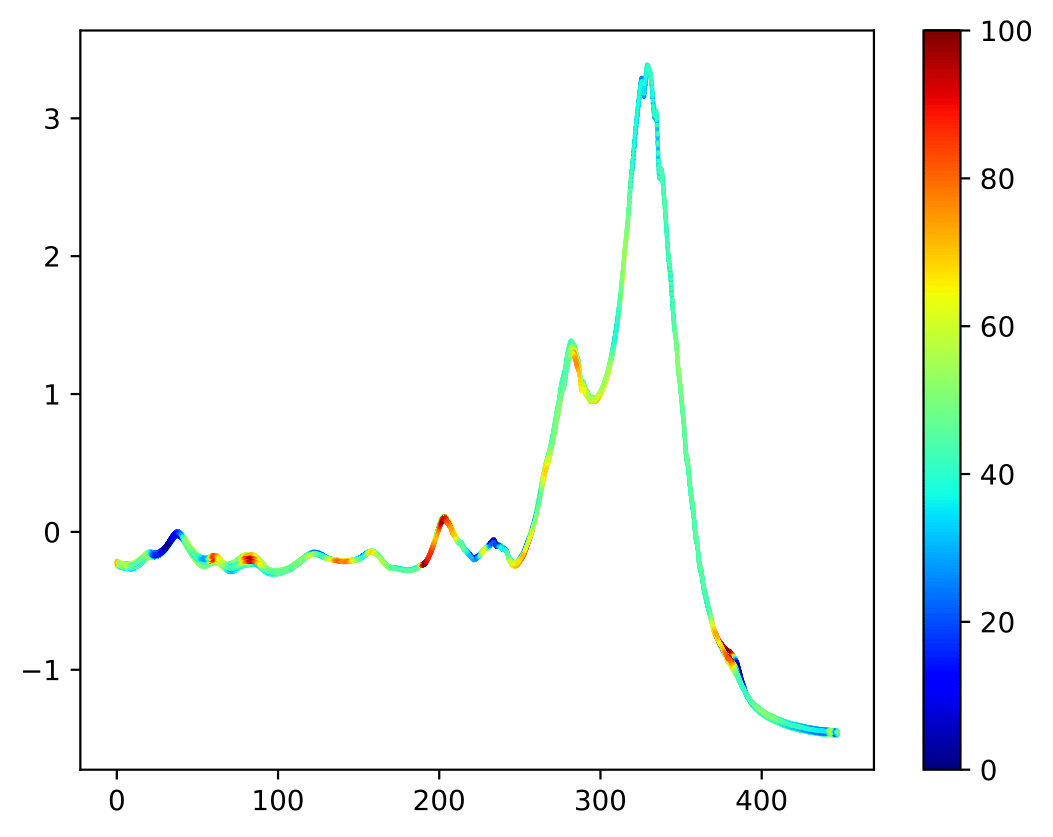}
    \label{sub-fcn-cam-Meat-class-3}}
    \subfloat[ResNet On Meat: Class-3]{
    \includegraphics[width=0.5\linewidth]{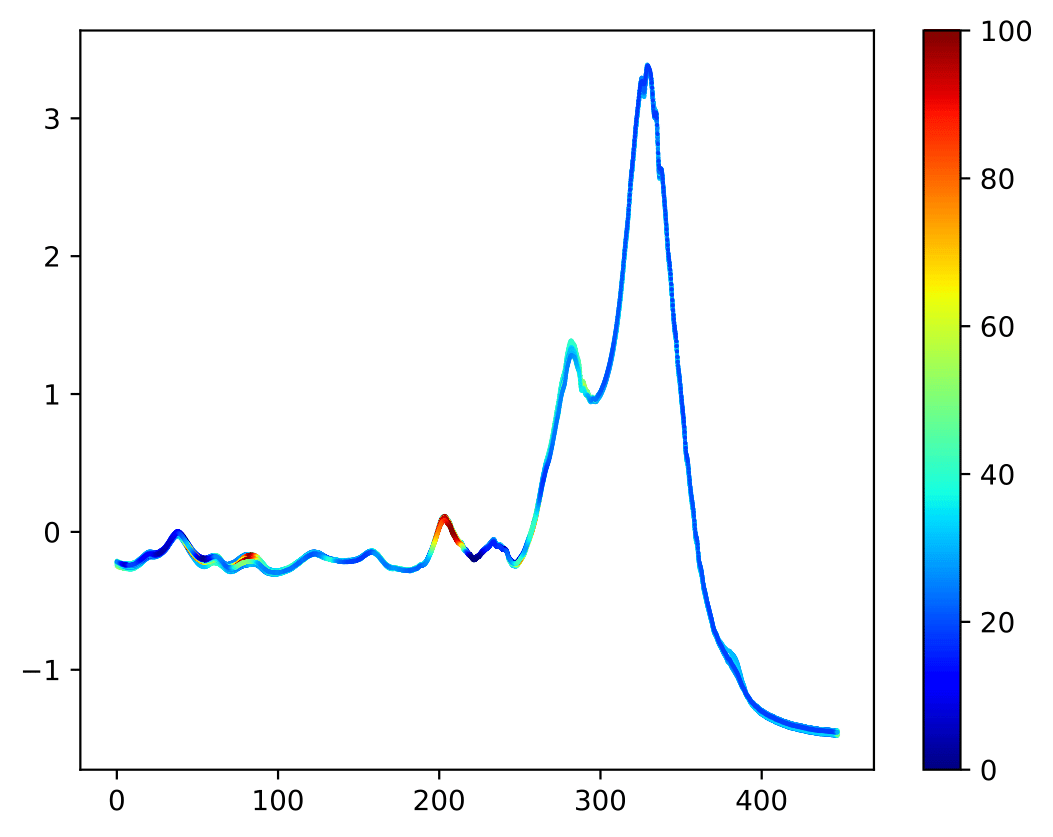}
    \label{sub-resnet-cam-Meat-class-3}}
    \caption{Highlighting with the Class Activation Map the contribution of each time series region for the three classes in Meat when using the FCN and ResNet classifiers.
    Red corresponds to high contribution and blue to almost no contribution to the correct class identification (smoothed for visual clarity and best viewed in color) (Color figure online).}
    \label{fig-cam-meat}
\begin{tikzpicture}[overlay]
\node at (2.3,24) (s11) {\small \textsf{discriminative}};
\node [below=-0.15 of s11] (s21) {\small  \textsf{red region}};
\node [below=0.2 of s11] (s31) {\small  \textsf{detected}};
\draw [-latex,thick,below right=0 of s31]edge[color=black]+(-3.55,21.95)coordinate(x1);
\draw [-latex,thick,below=0 of s31]edge[color=black]+(3.9,20.65)coordinate(x1);
\draw [-latex,thick,below left=0 of s31]edge[color=black]+(4.85,21.95)coordinate(x1);
\node at (2.3,15) (s1) {\small  \textsf{red region}};
\node [below=-0.15 of s1] (s2) {\small  \textsf{filtered-out}};
\node [below=0.2 of s1] (s3) {\small  \textsf{by ResNet}};
\draw [-latex,thick,below right=0 of s3]edge[color=black]+(-2.25,12.2)coordinate(x1);
\draw [-latex,thick,below left=0 of s3]edge[color=black]+(6,12.3)coordinate(x1);
\node at (2.7,7.7) (s12) {\small  \textsf{non-discriminative}};
\node [below=-0.15 of s12] (s22) {\small  \textsf{blue peak}};
\node [below=0.2 of s12] (s32) {\small  \textsf{detected}};
\draw [-latex,thick,above right=0 of s12]edge[color=black]+(5.3,9.1)coordinate(x1);
\draw [-latex,thick,above left=0 of s12]edge[color=black]+(-2.9,9.1)coordinate(x1);
\end{tikzpicture}
\end{figure}
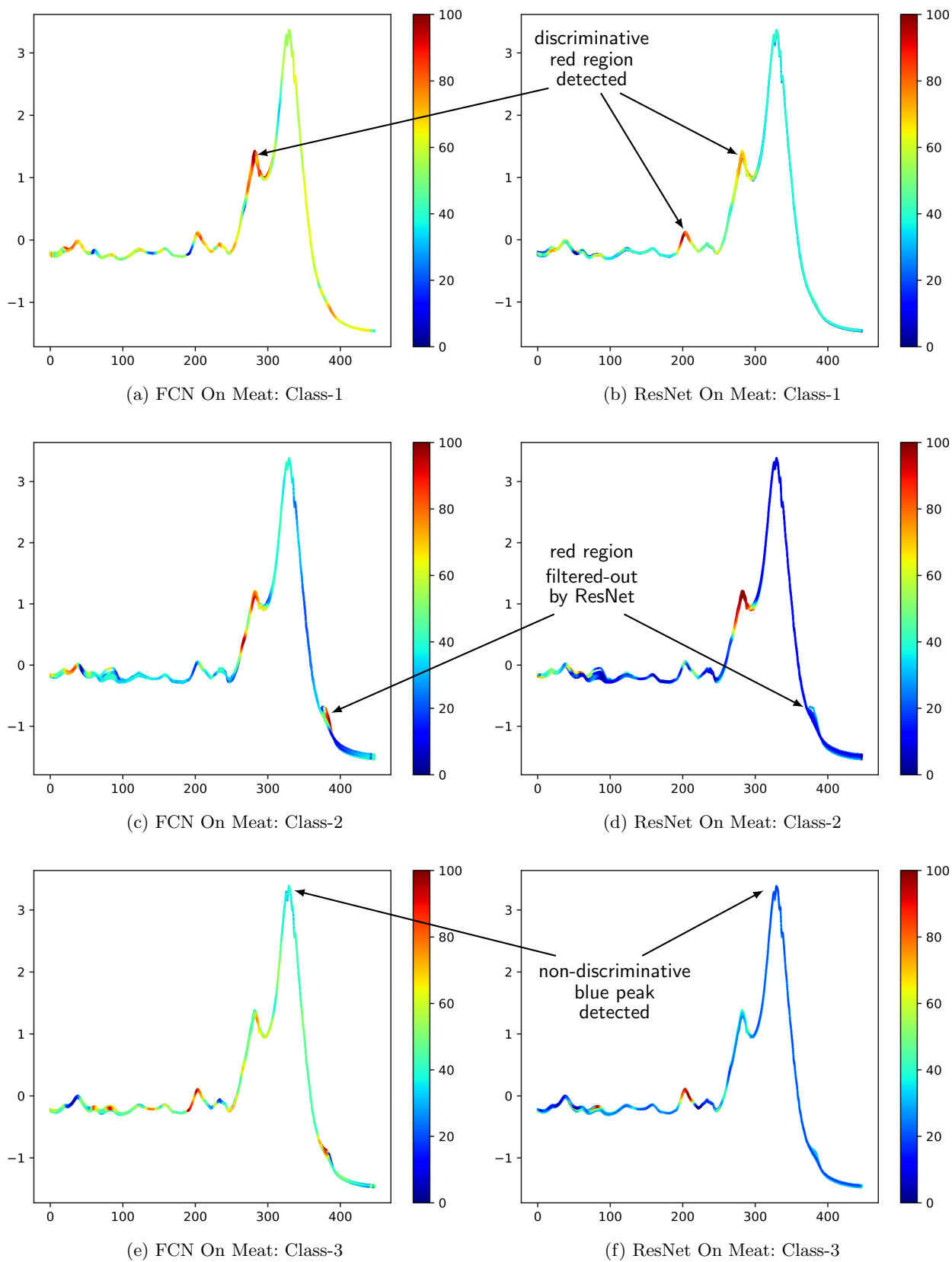 

Although the previous case study on GunPoint yielded interesting results in terms of showing that both models are localizing meaningful features, it failed to show the difference between the two most accurate deep learning classifiers: ResNet and FCN.
Therefore we decided to further analyze the CAM's result for the two models on the Meat dataset.

Meat is a food spectrograph dataset which are usually used in chemometrics to classify food types, a task that has obvious applications in food safety and quality assurance.
There are three classes in this dataset: Chicken, Pork and Turkey corresponding respectively to classes 1, 2 and 3 in Figure~\ref{fig-cam-meat}.  
\cite{alJowder1997mid} described how the data is acquired from 60 independent samples using Fourier transform infrared (FTIR) spectroscopy with attenuated total reflectance (ATR) sampling. 

Similarly to GunPoint, this dataset is easy to visualize and does not contain very noisy time series. 
In addition, with only three classes, the visualization is possible to understand and analyze.
Finally, unlike for the GunPoint dataset, the two approaches ResNet and FCN reached significantly different results on Meat with respectively $97\%$ and $83\%$ accuracy.

Figure~\ref{fig-cam-meat} enables the comparison between FCN's CAM (left) and ResNet's  CAM (right).  
We first observe that ResNet is much more firm when it comes to highlighting the regions. 
In other words, FCN's CAM contains much more smoother regions with cyan, green and yellow regions, whereas ResNet's CAM contains more dark red and blue subsequences showing that ResNet can filter out non-discriminative and discriminative regions with a higher confidence than FCN, which probably explains why FCN is less accurate than ResNet on this dataset. 
Another interesting observation is related to the red subsequence highlighted by FCN's CAM for class 2 and 3 at the bottom right of Figure~\ref{sub-fcn-cam-Meat-class-2} and~\ref{sub-fcn-cam-Meat-class-3}.
By visually investigating this part of the time series, we clearly see that it is a non-discriminative part since the time series of both classes exhibit this bump. 
This subsequence is therefore filtered-out by the ResNet model which can be seen by the blue color in the bottom right of Figure~\ref{sub-resnet-cam-Meat-class-2} and~\ref{sub-resnet-cam-Meat-class-3}. 
These results suggest that ResNet's superiority over FCN is mainly due to the former's  ability to filter-out non-distinctive regions of the time series. 
We attribute this ability to the main characteristic of ResNet which is composed of the residual connections between the convolutional blocks that enable the model to \emph{learn to skip} unnecessary convolutions by dint of its shortcut links ~\citep{he2016deep}.  

\subsection{Multi-Dimensional Scaling}
We propose the use of Multi-Dimensional Scaling (MDS)~\citep{kruskal1978multidimensional} with the objective to gain some insights on the spatial distribution of the input time series belonging to different classes in the dataset.
MDS uses a pairwise distance matrix as input and aims at placing each object in a N-dimensional space such as the between-object distances are preserved as well as possible. 
Using the Euclidean Distance (ED) on a set of input time series belonging to the test set, it is then possible to create a similarity matrix and apply MDS to display the set into a two dimensional space. 
This straightforward approach supposes that the ED is able to strongly separate the raw time series, which is usually not the case evident by the low accuracy of the nearest neighbor when coupled with ED~\citep{bagnall2017the}.

On the other hand, we propose to apply this MDS method to visualize the set of time series with its latent representation learned by the network. 
Usually in a deep neural network, we have several hidden layers and one can find several latent representation of the dataset. 
But since we are aiming at visualizing the class specific latent space, we chose to use the last latent representation of a DNN (the one directly before the softmax classifier), which is known to be a class specific layer~\citep{yosinski2014how}.  
We decided to apply this method only on ResNet and FCN for two reasons: (1) when evaluated on the UCR/UEA archive they reached the highest ranks; (2) they both employ a GAP layer before the softmax layer making the number of latent features invariant to the time series length. 

To better explain this process, for each input time series, the last convolution (for ResNet and FCN) outputs a multivariate time series whose dimensions are equal to the number of filters (128) in the last convolution, then the GAP layer averages the latter 128-dimensional  multivariate time series over the time dimension resulting in a vector of 128 real values over which the ED is computed. 
As we worked with the ED, we used metric MDS~\citep{kruskal1978multidimensional} that minimizes a cost function called \emph{Stress} which is a residual sum of squares:
\begin{equation}
    Stress_D(X_1,\ldots,X_N)=\Biggl(\frac{\sum_{i,j}\bigl(d_{ij}-\|x_i-x_j\|\bigr)^2}{\sum_{i,j}d_{ij}^2}\Biggr)^{1/2}
\end{equation}
where $d_{ij}$ is the ED between the GAP vectors of time series $X_i$ and $X_j$.
Obviously, one has to be careful about the interpretation of MDS output, as the data space is highly simplified (each time series $X_i$ is represented as a single data point $x_i$).

\begin{figure}
\centering
    \subfloat[GunPoint-MDS-Raw]{
 \includegraphics[width=0.5\linewidth]{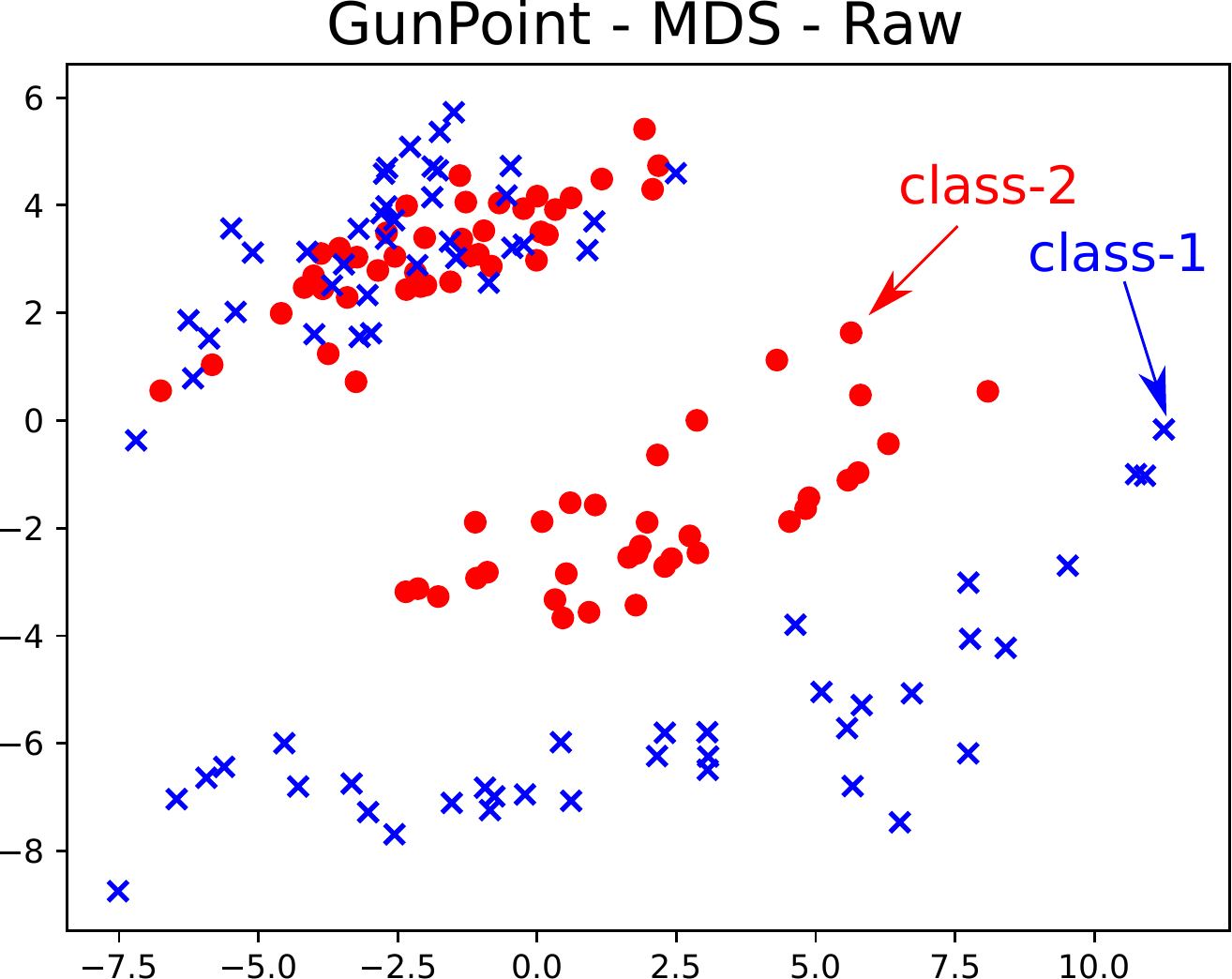}
      \label{sub-mds-gunpoint-raw}}\\
    \subfloat[GunPoint-MDS-GAP-FCN]{
 \includegraphics[width=0.5\linewidth]{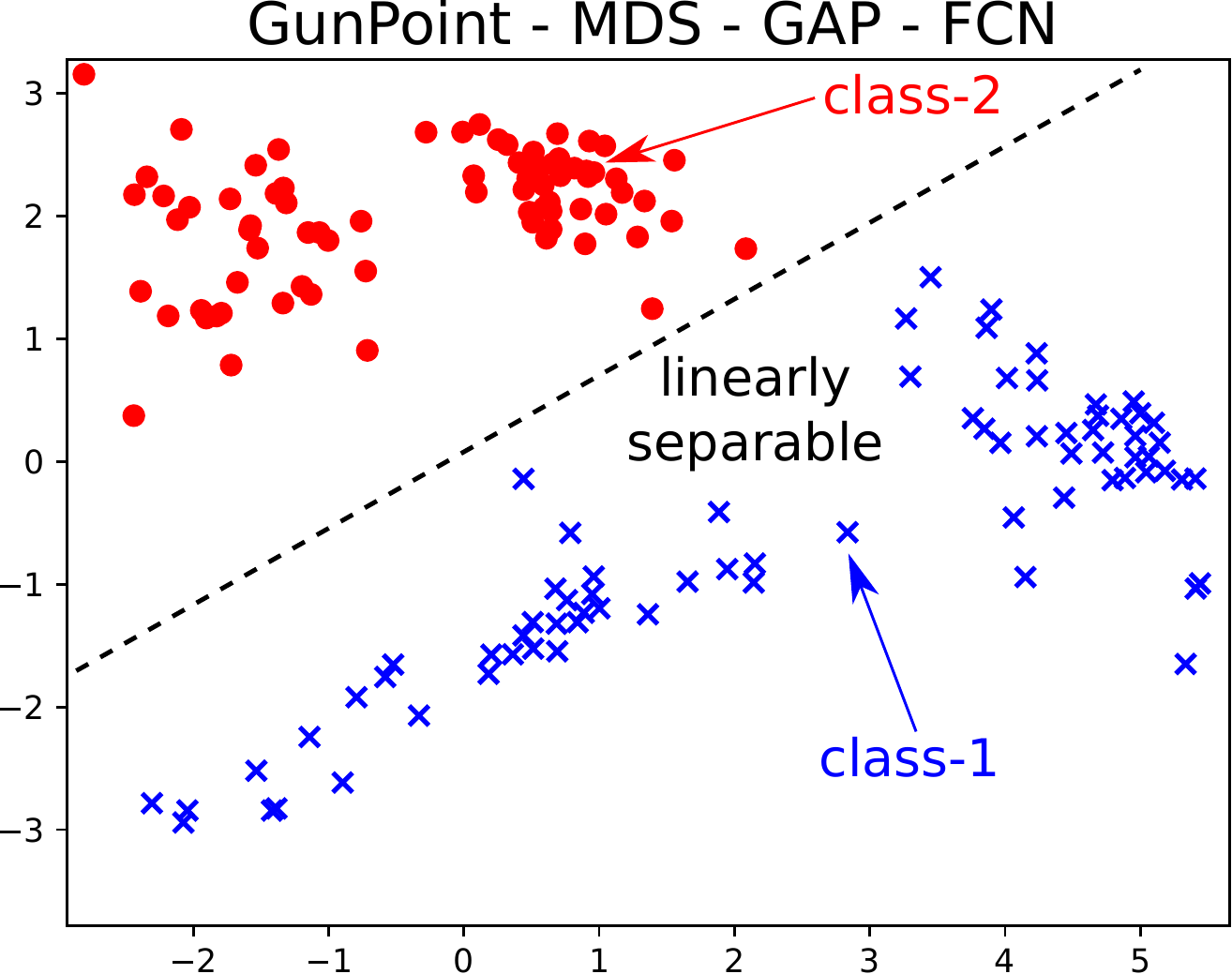}
      \label{sub-mds-gunpoint-gap-fcn}}
      \subfloat[GunPoint-MDS-GAP-ResNet]{
 \includegraphics[width=0.5\linewidth]{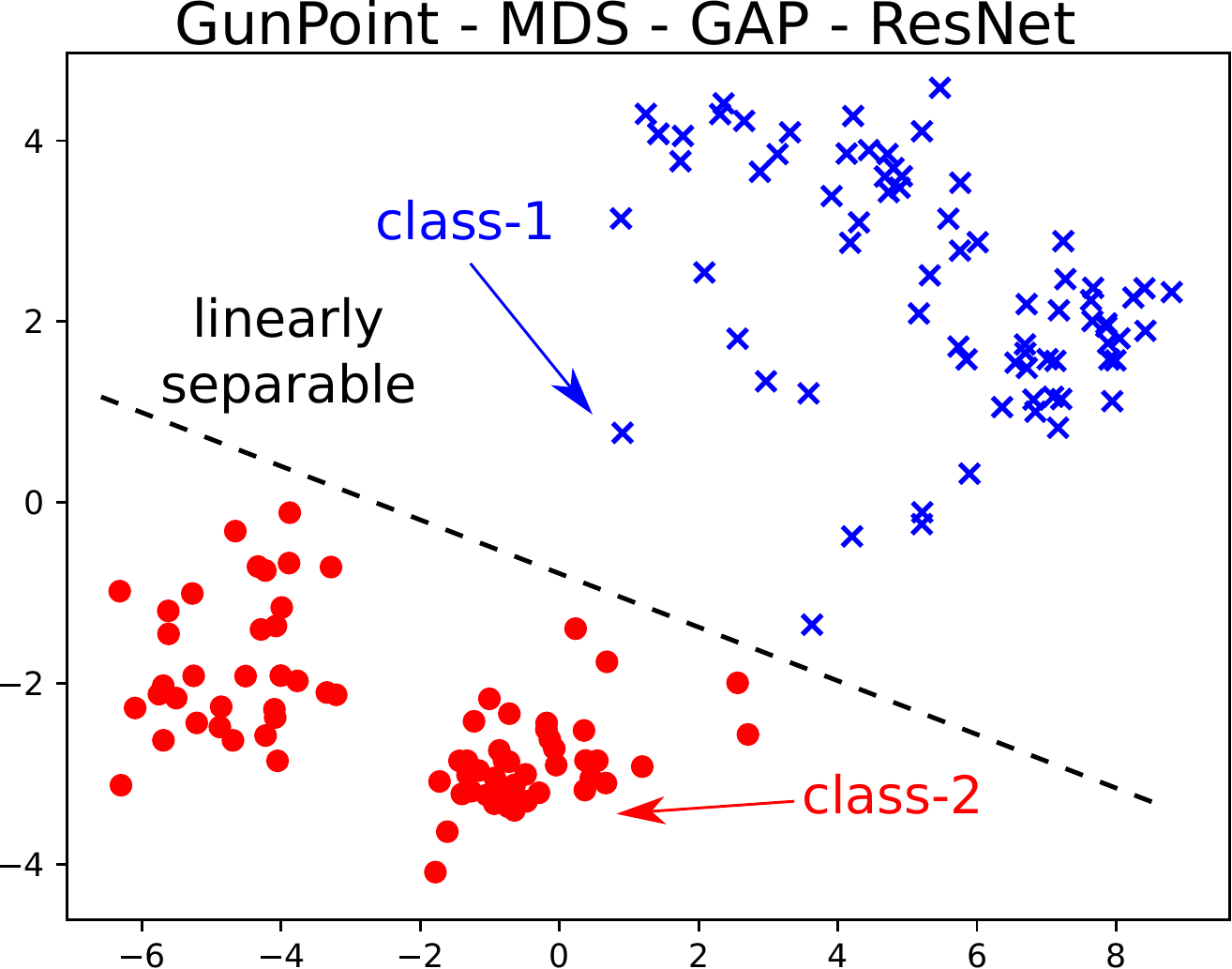}
      \label{sub-mds-gunpoint-gap-resnet}}
    \caption{Multi-Dimensional Scaling (MDS) applied on GunPoint for: (top) the raw input time series; (bottom) the learned features from the Global Average Pooling (GAP) layer for FCN (left) and ResNet (right) - (best viewed in color).
    This figure shows how the ResNet and FCN are projecting the time series from a non-linearly separable 2D space (when using the raw input), into a linearly separable 2D space (when using the latent representation) (Color figure online).}
    \label{fig-mds-gunpoint}
\end{figure} 

Figure~\ref{fig-mds-gunpoint} shows three MDS plots for the GunPoint dataset using: (1) the raw input time series (Figure~\ref{sub-mds-gunpoint-raw}); (2) the learned latent features from the GAP layer for FCN (Figure~\ref{sub-mds-gunpoint-gap-fcn}); and (3) the learned latent features from the GAP layer for ResNet (Figure~\ref{sub-mds-gunpoint-gap-resnet}). 
We can easily observe in Figure~\ref{sub-mds-gunpoint-raw} that when using the raw input data and projecting it into a 2D space, the two classes are not linearly separable.  
On the other hand, in both Figures~\ref{sub-mds-gunpoint-gap-fcn} and~\ref{sub-mds-gunpoint-gap-resnet}, by applying MDS on the latent representation learned by the network, one can easily separate the set of time series belonging to the two classes. 
We note that both deep learning models (FCN and ResNet) managed to project the data from GunPoint into a linearly separable space which explains why both models performed equally very well on this dataset with almost 100\% accuracy. 

\begin{figure}
\centering
    \subfloat[Wine-MDS-Raw]{
 \includegraphics[width=0.5\linewidth]{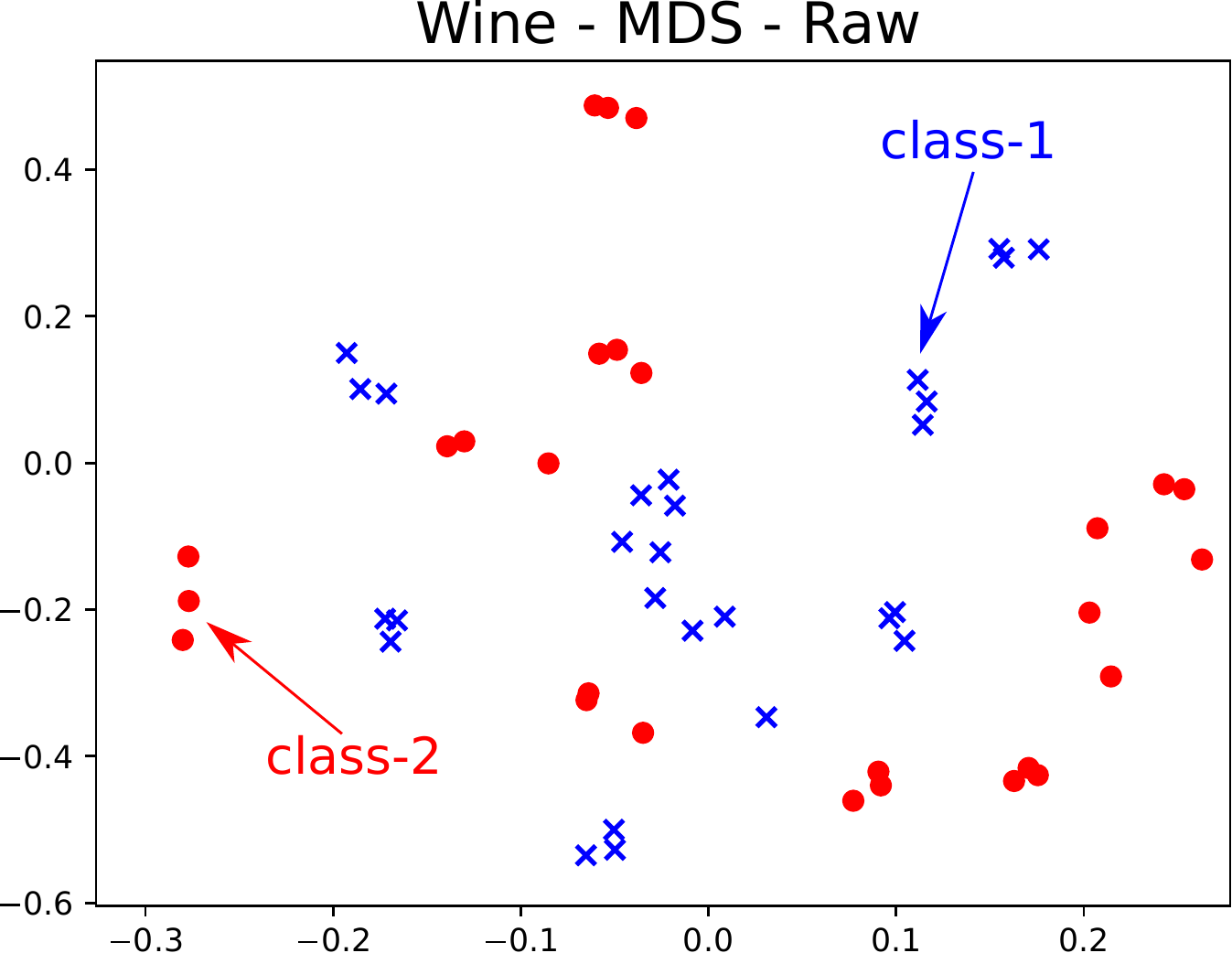}
      \label{sub-mds-wine-raw}}\\
    \subfloat[Wine-MDS-GAP-FCN]{
 \includegraphics[width=0.5\linewidth]{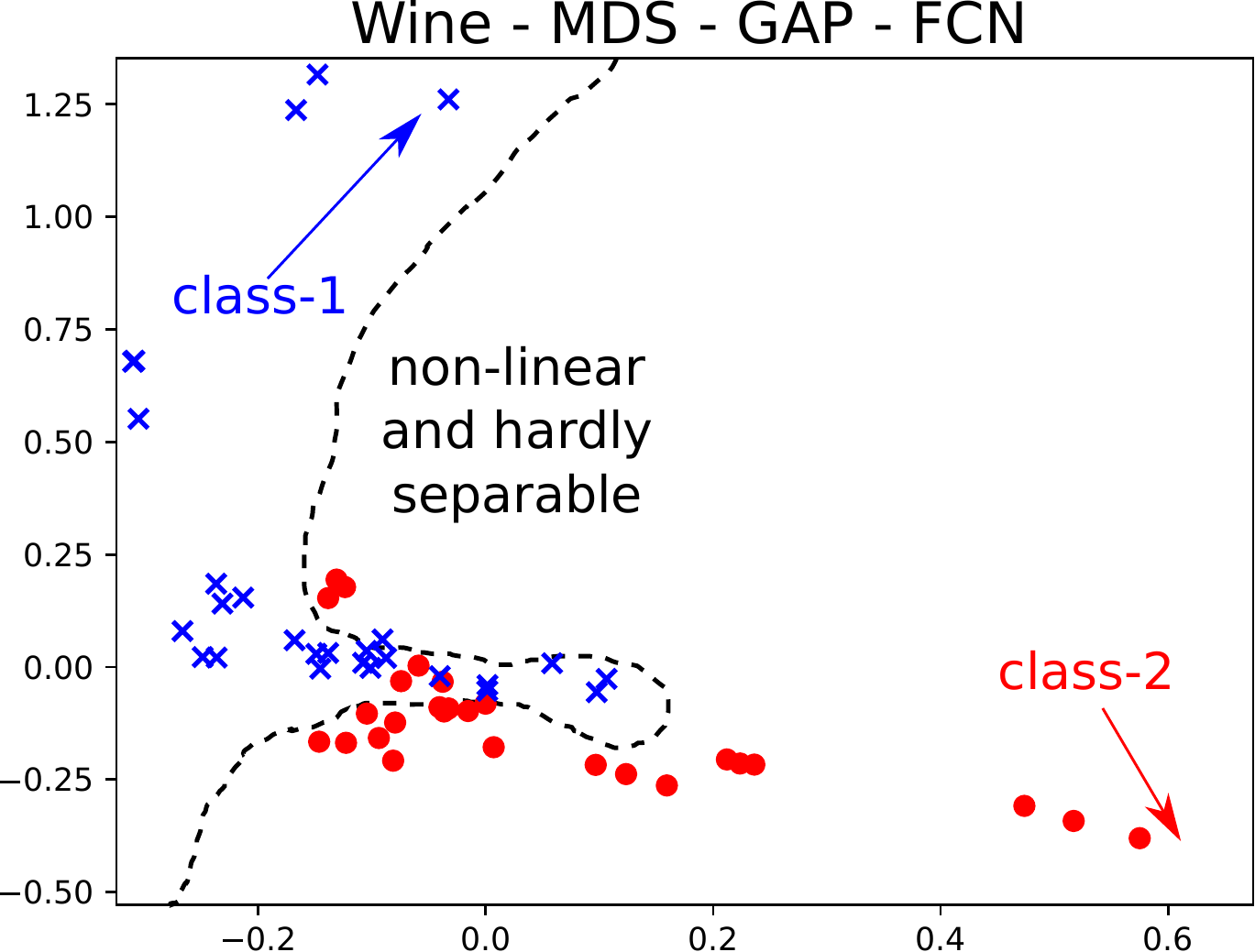}
      \label{sub-mds-wine-gap-fcn}}
      \subfloat[Wine-MDS-GAP-ResNet]{
 \includegraphics[width=0.5\linewidth]{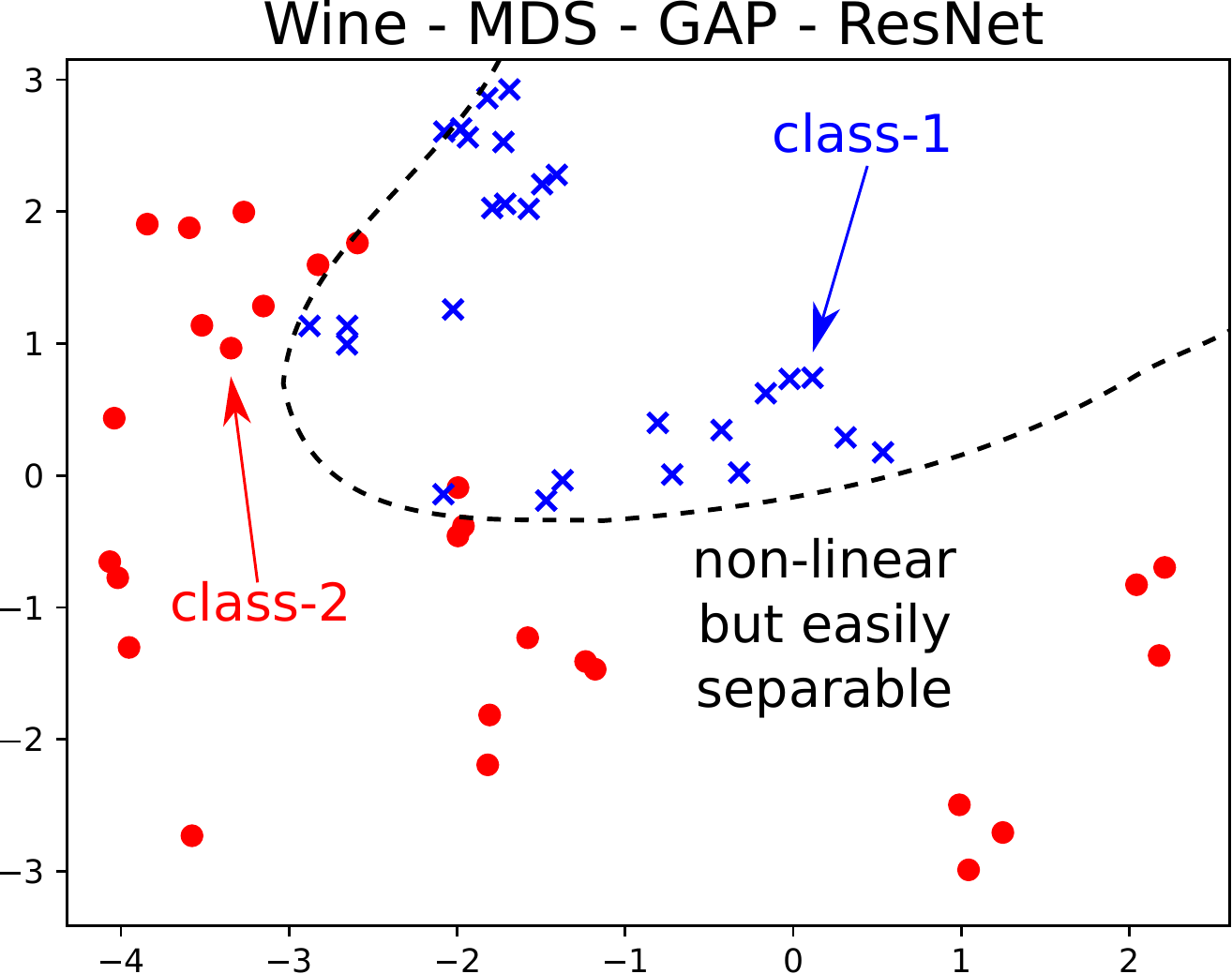}
      \label{sub-mds-wine-gap-resnet}}
    \caption{Multi-Dimensional Scaling (MDS) applied on Wine for: (top) the raw input time series; (bottom) the learned features from the Global Average Pooling (GAP) layer for FCN (left) and ResNet (right) - (best viewed in color).
    This figure shows how ResNet, unlike FCN, is able to project the data into an easily separable space when using the learned features from the GAP layer (Color figure online).}
    \label{fig-mds-wine}
\end{figure} 

Although the visualization of MDS on GunPoint yielded some interesting results, it failed to pinpoint the difference between the two deep learning models FCN and ResNet.
Therefore we decided to analyze another dataset where the accuracy of both models differed by almost 15\%. 
Figure~\ref{fig-mds-wine} shows three MDS plots for the Wine dataset using: (1) the raw input time series (Figure~\ref{sub-mds-wine-raw}); (2) the learned latent features from the GAP layer for FCN (Figure~\ref{sub-mds-wine-gap-fcn}); and (3) the learned latent features from the GAP layer for ResNet (Figure~\ref{sub-mds-wine-gap-resnet}). 
At first glimpse of Figure~\ref{fig-mds-wine}, the reader can conclude that all projections, even when using the learned representation, are not linearly separable which is evident by the relatively low accuracy of both models FCN and ResNet which is equal respectively to 58.7\% and 74.4\%. 
A thorough observation shows us that the learned hidden representation of ResNet (Figure~\ref{sub-mds-wine-gap-resnet}) separates the data from both classes in a much clearer way than the FCN (Figure~\ref{sub-mds-wine-gap-fcn}). 
In other words, FCN's learned representation has too many data points close to the decision boundary whereas ResNet's hidden features enables projecting data points further away from the decision boundary.   
This observation could explain why ResNet achieves a better performance than FCN on the Wine dataset.

\section{Conclusion}\label{sec-conclusion}
In this paper, we presented the largest empirical study of DNNs for TSC.
We described the most recent successful deep learning approaches for TSC in many different domains such as human activity recognition and sleep stage identification. 
Under a unified taxonomy, we explained how DNNs are separated into two main categories of generative and discriminative models.
We re-implemented nine recently published end-to-end deep learning classifiers in a unique framework which we make publicly available to the community. 
Our results show that end-to-end deep learning can achieve the current state-of-the-art performance for TSC with architectures such as Fully Convolutional Neural Networks and deep Residual Networks. 
Finally, we showed how the black-box effect of deep models which renders them uninterpretable, can be mitigated with a Class Activation Map visualization that highlights which parts of the input time series, contributed the most to a certain class identification. 

Although we have conducted an extensive experimental evaluation, deep learning for time series classification, unlike for computer vision and NLP tasks, still lacks a thorough study of data augmentation~\citep{ismailFawaz2018data,forestier2017generating} and transfer learning~\citep{IsmailFawaz2018transfer,serra2018towards}.
In addition, the time series community would benefit from an extension of this empirical study that compares in addition to accuracy, the training and testing time of these deep learning models. 
Furthermore, we think that the effect of z-normalization (and other normalization methods) on the learning capabilities of DNNs should also be thoroughly explored. 
In our future work, we aim to investigate and answer the aforementioned limitations by conducting more extensive experiments especially on multivariate time series datasets.
In order to achieve all of these goals, one important challenge for the TSC community is to provide one large generic \emph{labeled} dataset similar to the large images database in computer vision such as ImageNet~\citep{russakovsky2015imagenet} that contains 1000 classes.

In conclusion, with data mining repositories becoming more frequent, leveraging deeper architectures that can learn automatically from annotated data in an end-to-end fashion, makes deep learning a very enticing approach.

\begin{acknowledgements}
The authors would like to thank the creators and providers of the datasets: Hoang Anh Dau, Anthony Bagnall, Kaveh Kamgar, Chin-Chia Michael Yeh, Yan Zhu, Shaghayegh Gharghabi, Chotirat Ann Ratanamahatana, Eamonn Keogh and Mustafa Baydogan.
The authors would also like to thank NVIDIA Corporation for the GPU Grant and the M\'esocentre of Strasbourg for providing access to the cluster.
The authors would also like to thank Fran\c{c}ois Petitjean and Charlotte Pelletier for the fruitful discussions, their feedback and comments while writing this paper.
This work was supported by the ANR TIMES project (grant ANR-17-CE23-0015) of the French Agence Nationale de la Recherche.
\end{acknowledgements}

\bibliographystyle{spbasic}      
\bibliography{biblio}

\begin{thebibliography}{138}
\providecommand{\natexlab}[1]{#1}
\providecommand{\url}[1]{{#1}}
\providecommand{\urlprefix}{URL }
\expandafter\ifx\csname urlstyle\endcsname\relax
  \providecommand{\doi}[1]{DOI~\discretionary{}{}{}#1}\else
  \providecommand{\doi}{DOI~\discretionary{}{}{}\begingroup
  \urlstyle{rm}\Url}\fi
\providecommand{\eprint}[2][]{\url{#2}}

\bibitem[{Abadi et~al.(2015)Abadi, Agarwal, Barham, Brevdo, Chen, Citro,
  Corrado, Davis, Dean, Devin, Ghemawat, Goodfellow, Harp, Irving, Isard, Jia,
  Jozefowicz, Kaiser, Kudlur, Levenberg, Man\'{e}, Monga, Moore, Murray, Olah,
  Schuster, Shlens, Steiner, Sutskever, Talwar, Tucker, Vanhoucke, Vasudevan,
  Vi\'{e}gas, Vinyals, Warden, Wattenberg, Wicke, Yu, and
  Zheng}]{tensorflow2015whitepaper}
Abadi M, Agarwal A, Barham P, Brevdo E, Chen Z, Citro C, Corrado GS, Davis A,
  Dean J, Devin M, Ghemawat S, Goodfellow I, Harp A, Irving G, Isard M, Jia Y,
  Jozefowicz R, Kaiser L, Kudlur M, Levenberg J, Man\'{e} D, Monga R, Moore S,
  Murray D, Olah C, Schuster M, Shlens J, Steiner B, Sutskever I, Talwar K,
  Tucker P, Vanhoucke V, Vasudevan V, Vi\'{e}gas F, Vinyals O, Warden P,
  Wattenberg M, Wicke M, Yu Y, Zheng X (2015) {TensorFlow}: Large-scale machine
  learning on heterogeneous systems.
  \urlprefix\url{https://www.tensorflow.org/}, {A}ccessed 28 Feb 2019

\bibitem[{Al-Jowder et~al.(1997)Al-Jowder, Kemsley, and
  Wilson}]{alJowder1997mid}
Al-Jowder O, Kemsley E, Wilson R (1997) Mid-infrared spectroscopy and
  authenticity problems in selected meats: a feasibility study. Food Chemistry
  59(2):195 -- 201

\bibitem[{Aswolinskiy et~al.(2016)Aswolinskiy, Reinhart, and
  Steil}]{aswolinskiy2016time}
Aswolinskiy W, Reinhart RF, Steil J (2016) Time series classification in
  reservoir- and model-space: A comparison. In: Artificial Neural Networks in
  Pattern Recognition, pp 197--208

\bibitem[{Aswolinskiy et~al.(2017)Aswolinskiy, Reinhart, and
  Steil}]{aswolinskiy2017time}
Aswolinskiy W, Reinhart RF, Steil J (2017) Time series classification in
  reservoir- and model-space. Neural Processing Letters 48:789--809

\bibitem[{Bagnall and Janacek(2014)}]{bagnall2014a}
Bagnall A, Janacek G (2014) A run length transformation for discriminating
  between auto regressive time series. Journal of Classification 31(2):154--178

\bibitem[{Bagnall et~al.(2016)Bagnall, Lines, Hills, and
  Bostrom}]{bagnall2016time}
Bagnall A, Lines J, Hills J, Bostrom A (2016) Time-series classification with
  {COTE}: The collective of transformation-based ensembles. In: International
  Conference on Data Engineering, pp 1548--1549

\bibitem[{Bagnall et~al.(2017)Bagnall, Lines, Bostrom, Large, and
  Keogh}]{bagnall2017the}
Bagnall A, Lines J, Bostrom A, Large J, Keogh E (2017) The great time series
  classification bake off: a review and experimental evaluation of recent
  algorithmic advances. Data Mining and Knowledge Discovery 31(3):606--660

\bibitem[{{Bahdanau} et~al.(2015){Bahdanau}, {Cho}, and
  {Bengio}}]{bahdanau2015neural}
{Bahdanau} D, {Cho} K, {Bengio} Y (2015) {Neural Machine Translation by Jointly
  Learning to Align and Translate}. In: International Conference on Learning
  Representations

\bibitem[{Baird(1992)}]{baird1992document}
Baird HS (1992) Document Image Defect Models, Springer, Berlin, pp 546--556

\bibitem[{Banerjee et~al.(2017)Banerjee, Islam, Mei, Xiao, Zhang, Xu, Ji, and
  Li}]{banerjee2017a}
Banerjee D, Islam K, Mei G, Xiao L, Zhang G, Xu R, Ji S, Li J (2017) A deep
  transfer learning approach for improved post-traumatic stress disorder
  diagnosis. In: IEEE International Conference on Data Mining, pp 11--20

\bibitem[{Baydogan(2015)}]{baydogan2015mts}
Baydogan MG (2015) Multivariate time series classification datasets.
  \url{http://www.mustafabaydogan.com}, {A}ccessed 28 Feb 2019

\bibitem[{Baydogan et~al.(2013)Baydogan, Runger, and Tuv}]{baydogan2013a}
Baydogan MG, Runger G, Tuv E (2013) A bag-of-features framework to classify
  time series. IEEE Transactions on Pattern Analysis and Machine Intelligence
  35(11):2796--2802

\bibitem[{Bellman(2010)}]{bellman2010dynamic}
Bellman R (2010) Dynamic Programming. Princeton University Press

\bibitem[{Benavoli et~al.(2016)Benavoli, Corani, and
  Mangili}]{benavoli2016should}
Benavoli A, Corani G, Mangili F (2016) Should we really use post-hoc tests
  based on mean-ranks? Machine Learning Research 17(1):152--161

\bibitem[{Bengio et~al.(2013)Bengio, Yao, Alain, and
  Vincent}]{bengio2013generalized}
Bengio Y, Yao L, Alain G, Vincent P (2013) Generalized denoising auto-encoders
  as generative models. In: International Conference on Neural Information
  Processing Systems, pp 899--907

\bibitem[{{Bianchi} et~al.(2018){Bianchi}, {Scardapane}, {L{\o}kse}, and
  {Jenssen}}]{bianchi2018reservoir}
{Bianchi} FM, {Scardapane} S, {L{\o}kse} S, {Jenssen} R (2018) {Reservoir
  computing approaches for representation and classification of multivariate
  time series}. ArXiv \eprint{1803.07870}

\bibitem[{Bishop(2006)}]{bishop2006pattern}
Bishop C (2006) Pattern Recognition and Machine Learning. Springer

\bibitem[{Bostrom and Bagnall(2015)}]{bostrom2015binary}
Bostrom A, Bagnall A (2015) Binary shapelet transform for multiclass time
  series classification. In: Big Data Analytics and Knowledge Discovery, pp
  257--269

\bibitem[{Che et~al.(2017{\natexlab{a}})Che, Cheng, Zhai, Sun, and
  Liu}]{che2017boosting}
Che Z, Cheng Y, Zhai S, Sun Z, Liu Y (2017{\natexlab{a}}) Boosting deep
  learning risk prediction with generative adversarial networks for electronic
  health records. In: IEEE International Conference on Data Mining, pp 787--792

\bibitem[{Che et~al.(2017{\natexlab{b}})Che, He, Xu, and Liu}]{che2017decade}
Che Z, He X, Xu K, Liu Y (2017{\natexlab{b}}) {DECADE}: {A} deep metric
  learning model for multivariate time series. In: KDD Workshop on Mining and
  Learning from Time Series

\bibitem[{Chen et~al.(2013)Chen, Tang, Tino, and Yao}]{chen2013model}
Chen H, Tang F, Tino P, Yao X (2013) Model-based kernel for efficient time
  series analysis. In: ACM SIGKDD International Conference on Knowledge
  Discovery and Data Mining, pp 392--400

\bibitem[{Chen et~al.(2015{\natexlab{a}})Chen, Tang, Ti{\~n}o, Cohn, and
  Yao}]{chen2015model}
Chen H, Tang F, Ti{\~n}o P, Cohn A, Yao X (2015{\natexlab{a}}) Model metric
  co-learning for time series classification. In: International Joint
  Conference on Artificial Intelligence, pp 3387 -- 3394

\bibitem[{Chen et~al.(2015{\natexlab{b}})Chen, Keogh, Hu, Begum, Bagnall,
  Mueen, and Batista}]{ucrarchive}
Chen Y, Keogh E, Hu B, Begum N, Bagnall A, Mueen A, Batista G
  (2015{\natexlab{b}}) The {UCR} time series classification archive.
  \url{www.cs.ucr.edu/~eamonn/time_series_data/}, {A}ccessed 28 Feb 2019

\bibitem[{Chollet(2015)}]{chollet2015keras}
Chollet Fea (2015) Keras. \url{https://keras.io}, {A}ccessed 28 Feb 2019

\bibitem[{{Chouikhi} et~al.(2018){Chouikhi}, {Ammar}, and
  {Alimi}}]{chouikhi2018genesis}
{Chouikhi} N, {Ammar} B, {Alimi} AM (2018) Genesis of basic and multi-layer
  echo state network recurrent autoencoders for efficient data representations.
  ArXiv \eprint{1804.08996}

\bibitem[{{Cristian Borges Gamboa}(2017)}]{gamboa2017deep}
{Cristian Borges Gamboa} J (2017) Deep learning for time-series analysis. ArXiv
  \eprint{1701.01887}

\bibitem[{{Cui} et~al.(2016){Cui}, {Chen}, and {Chen}}]{cui2016multi}
{Cui} Z, {Chen} W, {Chen} Y (2016) Multi-scale convolutional neural networks
  for time series classification. ArXiv \eprint{1603.06995}

\bibitem[{Dau et~al.(2017)Dau, Silva, Petitjean, Forestier, Bagnall, and
  Keogh}]{dau2017judicious}
Dau HA, Silva DF, Petitjean F, Forestier G, Bagnall A, Keogh E (2017) Judicious
  setting of dynamic time warping's window width allows more accurate
  classification of time series. In: IEEE International Conference on Big Data,
  pp 917--922

\bibitem[{{Dau} et~al.(2018){Dau}, {Bagnall}, {Kamgar}, {Yeh}, {Zhu},
  {Gharghabi}, {Ratanamahatana}, and {Keogh}}]{dau2018the}
{Dau} HA, {Bagnall} A, {Kamgar} K, {Yeh} CCM, {Zhu} Y, {Gharghabi} S,
  {Ratanamahatana} CA, {Keogh} E (2018) {The UCR Time Series Archive}. ArXiv
  \eprint{1810.07758}

\bibitem[{Dem\v{s}ar(2006)}]{demsar2006statistical}
Dem\v{s}ar J (2006) Statistical comparisons of classifiers over multiple data
  sets. Machine Learning Research 7:1--30

\bibitem[{Deng et~al.(2013)Deng, Runger, Tuv, and Vladimir}]{deng2013a}
Deng H, Runger G, Tuv E, Vladimir M (2013) A time series forest for
  classification and feature extraction. Information Sciences 239:142 -- 153

\bibitem[{Esling and Agon(2012)}]{esling2012time}
Esling P, Agon C (2012) Time-series data mining. ACM Computing Surveys
  45(1):12:1--12:34

\bibitem[{Faust et~al.(2018)Faust, Hagiwara, Hong, Lih, and
  Acharya}]{faust2018deep}
Faust O, Hagiwara Y, Hong TJ, Lih OS, Acharya UR (2018) Deep learning for
  healthcare applications based on physiological signals: A review. Computer
  Methods and Programs in Biomedicine 161:1 -- 13

\bibitem[{Forestier et~al.(2017)Forestier, Petitjean, Dau, Webb, and
  Keogh}]{forestier2017generating}
Forestier G, Petitjean F, Dau HA, Webb GI, Keogh E (2017) Generating synthetic
  time series to augment sparse datasets. In: IEEE International Conference on
  Data Mining, pp 865--870

\bibitem[{Friedman(1940)}]{friedman1940a}
Friedman M (1940) A comparison of alternative tests of significance for the
  problem of $m$ rankings. The Annals of Mathematical Statistics 11(1):86--92

\bibitem[{{Gallicchio} and {Micheli}(2017)}]{gallicchio2017deep}
{Gallicchio} C, {Micheli} A (2017) Deep echo state network ({DeepESN}): {A}
  brief survey. ArXiv \eprint{1712.04323}

\bibitem[{Garcia and Herrera(2008)}]{garcia2008an}
Garcia S, Herrera F (2008) An extension on ``statistical comparisons of
  classifiers over multiple data sets'' for all pairwise comparisons. Machine
  learning research 9:2677--2694

\bibitem[{{Geng} and {Luo}(2018)}]{geng2018cost}
{Geng} Y, {Luo} X (2018) Cost-sensitive convolution based neural networks for
  imbalanced time-series classification. ArXiv \eprint{1801.04396}

\bibitem[{Glorot and Bengio(2010)}]{glorot2010understanding}
Glorot X, Bengio Y (2010) Understanding the difficulty of training deep
  feedforward neural networks. In: International Conference on Artificial
  Intelligence and Statistics, vol~9, pp 249--256

\bibitem[{Goldberg(2016)}]{goldberg2016a}
Goldberg Y (2016) A primer on neural network models for natural language
  processing. Artificial Intelligence Research 57(1):345--420

\bibitem[{Gong et~al.(2018)Gong, Chen, Yuan, and Yao}]{gong2018Multiobjective}
Gong Z, Chen H, Yuan B, Yao X (2018) Multiobjective learning in the model space
  for time series classification. IEEE Transactions on Cybernetics 99:1--15

\bibitem[{Grabocka et~al.(2014)Grabocka, Schilling, Wistuba, and
  Schmidt-Thieme}]{grabocka2014learning}
Grabocka J, Schilling N, Wistuba M, Schmidt-Thieme L (2014) Learning
  time-series shapelets. In: ACM SIGKDD International Conference on Knowledge
  Discovery and Data Mining, pp 392--401

\bibitem[{{Hatami} et~al.(2017){Hatami}, {Gavet}, and
  {Debayle}}]{hatami2017classification}
{Hatami} N, {Gavet} Y, {Debayle} J (2017) {Classification of time-series images
  using deep convolutional neural networks}. In: International Conference on
  Machine Vision

\bibitem[{He et~al.(2015)He, Zhang, Ren, and Sun}]{he2015delving}
He K, Zhang X, Ren S, Sun J (2015) Delving deep into rectifiers: Surpassing
  human-level performance on imagenet classification. In: IEEE International
  Conference on Computer Vision, pp 1026--1034

\bibitem[{He et~al.(2016)He, Zhang, Ren, and Sun}]{he2016deep}
He K, Zhang X, Ren S, Sun J (2016) Deep residual learning for image
  recognition. In: IEEE Conference on Computer Vision and Pattern Recognition,
  pp 770--778

\bibitem[{Hills et~al.(2014)Hills, Lines, Baranauskas, Mapp, and
  Bagnall}]{hills2014classification}
Hills J, Lines J, Baranauskas E, Mapp J, Bagnall A (2014) Classification of
  time series by shapelet transformation. Data Mining and Knowledge Discovery
  28(4):851--881

\bibitem[{Hinton et~al.(2012)Hinton, Deng, Yu, Dahl, Mohamed, Jaitly, Senior,
  Vanhoucke, Nguyen, Sainath, and Kingsbury}]{hinton2012deep}
Hinton G, Deng L, Yu D, Dahl GE, Mohamed AR, Jaitly N, Senior A, Vanhoucke V,
  Nguyen P, Sainath TN, Kingsbury B (2012) Deep neural networks for acoustic
  modeling in speech recognition: The shared views of four research groups.
  IEEE Signal Processing Magazine 29(6):82--97

\bibitem[{Hoerl and Kennard(1970)}]{hoerl1970ridge}
Hoerl AE, Kennard RW (1970) Ridge regression: Applications to nonorthogonal
  problems. Technometrics 12(1):69--82

\bibitem[{Holm(1979)}]{holm1979a}
Holm S (1979) A simple sequentially rejective multiple test procedure.
  Scandinavian Journal of Statistics 6(2):65--70

\bibitem[{H{\"o}ppner(2016)}]{hoppner2016improving}
H{\"o}ppner F (2016) Improving time series similarity measures by integrating
  preprocessing steps. Data Mining and Knowledge Discovery 31:851--878

\bibitem[{Hu et~al.(2016)Hu, Zhang, and Zhou}]{hu2016transfer}
Hu Q, Zhang R, Zhou Y (2016) Transfer learning for short-term wind speed
  prediction with deep neural networks. Renewable Energy 85:83 -- 95

\bibitem[{Ignatov(2018)}]{ignatov2018activity}
Ignatov A (2018) Real-time human activity recognition from accelerometer data
  using convolutional neural networks. Applied Soft Computing 62:915 -- 922

\bibitem[{Ioffe and Szegedy(2015)}]{ioffe2015batch}
Ioffe S, Szegedy C (2015) Batch normalization: Accelerating deep network
  training by reducing internal covariate shift. In: International Conference
  on Machine Learning, vol~37, pp 448--456

\bibitem[{Ismail~Fawaz et~al.(2018{\natexlab{a}})Ismail~Fawaz, Forestier,
  Weber, Idoumghar, and Muller}]{ismailFawaz2018data}
Ismail~Fawaz H, Forestier G, Weber J, Idoumghar L, Muller PA
  (2018{\natexlab{a}}) Data augmentation using synthetic data for time series
  classification with deep residual networks. In: International Workshop on
  Advanced Analytics and Learning on Temporal Data, {ECML} {PKDD}

\bibitem[{Ismail~Fawaz et~al.(2018{\natexlab{b}})Ismail~Fawaz, Forestier,
  Weber, Idoumghar, and Muller}]{ismailFawaz2018evaluating}
Ismail~Fawaz H, Forestier G, Weber J, Idoumghar L, Muller PA
  (2018{\natexlab{b}}) Evaluating surgical skills from kinematic data using
  convolutional neural networks. In: Medical Image Computing and Computer
  Assisted Intervention, pp 214--221

\bibitem[{Ismail~Fawaz et~al.(2018{\natexlab{c}})Ismail~Fawaz, Forestier,
  Weber, Idoumghar, and Muller}]{IsmailFawaz2018transfer}
Ismail~Fawaz H, Forestier G, Weber J, Idoumghar L, Muller PA
  (2018{\natexlab{c}}) Transfer learning for time series classification. In:
  IEEE International Conference on Big Data, pp 1367--1376

\bibitem[{Jaeger and Haas(2004)}]{jaeger2004Harnessing}
Jaeger H, Haas H (2004) Harnessing nonlinearity: Predicting chaotic systems and
  saving energy in wireless communication. Science 304(5667):78--80

\bibitem[{Kate(2016)}]{kate2016using}
Kate RJ (2016) Using dynamic time warping distances as features for improved
  time series classification. Data Mining and Knowledge Discovery
  30(2):283--312

\bibitem[{Keogh and Mueen(2017)}]{keogh2017curse}
Keogh E, Mueen A (2017) Curse of dimensionality. Encyclopedia of Machine
  Learning and Data Mining pp 314--315

\bibitem[{{Kim}(2014)}]{kim2014convolutional}
{Kim} Y (2014) Convolutional neural networks for sentence classification. In:
  Empirical Methods in Natural Language Processing

\bibitem[{Kingma and {Ba}(2015)}]{kingma2015adam}
Kingma DP, {Ba} J (2015) Adam: A method for stochastic optimization. In:
  International Conference on Learning Representations

\bibitem[{Kotsifakos and Papapetrou(2014)}]{kotsifakos2014model}
Kotsifakos A, Papapetrou P (2014) Model-based time series classification. In:
  Advances in Intelligent Data Analysis, pp 179--191

\bibitem[{Krasin et~al.(2017)Krasin, Duerig, Alldrin, Ferrari, Abu-El-Haija,
  Kuznetsova, Rom, Uijlings, Popov, Kamali, Malloci, Pont-Tuset, Veit,
  Belongie, Gomes, Gupta, Sun, Chechik, Cai, Feng, Narayanan, and
  Murphy}]{openimages}
Krasin I, Duerig T, Alldrin N, Ferrari V, Abu-El-Haija S, Kuznetsova A, Rom H,
  Uijlings J, Popov S, Kamali S, Malloci M, Pont-Tuset J, Veit A, Belongie S,
  Gomes V, Gupta A, Sun C, Chechik G, Cai D, Feng Z, Narayanan D, Murphy K
  (2017) {OpenImages}: A public dataset for large-scale multi-label and
  multi-class image classification.
  \url{https://storage.googleapis.com/openimages/web/index.html}, {A}ccessed 28
  Feb 2019

\bibitem[{Krizhevsky et~al.(2012)Krizhevsky, Sutskever, and
  Hinton}]{krizhevsky2012imagenet}
Krizhevsky A, Sutskever I, Hinton GE (2012) {ImageNet} classification with deep
  convolutional neural networks. In: Advances in Neural Information Processing
  Systems 25, pp 1097--1105

\bibitem[{Kruskal and Wish(1978)}]{kruskal1978multidimensional}
Kruskal JB, Wish M (1978) Multidimensional scaling. number 07--011 in sage
  university paper series on quantitative applications in the social sciences

\bibitem[{L\"angkvist et~al.(2014)L\"angkvist, Karlsson, and
  Loutfi}]{langkvist2014a}
L\"angkvist M, Karlsson L, Loutfi A (2014) A review of unsupervised feature
  learning and deep learning for time-series modeling. Pattern Recognition
  Letters 42:11 -- 24

\bibitem[{{Large} et~al.(2017){Large}, {Lines}, and {Bagnall}}]{large2017the}
{Large} J, {Lines} J, {Bagnall} A (2017) {The Heterogeneous Ensembles of
  Standard Classification Algorithms (HESCA): the Whole is Greater than the Sum
  of its Parts}. ArXiv \eprint{1710.09220}

\bibitem[{Le and Mikolov(2014)}]{le2014distributed}
Le Q, Mikolov T (2014) Distributed representations of sentences and documents.
  In: International Conference on Machine Learning, vol~32, pp
  II--1188--II--1196

\bibitem[{Le~Guennec et~al.(2016)Le~Guennec, Malinowski, and
  Tavenard}]{leguennec2016data}
Le~Guennec A, Malinowski S, Tavenard R (2016) Data augmentation for time series
  classification using convolutional neural networks. In: {ECML/PKDD Workshop
  on Advanced Analytics and Learning on Temporal Data}

\bibitem[{{LeCun} et~al.(1998{\natexlab{a}}){LeCun}, Bottou, Bengio, and
  Haffner}]{lecun1998gradient}
{LeCun} Y, Bottou L, Bengio Y, Haffner P (1998{\natexlab{a}}) Gradient-based
  learning applied to document recognition. Proceedings of the IEEE
  86(11):2278--2324

\bibitem[{{LeCun} et~al.(1998{\natexlab{b}}){LeCun}, Bottou, Orr, and
  M\"{u}ller}]{lecun1998efficient}
{LeCun} Y, Bottou L, Orr GB, M\"{u}ller KR (1998{\natexlab{b}}) Efficient
  backprop. In: Neural Networks: Tricks of the Trade, Springer, Berlin, pp
  9--50

\bibitem[{{LeCun} et~al.(2015){LeCun}, Bengio, and Hinton}]{lecun2015deep}
{LeCun} Y, Bengio Y, Hinton G (2015) Deep learning. Nature 521:436--444

\bibitem[{Lin and Runger(2018)}]{lin2018gcrnn}
Lin S, Runger GC (2018) {GCRNN}: Group-constrained convolutional recurrent
  neural network. IEEE Transactions on Neural Networks and Learning Systems
  99:1--10

\bibitem[{Lines and Bagnall(2015)}]{lines2015time}
Lines J, Bagnall A (2015) Time series classification with ensembles of elastic
  distance measures. Data Mining and Knowledge Discovery 29(3):565--592

\bibitem[{Lines et~al.(2016)Lines, Taylor, and Bagnall}]{lines2016hive}
Lines J, Taylor S, Bagnall A (2016) {HIVE-COTE}: The hierarchical vote
  collective of transformation-based ensembles for time series classification.
  In: IEEE International Conference on Data Mining, pp 1041--1046

\bibitem[{Lines et~al.(2018)Lines, Taylor, and Bagnall}]{lines2018time}
Lines J, Taylor S, Bagnall A (2018) Time series classification with
  {HIVE-COTE}: The hierarchical vote collective of transformation-based
  ensembles. ACM Transactions on Knowledge Discovery from Data
  12(5):52:1--52:35

\bibitem[{Liu et~al.(2018)Liu, Hsaio, and Tu}]{liu2018time}
Liu C, Hsaio W, Tu Y (2018) Time series classification with multivariate
  convolutional neural network. IEEE Transactions on Industrial Electronics
  66:1--1

\bibitem[{Lu et~al.(2015)Lu, Young, Arel, and Holleman}]{lu2015a}
Lu J, Young S, Arel I, Holleman J (2015) A 1 tops/w analog deep
  machine-learning engine with floating-gate storage in 0.13 $\mu$m cmos. IEEE
  Journal of Solid-State Circuits 50(1):270--281

\bibitem[{{Lucas} et~al.(2018){Lucas}, {Shifaz}, {Pelletier}, {O'Neill},
  {Zaidi}, {Goethals}, {Petitjean}, and {Webb}}]{lucas2018proximity}
{Lucas} B, {Shifaz} A, {Pelletier} C, {O'Neill} L, {Zaidi} N, {Goethals} B,
  {Petitjean} F, {Webb} GI (2018) {Proximity Forest: An effective and scalable
  distance-based classifier for time series}. Data Mining and Knowledge
  Discovery 28:851--881

\bibitem[{Ma et~al.(2016)Ma, Shen, Chen, Wang, Wei, and Yu}]{ma2016functional}
Ma Q, Shen L, Chen W, Wang J, Wei J, Yu Z (2016) Functional echo state network
  for time series classification. Information Sciences 373:1 -- 20

\bibitem[{{Malhotra} et~al.(2018){Malhotra}, {TV}, {Vig}, {Agarwal}, and
  {Shroff}}]{malhotra2017timenet}
{Malhotra} P, {TV} V, {Vig} L, {Agarwal} P, {Shroff} G (2018) {TimeNet:
  Pre-trained deep recurrent neural network for time series classification}.
  In: European Symposium on Artificial Neural Networks, Computational
  Intelligence and Machine Learning, pp 607--612

\bibitem[{Martinez et~al.(2018)Martinez, Perrin, Ramasso, and
  Rombaut}]{martinez2018a}
Martinez C, Perrin G, Ramasso E, Rombaut M (2018) {A deep reinforcement
  learning approach for early classification of time series}. In: {European
  Signal Processing Conference}

\bibitem[{Mehdiyev et~al.(2017)Mehdiyev, Lahann, Emrich, Enke, Fettke, and
  Loos}]{mehdiyev2017time}
Mehdiyev N, Lahann J, Emrich A, Enke D, Fettke P, Loos P (2017) Time series
  classification using deep learning for process planning: A case from the
  process industry. Procedia Computer Science 114:242 -- 249

\bibitem[{{Mikolov} et~al.(2013){Mikolov}, {Chen}, {Corrado}, and
  {Dean}}]{mikolov2013efficient}
{Mikolov} T, {Chen} K, {Corrado} G, {Dean} J (2013) {Efficient Estimation of
  Word Representations in Vector Space}. In: International Conference on
  Learning Representations - Workshop

\bibitem[{Mikolov et~al.(2013)Mikolov, Sutskever, Chen, Corrado, and
  Dean}]{mikolov2013distributed}
Mikolov T, Sutskever I, Chen K, Corrado G, Dean J (2013) Distributed
  representations of words and phrases and their compositionality. In: Neural
  Information Processing Systems, pp 3111--3119

\bibitem[{{Mittelman}(2015)}]{mittelman2015time}
{Mittelman} R (2015) {Time-series modeling with undecimated fully convolutional
  neural networks}. ArXiv \eprint{1508.00317}

\bibitem[{Neamtu et~al.(2018)Neamtu, Ahsan, Rundensteiner, Sarkozy, Keogh, Dau,
  Nguyen, and Lovering}]{neamtu2018generalized}
Neamtu R, Ahsan R, Rundensteiner EA, Sarkozy G, Keogh E, Dau HA, Nguyen C,
  Lovering C (2018) Generalized dynamic time warping: Unleashing the warping
  power hidden in point-wise distances. In: IEEE International Conference on
  Data Engineering

\bibitem[{Nguyen et~al.(2017)Nguyen, Gsponer, and Ifrim}]{nguyen2017time}
Nguyen TL, Gsponer S, Ifrim G (2017) Time series classification by sequence
  learning in all-subsequence space. In: IEEE International Conference on Data
  Engineering, pp 947--958

\bibitem[{Nwe et~al.(2017)Nwe, Dat, and Ma}]{nwe2017convolutional}
Nwe TL, Dat TH, Ma B (2017) Convolutional neural network with multi-task
  learning scheme for acoustic scene classification. In: Asia-Pacific Signal
  and Information Processing Association Annual Summit and Conference, pp
  1347--1350

\bibitem[{Nweke et~al.(2018)Nweke, Teh, Al-garadi, and Alo}]{nweke2018deep}
Nweke HF, Teh YW, Al-garadi MA, Alo UR (2018) Deep learning algorithms for
  human activity recognition using mobile and wearable sensor networks: State
  of the art and research challenges. Expert Systems with Applications 105:233
  -- 261

\bibitem[{Ord\'o\"nez and Roggen(2016)}]{ordonez2016deep}
Ord\'o\"nez FJ, Roggen D (2016) Deep convolutional and {LSTM} recurrent neural
  networks for multimodal wearable activity recognition. Sensors 16:115

\bibitem[{Pan and Yang(2010)}]{pan2010a}
Pan SJ, Yang Q (2010) A survey on transfer learning. IEEE Transactions on
  Knowledge and Data Engineering 22(10):1345--1359

\bibitem[{{Papernot} and {McDaniel}(2018)}]{papernot2018deep}
{Papernot} N, {McDaniel} P (2018) Deep k-nearest neighbors: Towards confident,
  interpretable and robust deep learning. ArXiv \eprint{1803.04765}

\bibitem[{Pascanu et~al.(2012)Pascanu, Mikolov, and
  Bengio}]{pascanu2012understanding}
Pascanu R, Mikolov T, Bengio Y (2012) Understanding the exploding gradient
  problem. ArXiv \eprint{1211.5063}

\bibitem[{Pascanu et~al.(2013)Pascanu, Mikolov, and Bengio}]{pascanu2013on}
Pascanu R, Mikolov T, Bengio Y (2013) On the difficulty of training recurrent
  neural networks. In: International Conference on Machine Learning, vol~28, pp
  III--1310--III--1318

\bibitem[{Petitjean et~al.(2016)Petitjean, Forestier, Webb, Nicholson, Chen,
  and Keogh}]{petitjean2016faster}
Petitjean F, Forestier G, Webb GI, Nicholson AE, Chen Y, Keogh E (2016) Faster
  and more accurate classification of time series by exploiting a novel dynamic
  time warping averaging algorithm. Knowledge Information Systems 47(1):1--26

\bibitem[{Poggio et~al.(2017)Poggio, Mhaskar, Rosasco, Miranda, and
  Liao}]{poggio2017why}
Poggio T, Mhaskar H, Rosasco L, Miranda B, Liao Q (2017) Why and when can
  deep-but not shallow-networks avoid the curse of dimensionality: A review.
  International Journal of Automation and Computing 14(5):503--519

\bibitem[{Rajan and Thiagarajan(2018)}]{rajan2018a}
Rajan D, Thiagarajan J (2018) A generative modeling approach to limited channel
  ecg classification. In: IEEE Engineering in Medicine and Biology Society, vol
  2018, p 2571

\bibitem[{{Rajkomar} et~al.(2018){Rajkomar}, {Oren}, {Chen}, {Dai}, {Hajaj},
  {Liu}, {Liu}, {Sun}, {Sundberg}, {Yee}, {Zhang}, {Duggan}, {Flores}, {Hardt},
  {Irvine}, {Le}, {Litsch}, {Marcus}, {Mossin}, {Tansuwan}, {Wang}, {Wexler},
  {Wilson}, {Ludwig}, {Volchenboum}, {Chou}, {Pearson}, {Madabushi}, {Shah},
  {Butte}, {Howell}, {Cui}, {Corrado}, and {Dean}}]{rajkomar2018scalable}
{Rajkomar} A, {Oren} E, {Chen} K, {Dai} AM, {Hajaj} N, {Liu} PJ, {Liu} X, {Sun}
  M, {Sundberg} P, {Yee} H, {Zhang} K, {Duggan} GE, {Flores} G, {Hardt} M,
  {Irvine} J, {Le} Q, {Litsch} K, {Marcus} J, {Mossin} A, {Tansuwan} J, {Wang}
  D, {Wexler} J, {Wilson} J, {Ludwig} D, {Volchenboum} SL, {Chou} K, {Pearson}
  M, {Madabushi} S, {Shah} NH, {Butte} AJ, {Howell} M, {Cui} C, {Corrado} G,
  {Dean} J (2018) {Scalable and accurate deep learning for electronic health
  records}. NPJ Digital Medicine 1:18

\bibitem[{Ratanamahatana and Keogh(2005)}]{ratanamahatana2005three}
Ratanamahatana CA, Keogh E (2005) Three myths about dynamic time warping data
  mining. In: SIAM International Conference on Data Mining, pp 506--510

\bibitem[{Rowe and Abbott(1995)}]{alistair1995daubechies}
Rowe ACH, Abbott PC (1995) Daubechies wavelets and mathematica. Computers in
  Physics 9(6):635--648

\bibitem[{Russakovsky et~al.(2015)Russakovsky, Deng, Su, Krause, Satheesh, Ma,
  Huang, Karpathy, Khosla, Bernstein, Berg, and
  Fei-Fei}]{russakovsky2015imagenet}
Russakovsky O, Deng J, Su H, Krause J, Satheesh S, Ma S, Huang Z, Karpathy A,
  Khosla A, Bernstein M, Berg AC, Fei-Fei L (2015) {ImageNet} large scale
  visual recognition challenge. International Journal of Computer Vision
  115(3):211--252

\bibitem[{Sainath et~al.(2013)Sainath, Mohamed, Kingsbury, and
  Ramabhadran}]{sainath2013deep}
Sainath TN, Mohamed AR, Kingsbury B, Ramabhadran B (2013) Deep convolutional
  neural networks for {LVCSR}. In: IEEE International Conference on Acoustics,
  Speech and Signal Processing, pp 8614--8618

\bibitem[{Santos and Kern(2017)}]{santos2017a}
Santos T, Kern R (2017) A literature survey of early time series classification
  and deep learning. In: International Conference on Knowledge Technologies and
  Data-driven Business

\bibitem[{Sch{\"a}fer(2015)}]{schafer2015the}
Sch{\"a}fer P (2015) The {BOSS} is concerned with time series classification in
  the presence of noise. Data Mining and Knowledge Discovery 29(6):1505--1530

\bibitem[{{Serr{\`a}} et~al.(2018){Serr{\`a}}, {Pascual}, and
  {Karatzoglou}}]{serra2018towards}
{Serr{\`a}} J, {Pascual} S, {Karatzoglou} A (2018) {Towards a universal neural
  network encoder for time series}. Artificial Intelligence Research and
  Development: Current Challenges, New Trends and Applications 308:120

\bibitem[{Silva et~al.(2018)Silva, Giusti, Keogh, and
  Batista}]{silva2018speeding}
Silva DF, Giusti R, Keogh E, Batista G (2018) Speeding up similarity search
  under dynamic time warping by pruning unpromising alignments. Data Mining and
  Knowledge Discovery 32(4):988--1016

\bibitem[{Srivastava et~al.(2014)Srivastava, Hinton, Krizhevsky, Sutskever, and
  Salakhutdinov}]{srivastava2014a}
Srivastava N, Hinton G, Krizhevsky A, Sutskever I, Salakhutdinov R (2014)
  Dropout: A simple way to prevent neural networks from overfitting. Journal of
  Machine Learning Research 15:1929--1958

\bibitem[{Strodthoff and Strodthoff(2019)}]{strodthoff2018detecting}
Strodthoff N, Strodthoff C (2019) Detecting and interpreting myocardial
  infarction using fully convolutional neural networks. Physiological
  Measurement 40(1):015001

\bibitem[{Susto et~al.(2018)Susto, Cenedese, and Terzi}]{susto2018time}
Susto GA, Cenedese A, Terzi M (2018) Time-series classification methods: Review
  and applications to power systems data. In: Big Data Application in Power
  Systems, pp 179 -- 220

\bibitem[{Sutskever et~al.(2014)Sutskever, Vinyals, and
  Le}]{sutskever2014sequence}
Sutskever I, Vinyals O, Le QV (2014) Sequence to sequence learning with neural
  networks. In: Neural Information Processing Systems, pp 3104--3112

\bibitem[{Szegedy et~al.(2015)Szegedy, Liu, Jia, Sermanet, Reed, Anguelov,
  Erhan, Vanhoucke, and Rabinovich}]{szegedy2015going}
Szegedy C, Liu W, Jia Y, Sermanet P, Reed S, Anguelov D, Erhan D, Vanhoucke V,
  Rabinovich A (2015) Going deeper with convolutions. In: IEEE Conference on
  Computer Vision and Pattern Recognition, pp 1--9

\bibitem[{Tanisaro and Heidemann(2016)}]{tanisaro2016time}
Tanisaro P, Heidemann G (2016) Time series classification using time warping
  invariant echo state networks. In: IEEE International Conference on Machine
  Learning and Applications, pp 831--836

\bibitem[{Tripathy and Acharya(2018)}]{tripathy2018use}
Tripathy R, Acharya UR (2018) Use of features from {RR}-time series and {EEG}
  signals for automated classification of sleep stages in deep neural network
  framework. Biocybernetics and Biomedical Engineering 38:890--902

\bibitem[{Uemura et~al.(2018)Uemura, Tomikawa, Miao, Souzaki, Ieiri, Akahoshi,
  Lefor, and Hashizume}]{uemura2018Feasibility}
Uemura M, Tomikawa M, Miao T, Souzaki R, Ieiri S, Akahoshi T, Lefor AK,
  Hashizume M (2018) Feasibility of an {AI}-based measure of the hand motions
  of expert and novice surgeons. Computational and Mathematical Methods in
  Medicine 2018

\bibitem[{{Ulyanov} et~al.(2016){Ulyanov}, {Vedaldi}, and
  {Lempitsky}}]{ulyanov2016instance}
{Ulyanov} D, {Vedaldi} A, {Lempitsky} V (2016) Instance normalization: The
  missing ingredient for fast stylization. ArXiv \eprint{1607.08022}

\bibitem[{Vaswani et~al.(2017)Vaswani, Shazeer, Parmar, Uszkoreit, Jones,
  Gomez, Kaiser, and Polosukhin}]{vaswani2017attention}
Vaswani A, Shazeer N, Parmar N, Uszkoreit J, Jones L, Gomez AN, Kaiser Lu,
  Polosukhin I (2017) Attention is all you need. In: Advances in Neural
  Information Processing Systems, pp 5998--6008

\bibitem[{Wang et~al.(2018)Wang, Chen, Hao, Peng, and Hu}]{wang2018deep}
Wang J, Chen Y, Hao S, Peng X, Hu L (2018) Deep learning for sensor-based
  activity recognition: A survey. Pattern Recognition Letters

\bibitem[{{Wang} et~al.(2018){Wang}, {Wang}, {Li}, and
  {Wu}}]{wang2018multilevel}
{Wang} J, {Wang} Z, {Li} J, {Wu} J (2018) Multilevel wavelet decomposition
  network for interpretable time series analysis. In: {ACM SIGKDD International
  Conference on Knowledge Discovery and Data Mining}, pp 2437--2446

\bibitem[{Wang et~al.(2016)Wang, Wang, and Liu}]{wang2016an}
Wang L, Wang Z, Liu S (2016) An effective multivariate time series
  classification approach using echo state network and adaptive differential
  evolution algorithm. Expert Systems with Applications 43:237 -- 249

\bibitem[{Wang et~al.(2017{\natexlab{a}})Wang, Hua, Hao, and Xie}]{wang2017a}
Wang S, Hua G, Hao G, Xie C (2017{\natexlab{a}}) A cycle deep belief network
  model for multivariate time series classification. Mathematical Problems in
  Engineering 2017:1--7

\bibitem[{{Wang} et~al.(2016{\natexlab{a}}){Wang}, {Chen}, {Wang}, {Rai}, and
  {Carin}}]{wang2016earliness}
{Wang} W, {Chen} C, {Wang} W, {Rai} P, {Carin} L (2016{\natexlab{a}})
  Earliness-aware deep convolutional networks for early time series
  classification. ArXiv \eprint{1611.04578}

\bibitem[{Wang and Oates(2015{\natexlab{a}})}]{wang2015Encoding}
Wang Z, Oates T (2015{\natexlab{a}}) Encoding time series as images for visual
  inspection and classification using tiled convolutional neural networks. In:
  Workshops at AAAI Conference on Artificial Intelligence, pp 40--46

\bibitem[{Wang and Oates(2015{\natexlab{b}})}]{wang2015imaging}
Wang Z, Oates T (2015{\natexlab{b}}) Imaging time-series to improve
  classification and imputation. In: International Conference on Artificial
  Intelligence, pp 3939--3945

\bibitem[{{Wang} and {Oates}(2015)}]{wang2015spatially}
{Wang} Z, {Oates} T (2015) Spatially encoding temporal correlations to classify
  temporal data using convolutional neural networks. ArXiv \eprint{1509.07481}

\bibitem[{{Wang} et~al.(2016{\natexlab{b}}){Wang}, {Song}, {Liu}, {Zhang},
  {Xue}, {Ye}, {Fan}, and {Xu}}]{wang2016representation}
{Wang} Z, {Song} W, {Liu} L, {Zhang} F, {Xue} J, {Ye} Y, {Fan} M, {Xu} M
  (2016{\natexlab{b}}) Representation learning with deconvolution for
  multivariate time series classification and visualization. ArXiv
  \eprint{1610.07258}

\bibitem[{Wang et~al.(2017{\natexlab{b}})Wang, Yan, and Oates}]{wang2017time}
Wang Z, Yan W, Oates T (2017{\natexlab{b}}) Time series classification from
  scratch with deep neural networks: A strong baseline. In: International Joint
  Conference on Neural Networks, pp 1578--1585

\bibitem[{Wilcoxon(1945)}]{wilcoxon1945individual}
Wilcoxon F (1945) Individual comparisons by ranking methods. Biometrics
  Bulletin 1(6):80--83

\bibitem[{Yang and Wu(2006)}]{yang200610}
Yang Q, Wu X (2006) 10 challenging problems in data mining research.
  Information Technology \& Decision Making 05(04):597--604

\bibitem[{Ye and Keogh(2011)}]{ye2011time}
Ye L, Keogh E (2011) Time series shapelets: a novel technique that allows
  accurate, interpretable and fast classification. Data Mining and Knowledge
  Discovery 22(1):149--182

\bibitem[{Yosinski et~al.(2014)Yosinski, Clune, Bengio, and
  Lipson}]{yosinski2014how}
Yosinski J, Clune J, Bengio Y, Lipson H (2014) How transferable are features in
  deep neural networks? In: International Conference on Neural Information
  Processing Systems, vol~2, pp 3320--3328

\bibitem[{{Zeiler}(2012)}]{zeiler2012adadelta}
{Zeiler} MD (2012) {ADADELTA}: An adaptive learning rate method. ArXiv
  \eprint{1212.5701}

\bibitem[{Zhang et~al.(2017)Zhang, Bengio, Hardt, Recht, and
  Vinyals}]{zhang2017understanding}
Zhang C, Bengio S, Hardt M, Recht B, Vinyals O (2017) {Understanding deep
  learning requires rethinking generalization}. In: International Conference on
  Learning Representations

\bibitem[{Zhao et~al.(2017)Zhao, Lu, Chen, Liu, and Wu}]{zhao2017convolutional}
Zhao B, Lu H, Chen S, Liu J, Wu D (2017) Convolutional neural networks for time
  series classification. Systems Engineering and Electronics 28(1):162--169

\bibitem[{Zheng et~al.(2014)Zheng, Liu, Chen, Ge, and Zhao}]{zheng2014time}
Zheng Y, Liu Q, Chen E, Ge Y, Zhao JL (2014) Time series classification using
  multi-channels deep convolutional neural networks. In: Web-Age Information
  Management, pp 298--310

\bibitem[{Zheng et~al.(2016)Zheng, Liu, Chen, Ge, and
  Zhao}]{zheng2016exploiting}
Zheng Y, Liu Q, Chen E, Ge Y, Zhao JL (2016) Exploiting multi-channels deep
  convolutional neural networks for multivariate time series classification.
  Frontiers of Computer Science 10(1):96--112

\bibitem[{Zhou et~al.(2016)Zhou, Khosla, Lapedriza, Oliva, and
  Torralba}]{zhou2016learning}
Zhou B, Khosla A, Lapedriza A, Oliva A, Torralba A (2016) Learning deep
  features for discriminative localization. In: IEEE Conference on Computer
  Vision and Pattern Recognition, pp 2921--2929

\bibitem[{Ziat et~al.(2017)Ziat, Delasalles, Denoyer, and
  Gallinari}]{ziat2017spatio}
Ziat A, Delasalles E, Denoyer L, Gallinari P (2017) Spatio-temporal neural
  networks for space-time series forecasting and relations discovery. In: IEEE
  International Conference on Data Mining, pp 705--714

\end{thebibliography}

\end{document}